UNIVERSITÄT TRIER

FACHBEREICH II

COMPUTERLINGUISTIK & DIGITAL HUMANITIES

# Language Segmentation

*Author:*
David ALFTER

*Supervisors:*
Prof. Dr. Caroline SPORLEDER
Dr. Sven NAUMANN

August 18, 2015

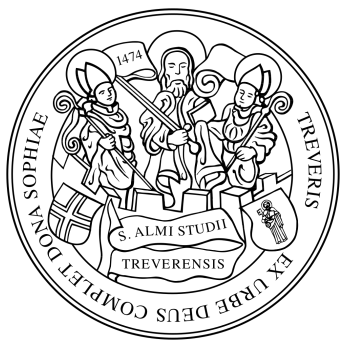

## Erklärung zur Masterarbeit

Hiermit erkläre ich, dass ich die Masterarbeit selbstständig verfasst und keine anderen als die angegebenen Quellen und Hilfsmittel benutzt und die aus fremden Quellen direkt oder indirekt übernommenen Gedanken als solche kenntlich gemacht habe.

Die Arbeit habe ich bisher keinem anderen Prüfungsamt in gleicher oder vergleichbarer Form vorgelegt. Sie wurde bisher nicht veröffentlicht.

_________________________
Datum

_________________________
Unterschrift

**Abstract**

Language segmentation consists in finding the boundaries where one language ends and another language begins in a text written in more than one language. This is important for all natural language processing tasks.

The problem can be solved by training language models on language data. However, in the case of low- or no-resource languages, this is problematic. I therefore investigate whether unsupervised methods perform better than supervised methods when it is difficult or impossible to train supervised approaches.

A special focus is given to *difficult* texts, i.e. texts that are rather short (one sentence), containing abbreviations, low-resource languages and non-standard language.

I compare three approaches: supervised n-gram language models, unsupervised clustering and weakly supervised n-gram language model induction. I devised the weakly supervised approach in order to deal with difficult text specifically. In order to test the approach, I compiled a small corpus of different text types, ranging from one-sentence texts to texts of about 300 words.

The weakly supervised language model induction approach works well on short and difficult texts, outperforming the clustering algorithm and reaching scores in the vicinity of the supervised approach. The results look promising, but there is room for improvement and a more thorough investigation should be undertaken.

## Acknowledgements

My thanks go to professor Caroline Sporleder for sharing her knowledge with me, for her inspiring ideas and for agreeing to supervise my Bachelor's and Master's Thesis despite her busy schedule. It was also thanks to the topic she suggested for my Bachelor's Thesis that I met Jürgen Knauth and later was able to get a research assistant position at the SeNeReKo project, collaborating closely with Jürgen.

Which brings me to the next person on the list. I would like to thank Jürgen Knauth for the wonderful collaboration, for his patience, for his contagious enthusiasm, and all the interesting conversations in passing that always lasted longer than intended.

I would like to thank Stephan Faber for his insightful comments when I couldn't see the wood for the trees, for his patience and optimism, for pushing me to go further and to persevere.

I would also like to thank Julian Vaudroz for accompanying me throughout the degree program. We both didn't know what we were in for when we started, but we persevered and it paid off. It wouldn't have been the same without you.

Finally, I would like to thank all the people that volunteered to proofread my thesis and all the people that helped me during the writing of this thesis. Unfortunately, I cannot list everyone. You know who you are!

# List of Figures

# List of Tables

# List of Algorithms

# Contents

# 1   Introduction

Language segmentation and identification are important for all natural language processing operations that are language-specific, such as taggers, parsers or machine translation (Jain and Bhat, 2014; Zubiaga et al., 2014). Indeed, using "traditional" monolingual natural language processing components on mixed language data leads to miserable results (Jain and Bhat, 2014). Even if the results are not terrible, language identification and segmentation can improve the overall results. For example, by identifying foreign language inclusions in an otherwise monolingual text, parser accuracy can be increased (Alex et al., 2007).

One important point that has to be borne in mind is the difference between language identification and language segmentation. Language identification is concerned with recognizing the language at hand. It is possible to use language identification for language segmentation. Indeed, by identifying the languages in a text, the segmentation is implicitly obtained. Language segmentation on the other hand is only concerned with identifying language boundaries. No claims about the languages involved are made.

After giving an overview over related work and different approaches that can be taken for language segmentation, I will present the theory behind supervised methods as well as unsupervised methods. Finally, I will introduce a weakly supervised method for language segmentation that I developed.

After the theoretical part, I will present experiments done with the different approaches, comparing their effectiveness on the task of language segmentation on different text types. A special focus will be given to *difficult* text types, such as short texts, texts containing under-resourced languages or texts containing a lot of abbreviations or other non-standard features.

A big advantage of unsupervised methods is language independence. If the approach used does not rely on language-specific details, the approach is more flexible as no language resources have to be adapted for the method to work on other languages. These advantages might be especially useful for under-resourced languages. When there is no or insufficient data available to train a supervised language model, an unsupervised approach might yield better results.

Another advantage is that unsupervised methods do not require prior training. They are not dependent on training data and thus cannot be skewed by the data. Indeed, supervised approaches that are trained on data are qualitatively tied to their training data; different training data will, in all probability, yield different models.

This thesis aims at answering the question whether unsupervised language segmentation approaches work better on difficult text types than supervised language approaches.

## 2 Related work

### 2.1 N-Grams and rank order statistics

Cavnar and Trenkle (1994) use an n-gram language model for language identification purposes. Their program 'Textcat' is intended to classify documents by language. The system calculates n-grams for $1 \leqslant n \leqslant 5$ from training data and orders the n-grams according to inverse frequency, i.e. from the most frequent n-grams to the most infrequent n-grams. The numerical frequency data is then discarded and only inherently present.

During training, the program calculates an n-gram profile consisting of these n-gram lists for each category (i.e. language to classify).

New data is classified by first calculating the n-gram profile and then comparing the profile to existing profiles. The category with the lowest difference score is taken as the category for the document.

The score they use for classification is called *out-of-place* metric. For each n-gram in the document n-gram profile, the corresponding n-gram in the category profile is looked up and the absolute difference of ranks is taken as score. The sum is calculated over all n-grams. More formally, the out-of-place metric $m_{oop}$ is calculated as:

$$m_{oop} = \sum_{i=1}^{n} (|r(x_i, d) - r(x_i, c)|)$$ (1)

With $n$ the number of n-grams in the document profile, $x_i$ the $i$-th n-gram, $r(x_i, d)$ the rank of the $i$-th n-gram in the document profile, $r(x_i, c)$ the rank of the $i$-th n-gram in the category profile.

Figure 1 illustrates the *out-of-place* metric.

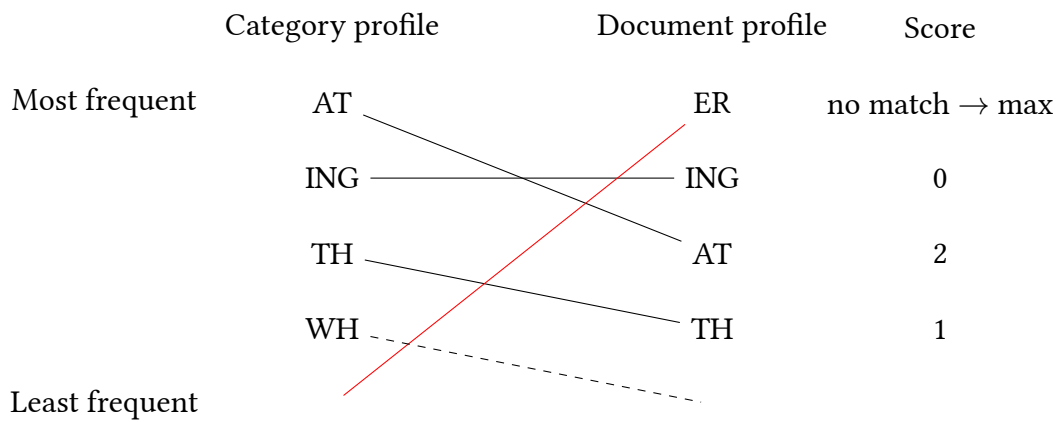

Figure 1: Out-of-place metric

In figure 1, the document profile has 'ER' as most frequent n-gram, at rank 1, followed by 'ING' at rank 2, etc. The category profile does not contain the n-gram 'ER'; in that case, an arbitrary fixed maximum value is assigned. The category profile contains the n-gram 'ING' at rank 2, the same rank as in the document profile; the difference is 0. The category profile contains the n-gram 'AT' at rank 1, while in the document profile, it occurs at rank 3. The absolute difference is 2. The out-of-place metric consists of the sum of all scores thus calculated.

Cavnar and Trenkle (1994) collected 3713 Usenet texts with a cultural theme in different languages. They filtered out non-monolingual texts and texts that had no useful content for language classification. In the end, they had 3478 articles ranging from a single line of text to 50 KB of text.

Their results indicated that length had no significant impact on the classification, contrary to what they thought. Also, they found that training the system with 400 n-grams yielded the best result with a precision of 99.8%.

They also showed that their approach could be used for subject classification of texts in the same language with reasonable precision. This finding indicates that language and domain are linked to a certain degree.

## 2.2 N-Grams and maximum likelihood estimator

Dunning (1994) also uses an n-gram language model for language identification purposes. The program calculates n-grams and their frequencies from the training data and estimates the probability P of a given string using the Maximum Likelihood Estimator (MLE) with Laplace add-one smoothing.More formally:

$$P(w_i|w_1,\ldots,w_{i-1}) = \frac{C(w_1,\ldots,w_i) + 1}{C(w_1,\ldots,w_{i-1}) + |V|} \tag{2}$$

with $C(w_1,\ldots,C_i)$ the number of times the n-gram $w_1,\ldots,w_i$ occurred, $C(w_1,\ldots,C_{i-1})$ the number of times the $(n-1)$-gram $w_1,\ldots,w_{i-1}$ occurred and $|V|$ the size of the vocabulary.

For a string $S$, the string is decomposed into n-grams and the *log* probability $l_k$ is calculated as:

$$l_k = \sum_{w_1,\ldots,w_k \in S} C(w_1,\ldots,w_k) \log P(w_k|w_1,\ldots,w_{k-1}) \tag{3}$$

where $k$ is the order of the n-gram ($k = n$) used.

In order to test the system, Dunning (1994) uses a specially constructed test corpus from a bilingual parallel translated English-Spanish corpus containing English and Spanish texts with 10 texts varying from 1000 to 50000 bytes for the training set and 100 texts varying from 10 to 500 bytes for the test set.

The results indicate that bigram models perform better for shorter strings and less training data while trigram models work better for larger strings and more training data.

Dunning (1994) criticizes Cavnar and Trenkle (1994) for saying that their system would be insensitive to the length of the string to be classified, as the shortest text they classified was about 50 words. The system implemented by Dunning (1994) can classify strings of 10 characters in length "moderately well", while strings of 50 characters or more are classified "very well". Accuracies given vary from 92% for 20 bytes of training data to 99.9% for 500 bytes of text.

## 2.3   Trigrams and short words

Grefenstette (1995) compares trigrams versus short words for language identification. Short words are often function words that are typical for and highly frequent in a given language.

The trigram language guesser was trained on one million characters of text in 10 languages: Danish, Dutch, English, French, German, Italian, Norwegian, Portuguese, Spanish and Swedish. From the same texts, all words with 5 or less characters were counted for the short-word-strategy.

The results indicate that the trigram approach works better for small text fragments of up to 15 words, while for any text longer than 15 words, both methods work equally well with reported accuracies of up to 100% in the 11-15 word range.

## 2.4   N-Grams and clustering

Gao et al. (2001) present a system that augments n-gram language models with clustering techniques. They cluster words by similarity and use these clusters in order to overcome the data sparsity problem.

In traditional cluster-based n-gram models, the probability $P(w_i)$ is defined as the product of the probability of a word given a cluster $c_i$ and the probability of the cluster $c_i$ given the preceding clusters. For a trigram model, the probability $P(w_i)$ of a word $w_i$ is calculated as

$$P(w_i|w_{i-2}w_{i-1}) = P(w_i|c_i) \times P(c_i|c_{i-2}c_{i-1}) \tag{4}$$

The probability of a word given a cluster is calculated as

$$P(w_i|c_i) = \frac{C(w_i)}{C(c_i)} \tag{5}$$

with $C(w_i)$ the count of the word $w_i$ and $C(c_i)$ the count of the cluster $c_i$.

The probability of a cluster given the preceding clusters is calculated using the Maximum Likelihood Estimator

$$P(c_i|c_{i-2}c_{i-1}) = \frac{C(c_{i-2}c_{i-1}c_i)}{C(c_{i-2}c_{i-1})} \qquad (6)$$

Gao et al. (2001) derive from this three ways of using clusters to augment language models: *predictive* clustering (7), *conditional* clustering (8) and *combined* clustering (9).

$$P(w_i|w_{i-2}w_{i-1}) = P(c_i|w_{i-2}w_{i-1}) \times P(w_i|w_{i-2}w_{i-1}c_i) \qquad (7)$$

$$P(w_i|w_{i-2}w_{i-1}) = P(w_i|c_{i-2}c_{i-1}) \qquad (8)$$

$$P(w_i|w_{i-2}w_{i-1}) = P(c_i|c_{i-2}c_{i-1}) \times P(w_i|c_{i-2}c_{i-1}c_i) \qquad (9)$$

Similarly, Dreyfuss et al. (2007) use clustering to cluster words by their context in order to improve trigram language models. In addition to Gao et al. (2001), they also use information about the subject-verb and verb-object relations of the sentence.

They show that their model, using clustering, subject-verb information, verb-object information, and the Porter stemmer outperforms a traditional trigram model.

Carter (1994) clusters training sentences (i.e. the corpus) into subcorpora of similar sentences and calculates separate language model parameters for each subcorpus in order to capture contextual information. In contrast to other works, Carter (1994) clusters sentences instead of single words (compare Pereira et al. (1993) and Ney et al. (1994)). Carter (1994) shows that the subdivision into smaller clusters increases the accuracy of bigram language models, but not trigram models.

## 2.5 Inclusion detection

Beatrice Alex (cf. Alex (2005, 2006, 2007); Alex et al. (2007); Alex and Onysko (2010)) addresses the problem of English inclusions in mainly non-English texts. For the language pair German-English, inclusions are detected using a German and an English lexicon as first resource. If a word is found only in the English lexicon, it is tagged as unambiguously English. If the word is found in neither lexicon, a web search is conducted, restricting the search options to either German or English and counting the number of results. If the German search yields more results, the word is tagged as German, otherwise as English inclusion. If a word is found in both lexicons, a post-processing module resolves the ambiguity.

Alex is mainly concerned with the improvement of parsing results by inclusion detection. For example in (Alex et al., 2007) they report an increase in F-Score of 4.3 by using inclusion detection when parsing a German text with a parser trained on the TIGER corpus (Brants et al., 2002).

## 2.6    Clustering and speech

In the area of clustering and spoken language identification, Yin et al. (2007) present a hierarchical clusterer for spoken language. They cluster 10 languages[1] using prosodic features and *Mel Frequency Cepstral Coefficients* (MFCC). MFCC vectors are a way of representing acoustic signals (Logan et al., 2000). The signal is first divided into smaller 'frames', each frame is passed through the discrete Fourier transform and only the logarithm of the amplitude spectrum is retained (Logan et al., 2000). The spectrum is then projected onto the 'Mel frequency scale', a scale that maps actual pitch to perceived pitch, "as apparently the human auditory system does not perceive pitch in a linear manner" (Logan et al., 2000). Finally, a discrete cosine transform is applied to the spectrum to get the MFCC representations of the original signal (Logan et al., 2000).

Yin et al. (2007) show that their hierarchical clusterer outperforms traditional *Acoustic Gaussian Mixture Model* systems.

As spoken language will not be further investigated in this thesis, I will not dive deeper into the matter at this point.

## 2.7    Monolingual training data

Yamaguchi and Tanaka-Ishii (2012), King and Abney (2013) and Lui et al. (2014) use monolingual training data in order to train a system capable of recognizing the languages in a multilingual text.

Yamaguchi and Tanaka-Ishii (2012) use a dynamic programming approach to segment a text by language. Their test data contains fragments of 40 to 160 characters and achieves F-scores of 0.94 on the relatively 'closed' data set of the Universal Declaration of Human Rights[2] and 0.84 on the more 'open' Wikipedia data set. However, the approach is computationally intensive, not to say prohibitive; while Yamaguchi and Tanaka-Ishii (2012) self-report a processing time of 1 second for an input of 1000 characters, Lui et al. (2014) found that with 44 languages, the approach by Yamaguchi and Tanaka-Ishii (2012) takes almost 24 hours to complete the computation on a 16 core workstation.

King and Abney (2013) use weakly supervised methods to label the languages of words. They consider the task as sequence labeling task. They have limited themselves to bilingual documents with a single language boundary and the task consists in discriminating between English and non-English text. They found that a Conditional Random Field model augmented with Generalized Expectation criteria worked best, yielding accuracies of 88% with as little as 10 words used for training.

---

[1]The authors do not explicitly list the languages clustered, except for two-letter abbreviations which seem to correspond to ISO 639-1. The languages under investigation could have been Vietnamese, German, Farsi, French, Japanese, Spanish, Korean, English, Tamil, and 'ma', though it is impossible to tell.

[2]`http://www.un.org/en/documents/udhr/`

Lui et al. (2014) consider the task as multi-label classification task. They represent a document as an n-gram distribution of byte sequences in a bag-of-words manner. They report F-scores of 0.957 and 0.959. They note that similar languages will pose problems when trying to identify a language, and solve this problem by identifying a set of languages that most probably are correct instead of a single language.

One problem that these approaches all have is that they need to know the languages that will occur in the test data (King and Abney, 2013; Lui et al., 2014).

## 2.8 Predictive suffix trees

Seldin et al. (2001) propose a system for automatic unsupervised language segmentation and protein sequence segmentation. Their system uses Variable Memory Markov (VMM) sources, an alternative to Hidden Markov Models (HMM) implemented as Predictive Suffix Trees (PST).

Whereas HMMs require substantial amounts of training data and a deep understanding of the problem in order to restrict the model architecture, VMMs are simpler and less expressive than HMMs, but have been shown to "solve many applications with notable success" (Begleiter et al., 2004). In contrast to n-gram models that estimate the probability of w as $P(w|N)$ with $N$ the context (typically the $n$ previous words), VMMs can vary $N$ in function of the available context (Begleiter et al., 2004). Thus, they can capture both small and large order dependencies, depending on the training data (Begleiter et al., 2004).

There is no single VMM algorithm, but rather a family of related algorithms. One of these algorithms is called Predictive Suffix Tree (PST) (Ron et al., 1996). A PST is a tree over an alphabet $\Sigma$, with each node either having 0 (leaf nodes) or $|\Sigma|$ children (non-terminal nodes) (Ron et al., 1996). Each node is labeled with the result of the walk from that node up to the root (Ron et al., 1996). Each edge is labeled by a symbol $s \in \Sigma$ and the probability for the next symbol being $s$ (Ron et al., 1996).

By modifying the Predictive Suffix Tree (PST) algorithm using the Minimum Description Length (MDL) principle, Seldin et al. (2001) end up with a non-parametric self-regulating algorithm. The MDL principle avoids overfitting of the model by favoring low complexity over goodness-of-fit (Grünwald, 2007).

They embed the algorithm in a deterministic annealing (DA) procedure to refine the results. Finally, they use the *Blahut-Arimoto* algorithm, a rate-distortion function, until convergence of the system.

For the language segmentation task, they use 150000 letters of text, 30000 from each of the following languages: English, German, French, Italian, transliterated Russian. They used continuous language fragments of approximately 100 letters, yielding a synthetic multilingual text that switches language approximately every two sentences. One important point that they note is that "too short segments do not enable reliable discrimination between different models". Therefore, they disallow switching models after every word.

They report very good results on the language segmentation task (and on the protein segmentation task). After 2000-3000 iterations of the Blahut-Arimoto algorithm, the correct number of languages is identified and the segmentation is accurate up to a few letters.

# 3 Theory

## 3.1 Supervised language model

### 3.1.1 N-Gram models

Among supervised language models, $n$-gram models are very popular (Gao et al., 2001). An $n$-gram is a slice from the original string (Cavnar and Trenkle, 1994). These slices can be contiguous or not. Non-contiguous n-grams are also called *skip-grams* (Guthrie et al., 2006). In skip-grams, an additional parameter $k$ indicates the maximum distance that is allowed between units. In this parlance, contiguous n-grams can be regarded as 0-skip-n-grams (Guthrie et al., 2006).

The following example demonstrates the difference between (traditional) n-grams and skip-grams. Given the following sentence:

$$\text{This is a sample sentence.}$$

We can construct, for example, the following word $k$-skip-$n$-grams:
**(0-skip-)2-grams:** This is, is a, a sample, sample sentence
**2-skip-2-grams:** This is, This a, This sample, is a, is sample, is sentence, a sample, a sentence, sample sentence
**(0-skip-)3-grams:** This is a, is a sample, a sample sentence
**2-skip-3-grams:** This is a, This is sample, This is sentence, This a sample, This a sentence, This sample sentence, is a sample, is a sentence, is sample sentence, a sample sentence

The results for 2-skip-2-grams does not include the skip-gram "This sentence", as the distance in words between these two words is 3, higher than the allowed $k$ of 2. As can be seen from this example, the number of skip-grams is more than two times higher than the number of contiguous n-grams, and this trend continues the more skips are allowed (Guthrie et al., 2006). Skip-grams, unlike n-grams, do not incur the problem of data sparseness with an increase of $n$.

Instead of using words as unit for n-gram decompositions, we can also choose characters. Each word is then decomposed into sequences of $n$ characters. For example, the word

$$\text{model}$$

can be decomposed into the 2-grams: mo, de, el. Often, the word to decompose is padded with start and end tags in order to improve the model (Cavnar and Trenkle, 1994). If we pad the word with <w> and </w>, the 2-gram decomposition yields: <w>m, mo, de, el, l </w>. The use of paddings allows the model to capture details about character distribution with regard to the start and end of words (Cavnar and Trenkle, 1994). For example, in English the letter 'y' occurs more often at the end of words than at the beginning of words, while the letter 'w' occurs mainly at the beginning of words (Taylor, 2015). A non-padding model cannot capture this distinction, while a padding model can.

One advantage of n-gram models is that the decomposition of a string into smaller units reduces the impact of typing errors (Cavnar and Trenkle, 1994). Indeed, a typing error only affects a limited number of units (Cavnar and Trenkle, 1994). Due to this property, n-gram models have been shown to be able to deal well with noisy text (Cavnar and Trenkle, 1994).

### 3.1.2 Formal definition

Traditional n-gram language models predict the next word $w_i$ given the previous words $w_1, \ldots, w_{i-1}$. This prediction uses the conditional probability $P(w_i|w_1, \ldots, w_{i-1})$. Instead of using the entire history $w_1, \ldots, w_{i-1}$, the probability is approximated by using only the $n$ previous words $w_{i-n+1}, \ldots, w_{i-1}$.

$$P(w_i|w_1, \ldots, w_{i-1}) = P(w_i|w_{i-n+1}, \ldots, w_{i-1}) \tag{10}$$

The probability can be estimated using the Maximum Likelihood Estimation (MLE):

$$P(w_i|w_{i-n+1}, \ldots, w_{i-1}) = \frac{C(w_{i-n+1}, \ldots, w_i)}{C(w_{i-n+1}, \ldots, w_{i-1})} \tag{11}$$

Where $C(w_{i-n+1}, \ldots, w_i)$ represents the number of times the $n$-gram sequence $w_{i-n+1}, \ldots, w_i$ occurred in the training corpus and $C(w_{i-n+1}, \ldots, w_{i-1})$ represents the number of times the $(n-1)$-gram sequence $w_{i-n+1}, \ldots, w_{i-1}$ was seen in the training corpus.

### 3.1.3 Smoothing

The problem with MLE is that sequences not seen during training will have a probability of zero. In order to avoid this problem, different smoothing techniques can be used (Chen and Goodman, 1996). The simplest smoothing technique is additive (Laplace) smoothing (Chen and Goodman, 1996). Let $V$ be the vocabulary size (i.e. the total number of unique words in the test corpus). The smoothed probability $P_{Laplace}$ becomes:

$$P_{Laplace}(w_i|w_{i-n+1}, \ldots, w_{i-1}) = \frac{C(w_{i-n+1}, \ldots, w_i) + \lambda}{C(w_{i-n+1}, \ldots, w_{i-1}) + \lambda V} \tag{12}$$

With $\lambda$ the smoothing factor. If we choose $\lambda = 1$, we speak of "add one" smoothing (Jurafsky and Martin, 2000). In practice, $\lambda < 1$ is often chosen (Manning and Schütze, 1999).

An important estimation is the Good-Turing estimation (Chen and Goodman, 1996). While not directly a smoothing method, it estimates the frequency of a given observation with

$$c^* = (c + 1)\frac{N_{c+1}}{N_c} \tag{13}$$

where $c$ is the number of times the observation was made, $N_c$ is the number of times the frequency $c$ was observed and $N_{c+1}$ the frequency of the frequency $c + 1$. Thus, instead of using the actual count $c$, the count is taken to be $c^*$ (Chen and Goodman, 1996).

Another way to avoid assigning probabilities of zero to unseen sequences is by using back-off models. There are linear and non-linear back-off models. In non-linear back-off models, if the original n-gram probability falls below a certain threshold value, the probability is estimated by the next lowest n-gram model. Katz's back-off model (Katz, 1987) for instance calculates probability $P_{bo}$ using the formula:

$$P_{bo} = \begin{cases} d_{w_{i-n+1},\dots,w_i} \frac{C(w_{i-n+1},\dots,w_i)}{C(w_{i-n+1},\dots,w_{i-1})} & \text{if } C(w_{i-n+1},\dots,w_i) > k \\ \alpha_{w_{i-n+1},\dots,w_{i-1}} P_{bo}(w_i|w_{i-n+2},\dots,w_{i-1}) & \text{otherwise} \end{cases} \tag{14}$$

With $d$ and $\alpha$ as smoothing parameters. The parameter $k$ is often chosen $k = 0$. This means that if the probability given a high-order n-gram model is zero, we back off to the next lowest model. For tri-gram models, the formula becomes:

$$P_{bo}(w_i|w_{i-2}, w_{i-1}) = \begin{cases} P(w_i|w_{i-2}, w_{i-1}) & \text{if } C(w_{i-2}, w_{i-1}) > 0 \\ \alpha_1 P(w_i|w_{i-1}) & \text{if } C(w_{i-2}, w_{i-1}) = 0 \text{ and } C(w_{i-1}, w_i) > 0 \\ \alpha_2 P(w_i) & \text{otherwise} \end{cases} \tag{15}$$

In contrast, linear back-off models use an interpolated probability estimate by combining multiple probability estimates and weighting each estimate. The probability $P_{LI}$ for a tri-gram model is:

$$P_{LI}(w_i|w_{i-2}, w_{i-1}) = \lambda_3 P(w_i|w_{i-2}, w_{i-1}) + \lambda_2 P(w_i|w_{i-1}) + \lambda_1 P(w_i) \tag{16}$$

with $\sum \lambda_i = 1$

## 3.2 Unsupervised clustering

Clustering consists in the grouping of objects based on their mutual similarity (Biemann, 2006). Objects to be clustered are typically represented as feature vectors (Biemann, 2006); from the original objects, a feature representation is calculated and used for further processing.

Clustering can be partitional or hierarchical (Yin et al., 2007). Partitional clustering divides the initial objects into separate groups in one step, whereas hierarchical clustering builds a hierarchy of objects by first grouping the most similar objects together and then clustering the next level hierarchy with regard to the existing clusters (Yin et al., 2007).

The clustering algorithm uses a *distance* metric to measure the distance between the feature vectors of objects (Biemann, 2006). The distance metric defines the similarity of objects based on the feature space in which the objects are represented (Jain et al., 1999). There are different metrics available. A frequently chosen metric is the cosine similarity that calculates the distance between two vectors, i.e. the angle between them (Biemann, 2006).

In order for a clustering algorithm to work, features that represent the object to be clustered have to be defined (Jain et al., 1999). Features can be quantitative (e.g. word length) or qualitative (e.g. word starts with a capital letter) (Jain et al., 1999).

Most clustering algorithms, e.g. $k$-means, need the number of clusters to generate (Jain et al., 1999). The question how to best choose this key number has been addressed in-depth by Dubes (1987).

Clustering can be *soft* or *hard*. When hard-clustering, an object can belong to one class only, while in soft-clustering, an object can belong to one or more classes, sometimes with different probabilities (Jain et al., 1999).

### 3.3 Weakly supervised language model induction

The main idea behind language model induction is that by inducing language models from the text itself, the models are highly specialized but the approach is generally more flexible since genre or text specific issues do not arise.

This approach is similar in character to the work by Seldin et al. (2001) in that the text itself is used as data set. However, the realization differs greatly. Whereas Seldin et al. (2001) use predictive suffix trees, I use n-gram language models.

The intuition is to learn the language models from the text itself, in an iterative manner. Suppose we have a document as follows where $w_i$ represents the word at position $i$ in the text. Suppose the text contains two languages, marked in red and blue.

$$w_1 \quad w_2 \quad w_3 \quad w_4 \quad w_5 \quad w_6 \quad w_7 \quad w_8 \quad w_9 \quad w_{10} \quad \cdots$$

Figure 2: Simple text illustration

We take the first word and create a language model $m_1$ from that word.

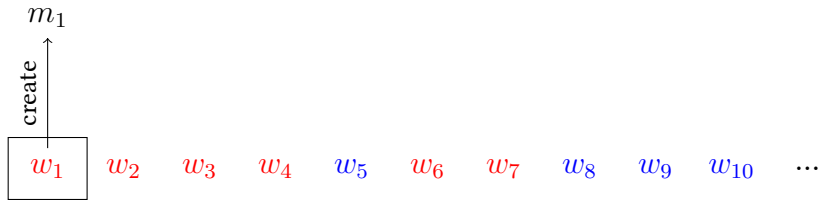

Figure 3: Initial model creation

We then evaluate the second word using the first language model. If the language model score is high enough, we update the language model with the second word.

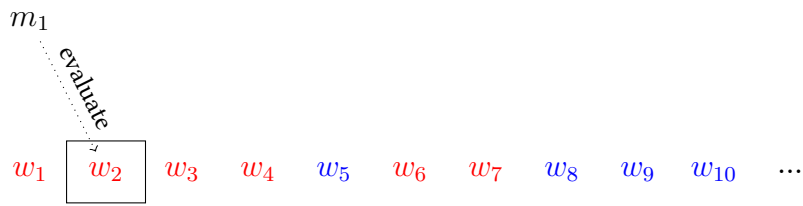

Figure 4: Initial model evaluation

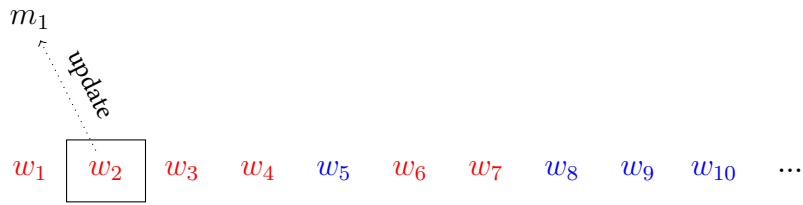

Figure 5: Model update

If the score is below a certain threshold, the existing language model does not model the word well enough and a new model is created.

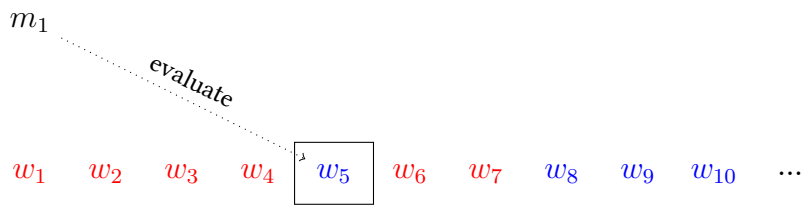

Figure 6: Evaluation

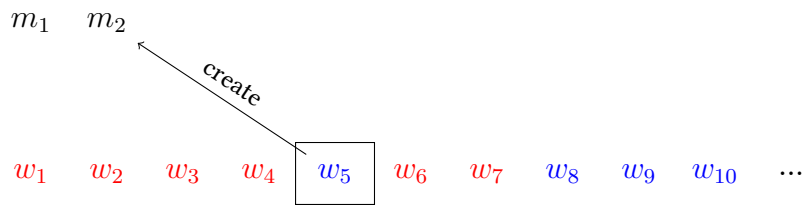

Figure 7: New model creation

When there is more than one language model, each word is evaluated by every language model, and the highest scoring model is updated, or a new model is created if no language model models the word well enough.

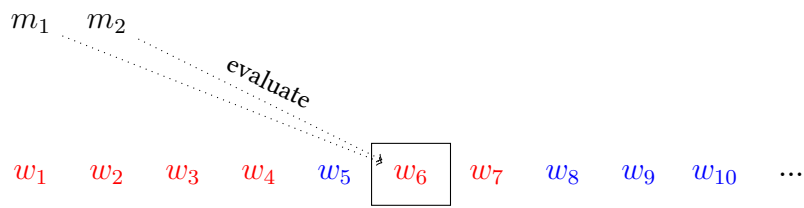

Figure 8: Multiple model evaluation

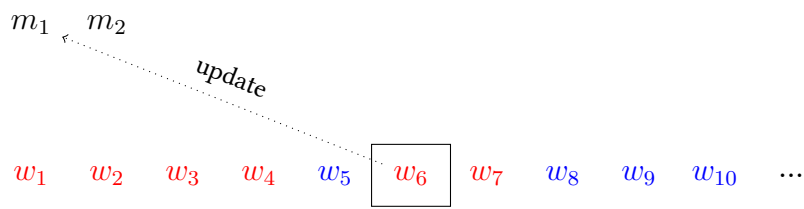

Figure 9: Updating relevant model

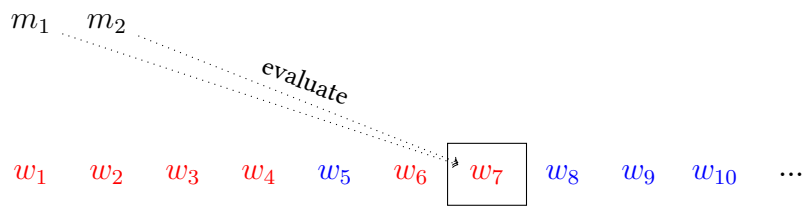

Figure 10: Multiple model evaluation 2

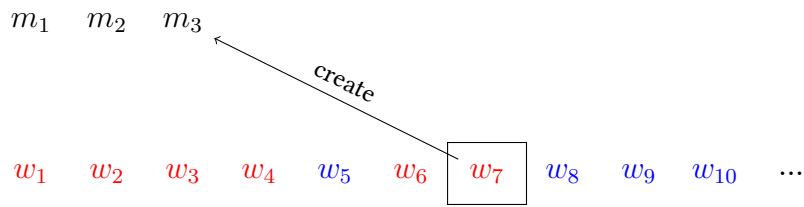

Figure 11: New model creation 2

The last example shows that it is not necessarily the case that exactly one language model is created per language; it often is the case that many language models are created for one language.

At the beginning, the models are not very reliable, as they only have a few words as basis, but the more text is analyzed, the more reliable the models become.

However, the approach is problematic in that the text structure itself influences the language models created. If the text starts with a foreign language inclusion, as illustrated in figure 12, the initial model might be too frail to recognize the following words as being a different language, updating the first model with the second and third word and so on. Thus, the approach would fail at recognizing the foreign language inclusion.

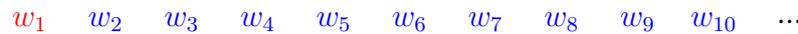

Figure 12: Problematic text sample

If we were to start from the end of the text and work towards the beginning, the probability of having a relatively robust language model for the 'blue' language would be high, and so, it would theoretically be easier to recognize the first word as not being 'blue'.

Therefore, one induction step involves one forward generation and one backwards generation. This yields two sets, the set of models from the forward generation $F = \{f_1, f_2, \ldots, f_n\}$ and the set from the backwards generation $B = \{b_1, b_2, \ldots, b_m\}$. Then, from the two sets of models, the most similar models are selected. For this, every model from $F$ is compared to every model from $B$, as figure 13 shows. The most similar models are then merged, as illustrated in figure 14. Indeed, if both the forward and backwards generation yielded a similar language model, it is probable that the model is correct.

Even so, both forward and backwards generation can not guarantee ideal results, there is the option to run the generation from a random position. This random induction picks a random position in the text and runs one induction step from that position, meaning one forward and one backwards generation. Finally, the most similar models are merged as for the general generation.

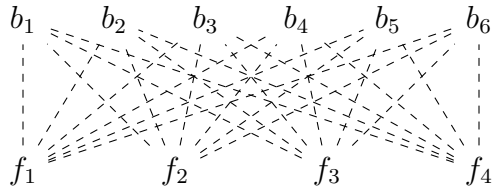

Figure 13: Finding the most similar models

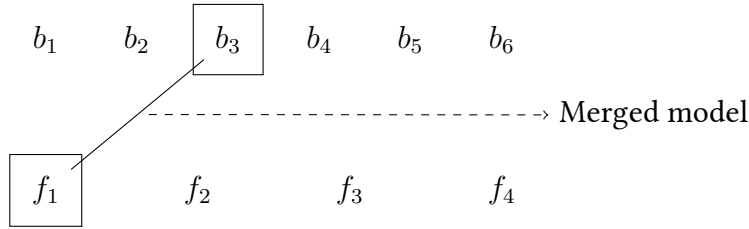

Figure 14: Merging most similar models

This only yields one probable language model, therefore the induction is repeated with the difference that all probable models are taken into consideration as well. For each word, if a probable model models the word well enough, no new model is created, otherwise a new model is created.

At the end of the induction loop, the set of probable models $P$ is examined. As long as there are two models that have a similarity score below a certain threshold, the two most similar models are merged.

Finally, after the language models have been induced, another pass is made over the text and each word is assigned to the language model which yields the highest score for that word, resulting in a word-to-model assignment as illustrated in figure 15.

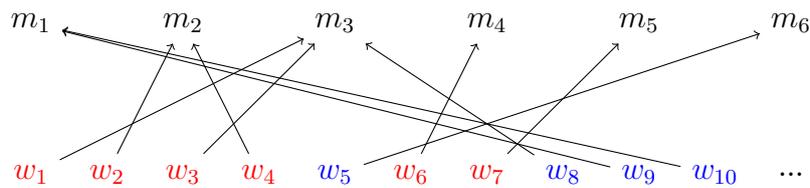

Figure 15: Word-Model assignment

I have made the approach parametric with parameters being:

- Induction iterations: Number of induction iterations

- Random iterations: Number of random iterations

- Forward/Backwards threshold: Threshold for forward/backwards merging

- Silver threshold: Threshold for $P$ model merging

These parameters can be adapted, in the hope that some parameter configurations will work better on certain data sets than other configurations. Since the approach has parameters that have to be learned from a development set, the approach is said to be weakly supervised; the development set is not used to train any language specifics, only for the estimation of the parameters of the approach.

# 4   Experimental setup

In this chapter I present experiments done using the approaches delineated in the previous section in order to find out whether there are approaches that work better on certain types of text.

The central hypothesis is that unsupervised language segmentation approaches are more successful on difficult data. Difficult data is data for which there is not enough data to train a language model or data which contains a lot of non-standard language such as abbreviations.

First, I present the data used to test the language segmentation systems and elaborate on the different aspects that had to be considered for the data compilation.

I then present two supervised language segmentation experiments using n-gram language models and Textcat.

For unsupervised language segmentation, I will first present experiments using clustering algorithms before presenting experiments using language model induction.

## 4.1   Data

In order to test the different language segmentation approaches, I compiled different sets of test data. As I want to focus on short texts, most texts from the test corpus are rather small, sometimes consisting of only one sentence. However, in order to test the general applicability of the approach, the test corpus also contains larger text samples.

The test corpus can be subdivided into different sub-corpora:

- Latin-based: Texts consisting of languages using Latin-based scripts, such as German, English, Finnish or Italian

- Mixed script: Texts consisting of languages using Latin-based scripts and languages using non-Latin-based scripts

- Twitter data: Short texts taken from Twitter

- Pali dictionary data: Unstructured texts containing many different language inclusions such as Vedic Sanskrit, Sanskrit, Indogermanic reconstructions, Old Bulgarian, Lithuanian, Greek, Latin, Old Irish, many abbreviations and references to text passages

As every outcome has to be manually checked, the test corpus is rather small. Every category consists of five texts. Each texts consists of two or three languages with the exception of the Pali dictionary data that often contains inclusions from many different languages in the etymological explanations.

For each text, I also created a gold standard version with the expected clusters. In some cases it is not clear how to cluster certain objects. In that case, I use a clustering that makes sense to me, but this need not mean that it is the correct or only possible clustering.

For the parameter estimation of the language model induction approach, I also compiled a set of development data. All texts can be found in the appendix under 8.1 and 8.2.

## 4.2 Supervised language model

### 4.2.1 Implementation

For the supervised language segmentation method, I implemented an n-gram language model as described by Dunning (1994). The n-gram language model is implemented as a character trigram model with non-linear back-off to bigram and unigram models. The conditional probability $P$ is calculated using the formula:

$$P(w_i|w_{i-2}, w_{i-1}) = \begin{cases} \alpha_1 \frac{C(w_{i-2}, w_{i-1}, w_i)}{C(w_{i-2}, w_{i-1})} & \text{if } C(w_{i-2}, w_{i-1}, w_i) > 0 \\ \alpha_2 \frac{C(w_{i-1}, w_i)}{C(w_{i-1})} & \text{if } C(w_{i-1}, w_i) > 0 \\ \alpha_3 \frac{C(w_i)}{V} & \text{if } C(w_i) > 0 \\ \alpha_4 \frac{1}{V+W+X} & \text{otherwise} \end{cases} \tag{17}$$

with $\alpha_1 = 0.7$, $\alpha_2 = 0.2$, $\alpha_3 = 0.09$, $\alpha_4 = 0.01$, $V$ the number of unigrams, $W$ the number of bigrams and $X$ the number of trigrams.

Each word is padded by two different start symbols and two different end symbols. The joint probability for a word $w$ of length $n$ is calculated as

$$P(w) = \frac{1}{\sum_{i=2}^{n} |\log P(w_i|w_{i-2}, w_{i-1})|} \tag{18}$$

In the denominator, I use the *log probability* instead of the probability to increase numerical stability. Indeed, multiplying very small numbers can lead to the result being approximated as zero by the computer when the numbers become too small to be represented as normalized number (Goldberg, 1991). Using the sum of logarithms avoids this problem and is less computationally expensive (Bürgisser et al., 1997).

As the logarithm of a number approaching zero tends to infinity, rare observations get a higher score than frequent observations. As such, the denominator can be seen as a scale of *rarity*, with a higher score corresponding to a rarer word. By taking the inverse of this scale, we get a score corresponding to the "commonness" ($\approx$ frequency) of a word.

### 4.2.2 Training phase

First, models are trained on training data in the relevant languages. I have not included the languages from the Pali dictionary data, as there are too many different languages and there are typically only small inclusions of different languages in a dictionary entry; as such, it would not have made sense to train a language model just to recognize a single word. Another reason for not using the Pali dictionary data languages is that sometimes it is not possible to find data for a language, e.g. Old Bulgarian or reconstructed Indogermanic. In some cases, it would have been conceivable to train models on similar languages, but again, the effort of training a model is disproportionately high compared to the (uncertain) result of recognizing a single inclusion. Instead, an additional catch-all language model is used to capture words that do not seem to belong to a trained model.

| Language | Size in MB |
| --- | --- |
| Amharic | 9 |
| Arabic | 747 |
| Chinese | 1005 |
| English | 2097 |
| Finnish | 570 |
| French | 2097 |
| German | 2097 |
| Greek | 464 |
| Italian | 2097 |
| Polish | 2097 |
| Russian | 2097 |
| Spanish | 2097 |
| Turkish | 386 |
| Ukrainian | 1456 |

Table 1: Training data: Size

The training data consists of *Wikipedia dumps* from the months June and July 2015; a dump is a copy of the whole encyclopedia for a given language. Due to the difference in size of the Wikipedia of the different languages, I choose the full dump for languages with less than 3 GB of compressed data and limited the amount of data to maximally 3 GB of compressed data.

The Wikipedia data was processed using the Wikipedia Extractor[3] version 2.8 in order to extract the textual content from the article pages. Indeed, the Wikipedia pages are written using the *MediaWiki Markup Language*[4]. While this markup is useful for meta-data annotation and cross-referencing, the encoded information is superfluous for language model training and has to be removed before training a model on the data. Table 1 shows the size of the training data per language after text extraction.

---

[3] `http://medialab.di.unipi.it/wiki/Wikipedia_Extractor`

[4] `https://www.mediawiki.org/wiki/Help:Formatting`

As the test data only contains transliterated Amharic text, the Wikipedia data, written in the Ge'ez script, had to be transliterated. The text was transliterated according to the EAE transliteration scheme by the Encyclopaedia Aethiopica.

As the test data contains transliterated Greek, the Greek data was used once as-is and once transliterated according to the ELOT (Hellenic Organization for Standardization) transliteration scheme for Modern monotonic Greek.

It should be borne in mind that the training data influences the quality and accuracy of the model. Furthermore, a model might work well on certain text types and less well on other text types. It is not possible to train a perfect, universal model.

### 4.2.3  Application of the approach

In the second step, an input text is segmented into words. Then, each word is evaluated by each language model and the model with the highest score is assigned as the word's language model.

The approach taken consists in classifying words as either belonging to a trained language model or to the additional, catch-all model *other*, which simply means that the word could not be assigned to a trained model class.

### 4.2.4  Textcat and language segmentation

I also tested how well Textcat is suited to the task of language segmentation. The approach is similar to the n-gram approach, with the exception that I do not train any models and rely on Textcat's classifier for language prediction.

In the first step, an input text is segmented into words. Then, each word is passed to Textcat and the guess made by Textcat is taken as the word's language.

## 4.3  Unsupervised clustering

In order to test the efficiency of clustering algorithms on the task of language segmentation, I looked at various algorithms readily available through *WEKA*, "a collection of machine learning algorithms for data mining tasks" by the University of Waikato in New Zealand (Hall et al., 2009) and the Environment for Developing KDD-Applications Supported by Index-Structures (ELKI), "an open source data mining software [...] with an emphasis on unsupervised methods in cluster analysis and outlier detection" by the Ludwig-Maximilians-Universität München (Achtert et al., 2013). I also looked at *JavaML*, "a collection of machine learning and data mining algorithms" (Abeel et al., 2009), in order to integrate clusterers into my own code framework. JavaML offers different clustering algorithms and also offers access to WEKA's clustering algorithms. In contrast to WEKA and ELKI, which can be used in stand-alone mode, JavaML is meant to be integrated into bigger programs and provides an application programming interface (API) that allows the provided algorithms to be accessed in a programmatic way, i.e. from inside a program.

### 4.3.1 Preprocessing

However, in order for the clustering algorithms to work, the document to segment has to be preprocessed in a number of ways, as shown in figure 16.

Figure 16: Clustering preprocessor

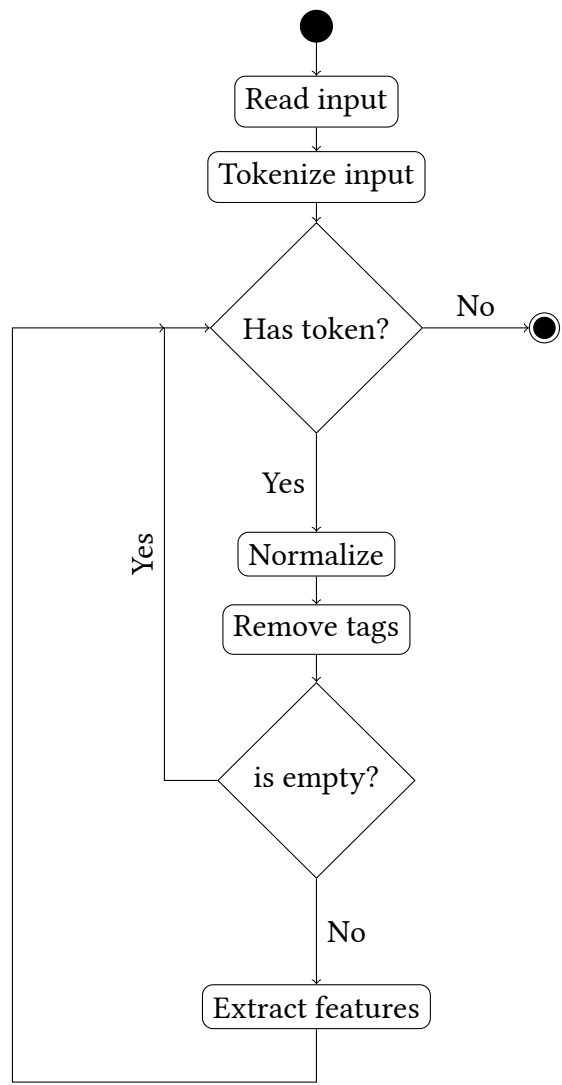

First of all, the document has to be read in by the program. This step is straightforward.

The document then has to be tokenized. Tokenization is not trivial and depends on the definition of a 'word'. For this task I have used a whitespace tokenizer that defines a word as a continuous sequence of character literals separated by one or more whitespace characters. While it can be objected that for scripts that don't use whitespace to separate words, such as Chinese, tokenization fails, this is not too big a concern. Indeed, if a continuous block of Chinese characters is treated as one word, it is likely to be clustered separately due to the different in "word" length and the different character set. If, however, a document contains two scripts that do not separate words by whitespace, the approach totally fails. It is beyond the scope of this thesis, and possibly of any thesis, to implement a universal tokenizer that works regardless of language without prior knowledge about the languages at hand.

Each token is then *normalized*. Normalization of a non-Latin-based input (e.g. Arabic or Cyrillic script) returns the input without modification. Otherwise, the following modifications are made, if applicable:

- remove leading and trailing whitespace

- remove punctuation

- remove control characters

Control characters are defined as the set

( [ ] ) \

Punctuation is defined as the set

. , " ' : ; ! ? —

The token is then stripped of XML-like tags, if applicable. The following example illustrates this step. Let us assume we have the following token:

< word  id = " 1 "  lemma = " go " > g o e s < / word >

The token is replaced by the text content of the node, thus the resulting token is 'goes'.

If, after all these modifications, the token corresponds to the empty string, we continue with the next token. Otherwise, the token is passed on to the feature extraction module. The algorithm terminates when all tokens have been consumed.

### 4.3.2 Defining features

The final step consists in defining features by which to cluster and implementing feature extractors that build the feature vectors from the input. Since the features are to be language independent, using features such as 'occurs in an English lexicon' cannot be used. The following features were devised:

1. word length: the length of the word in characters

2. X tail bigrams: bigrams calculated from the end of the word

3. Y tail trigrams: trigrams calculated from the end of the word

4. X first bigrams: bigrams calculcated from the beginning of the word

5. Y first trigrams: trigrams calculated from the beginning of the word

6. latin basic: is the word latin basic?

7. latin extended: is the word latin extended?

8. capitalized: is the word capitalized?

9. contains non-word: does the word contain a non-word?

10. is non-word: is the word a non-word?

11. number of latin letters: number of latin letters

12. number of non-latin letters: number of non-latin letters

13. vowel ratio: number of vowels divided by the word length

14. basic latin letter ratio: number of latin letters divided by the word length

15. max consonant cluster: the longest consonant cluster size in characters

16. is digit: is the word a digit?

17. is ideographic: is the word ideographic?

18. directionality: what directionality does first character of the word have?

19. is BMP codepoint: does the word contain non-BMP characters?

20. general type: what is the general type of the first character of the word?

The last two features are based on the Java Character class. This class provides methods to check for specific implementation-based properties of characters.

While most features are rather self-explanatory, a few require further explanation. For the n-grams, the number of n-grams is restricted so as to keep the resulting vectors the same size. This is important because the clustering algorithm considers one data column as one feature, and having vectors of different length would disrupt this precondition. Implementing the comparison of vectors of different lengths, or rather or vectors containing vectors as features would have been possible, but rather time-consuming. If a word is too short to generate the required number of n-grams, only the possible n-grams are generated and all other positions filled with 0.

The 'latin' features check whether the word consists only of the basic latin letters A-Z and a-z ('basic') while the 'extended' feature also covers letters derived from the latin letters (e.g. ë, ç, m̧, ñ).

Non-words are defined as anything not consisting of letters, such as punctuation marks or digits.

Directionality indicates which direction a character should be written. While the actual list is much more exhaustive, this property basically indicates whether the character is written from left to right or from right to left. [5]

BMP stands for Basic Multilingual Plane and refers to an encoding unit known as *plane*, which consists of $2^{16} = 65536$ codepoints (i.e. encoding slots for characters) (The Unicode Consortium, 2014). The BMP is the first plane, covering the codepoints U+0000 to U+FFFF (The Unicode Consortium, 2014). While it is not important to understand the technical details fully, it is interesting to note that most characters are covered by the BMP, including Chinese, Japanese and Korean characters (The Unicode Consortium, 2014). The next plane, called Supplementary Multilingual Plane or Plane 1 contains historic scripts such as Egyptian hieroglyphs and cuneiform scripts, but also musical notation, game symbols and various other scripts and symbols (The Unicode Consortium, 2014). There are 17 planes in total (The Unicode Consortium, 2014).

The last feature in the list, General Type is also an implementation-related property. Type can be, for example[5], END_PUNCTUATION, LETTER_NUMBER or MATH_SYMBOL. These constants are represented as numbers internally, which are taken as feature for the clustering algorithm.

### 4.3.3 Mapping features to a common scale

As JavaML requires numerical features, all features were mapped to numerical scales:

- Binary features were mapped to 0 (false) and 1 (true)

- Ternary features were mapped to 0 (false), 1 (true) and 99 (not applicable)

- Numerical features were represented as themselves, either as whole numbers (e.g. word length) or as floating point numbers (e.g. vowel ratio)

- Java specific features (18,20) take the underlying numerical value as feature

- N-grams were encoded numerically using algorithm 1

---

[5]The full list can be found under the documentation of the Java Character class http://docs.oracle.com/javase/7/docs/api/java/lang/Character.html

---

**Algorithm 1** N-gram numerical encoding

---

1: **function** ENCODE($word$)
2:     $sum \leftarrow 0$
3:     **for** character in word **do**
4:         $value \leftarrow$ code-point of character
5:         $sum \leftarrow sum + value$
6:     **end for**
7:     **return** $sum$
8: **end function**

---

While algorithm 1 does not encode n-grams in an unambiguous way ("en" and "ne" are both encoded as 211), it provides a sufficiently good encoding.

### 4.3.4   The problem of unambiguous encoding

I have tried using unambiguous encodings. The main problem with unambiguous encoding is that the notion of "distance" is distorted. The idea behind the unambiguous encoding is that each "word" (i.e. string of characters) is encoded numerically so that no two "words" are represented as the same number. Besides the encoding of each separate character, the position of the character inside the string also has to be encoded. A possible encoding $e$ for a string $w_1 w_2 w_3$ could be

$$e_{w_1 w_2 w_3} = n(w_1) + x * n(w_2) + y * n(w_3) \tag{19}$$

with $w_i$ the character of the string at position $i$, $n(w_i)$ the numerical encoding of the character $w_i$ and $x$ and $y$ parameters. If $|A|$ is the alphabet size of the alphabet $A$ in which the word is encoded, the following constraints must be true for the encoding to be unambiguous:

$$x \geq |A| \tag{20}$$

$$y \geq |A|^2 \tag{21}$$

If we take for example the English alphabet with 26 lowercase and 26 uppercase letters, not counting punctuation, digits and other characters, it has to be true that $x \geq 52$ and $y \geq 2704$. The problem is that we cannot know in advance what size the alphabet will be. If we have English and German texts, the size can be estimated around 60. However, if we have English, Russian and Arabic text, the size drastically increases. We could choose any two very big numbers, but if we want to guarantee our encoding to be unambiguous, we run the risk of ending up with numbers too big to be represented efficiently.

In this encoding scheme, distance is skewed: changes to the first character result in linear distance. 'man' and 'nan' have a distance of 1, because 'm' and 'n' have a distance of 1. 'man' and 'lan' have a distance of 2, etc. Changes to the second character are multiplied by $x$. 'man' and 'men' have a distance of $x * (distance(a, e)) = 4 * x$. Changes to the third character are scaled by $y$. For any sufficiently big $x$ and $y$, the distances are too skewed to be used for automatic cluster analysis. Let us consider the following example with only two characters for simplicity. For this example, let us assume $x = 1373$.

|      | na   | ma   | ne   | me   |
|------|------|------|------|------|
| na   | 0    | 1    | 5492 | 5491 |
| ma   | 1    | 0    | 5493 | 5492 |
| ne   | 5492 | 5493 | 0    | 1    |
| me   | 5491 | 5492 | 1    | 0    |

Table 2: Unambiguous encoding: distances

It should be apparent from table 2 that the notion of "distance" is distorted. In comparison, table 3 shows the encoding achieved with algorithm 1.

|      | na | ma | ne | me |
|------|----|----|----|----|
| na   | 0  | 1  | 4  | 3  |
| ma   | 1  | 0  | 5  | 4  |
| ne   | 4  | 5  | 0  | 1  |
| me   | 3  | 4  | 1  | 0  |

Table 3: Simplified encoding: distances

While this encoding is not unambiguous, it is considered sufficiently good for our purposes.

### 4.3.5 The clusterer

Most clustering algorithms such as $k$-means need to be passed the number of clusters to generate. As we want to work as flexibly as possible, I ignored all algorithms that need the number of clusters before clustering. In contrast, the $x$-means algorithm (Pelleg and Moore, 2000) estimates the number of clusters to generate itself. This algorithm has been chosen to perform the language clustering tasks.

While WEKA and ELKI offer a graphical user interface and various graphical representations of the results, the output is not easily interpretable. Indeed, we can get a visualization of a clustering operation as shown in figures 17 (WEKA) and 18 (ELKI). However, all data points have to be manually checked by either clicking each point in order to get additional information about that data point (WEKA) or by hovering over the data points after having selected the Object Label Tooltip option (ELKI). Figure 18 shows the information for the lowest orange rectangle data point in the ELKI visualization.

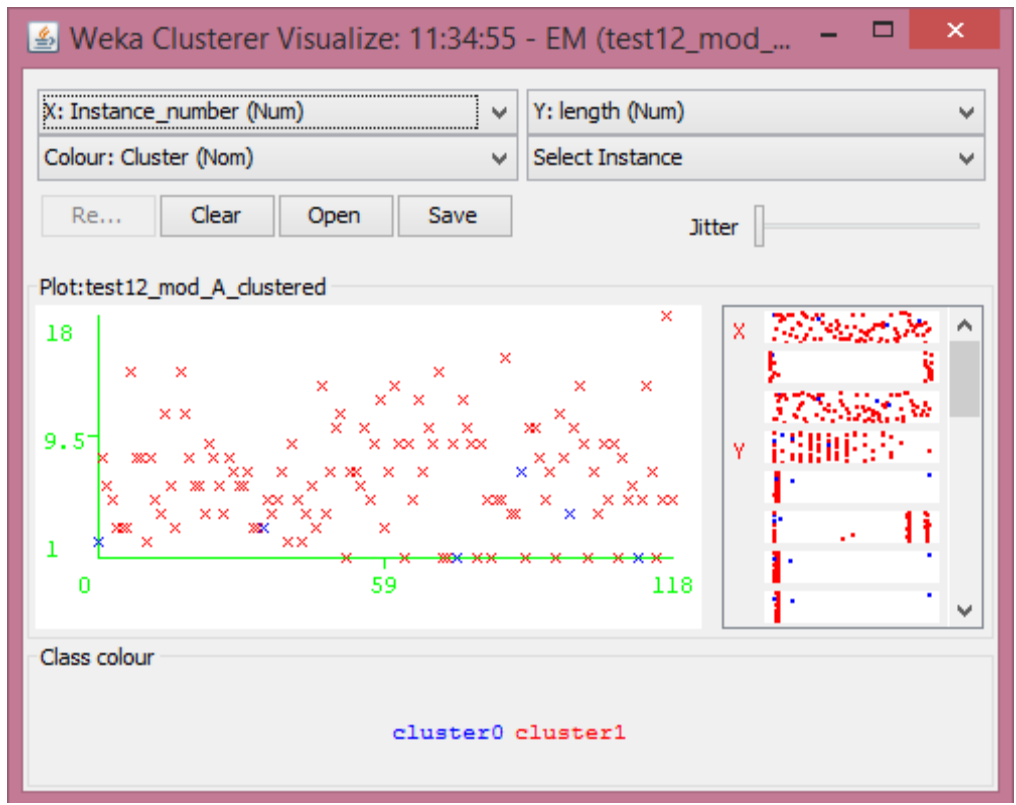

Figure 17: WEKA: Cluster visualization

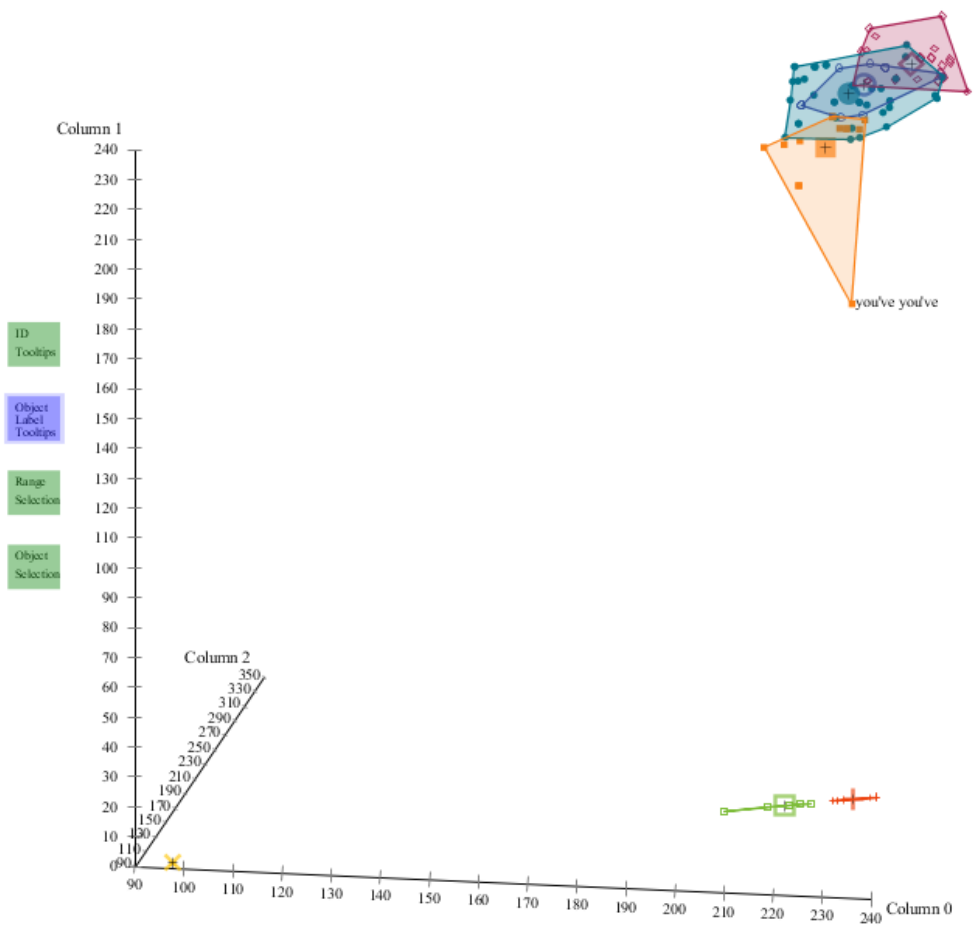

Figure 18: ELKI: Cluster visualization

Therefore, I have decided to embed the $x$-means clustering algorithm into a custom framework. Originally part of the WEKA algorithms, the $x$-means algorithm has been integrated into a Java program via the JavaML library. The framework takes an input file, constructs the aforementioned feature vectors from the input, performs normalization, passes the calculated feature vectors to the clustering algorithm and displays the results in a text-based easily interpretable manner.

Preliminary analyses have shown that the first clustering result often is not discriminating enough. Hence, I perform a first clustering analysis, followed by a second clustering analysis on the clusters obtained from the first analysis.

### 4.3.6 Evaluating clusterings

The clustering results are evaluated using four common similarity measures used in evaluating the accuracy of clustering algorithms. These methods are based on counting pairs (Wagner and Wagner, 2007).

Let us consider the clustering $C = \{C_1, \ldots, C_k\}$. $C$ is a set of non-empty disjoint clusters $C_1, \ldots, C_k$. Let us consider the reference clustering $C' = \{C_1, \ldots, C_l\}$. We define the following sets.

- $S_{11}$: set of pairs that are in the same cluster in $C$ and $C'$

- $S_{00}$: set of pairs that are in different clusters in $C$ and $C'$

- $S_{10}$: set of pairs that are in the same cluster in $C$ and in different clusters in $C'$

- $S_{01}$: set of pairs that are in different clusters in $C$ and in the same cluster in $C'$

Let $n_{ij} = |S_{ij}|$, with $i, j \in \{0, 1\}$ be the size of a given set $S_{ij}$.
The *Rand Index* is defined as

$$RI = \frac{n_{11} + n_{00}}{n_{11} + n_{10} + n_{01} + n_{00}} \tag{22}$$

The Rand Index measures the accuracy of the clustering given a reference partition (Wagner and Wagner, 2007). However, it is criticized for being highly dependent on the number of clusters (Wagner and Wagner, 2007).

The *Jaccard Index* measures the similarity of sets. It is similar to the Rand Index, but it disregards $S_{00}$, the set of pairs that are clustered into different clusters in $C$ and $C'$ (Wagner and Wagner, 2007). It is calculated as

$$J = \frac{n_{11}}{n_{11} + n_{10} + n_{01}} \tag{23}$$

The *Fowlkes-Mallows Index* measures precision. It is calculated as

$$FM = \frac{n_{11}}{\sqrt{(n_{11} + n_{10})(n_{11} + n_{01})}} \tag{24}$$

The Fowlkes-Mallows Index has the undesired property of yielding high values when the number of clusters is small (Wagner and Wagner, 2007).

Finally, I will indicate the F-Score. According to Manning et al. (2008), in the context of clustering evaluation the *F(β) score* is defined as

$$F(\beta) = \frac{(\beta^2 + 1) * P * R}{(\beta^2)P + R} \tag{25}$$

with precision $P$ and recall $R$ defined as

$$P = \frac{n_{11}}{n_{11} + n_{10}} \tag{26}$$

$$R = \frac{n_{11}}{n_{11} + n_{01}} \tag{27}$$

By varying $\beta$, it is possible to give more weight to either precision ($\beta < 0$) or recall ($\beta > 1$) (Manning et al., 2008). As I value recall higher than precision, I will indicate F1 ($\beta = 1$) and F5 ($\beta = 5$) scores. Indeed, I want to penalize the algorithm for clustering together pairs that are separate in the gold standard while not penalizing the algorithm for splitting pairs that are together in the gold standard.

All measures of similarity fall between $[0, 1]$ with $0$ being most dissimilar and $1$ being identical. As there is no ultimate measure and all measures of similarity have their drawbacks (Wagner and Wagner, 2007), all measures will be indicated in the results section.

## 4.4  Weakly supervised language model induction

The language model induction approach works in two stages. In the first stage, n-gram language models are induced from the text. In the second stage, the text is mapped to the induced models. The algorithm for the language model induction is as follows:

---

**Algorithm 2** Model induction

---

1:  CREATEINITIALMODEL
2:  **for** word in words **do**
3:      $modelAndScore \leftarrow$ MAXMODELSCORE(word)
4:      $score \leftarrow modelAndScore.score$
5:      **if** $score < threshold$ **then**
6:          $model \leftarrow$ CREATEMODEL(word)
7:          $models.add(model)$
8:      **else**
9:          $maxModel \leftarrow modelAndScore.model$
10:          $maxModel.update(word)$
11:      **end if**
12: **end for**

---

First of all, an initial language model is created. For each word, the maximum model and maximum score is calculated. These values correspond to the language model that yielded the highest probability for the word in question, and the associated probability. If the score falls below a threshold $t$ (i.e. none of the existing language models model the word well enough), a new language model is created on the basis of the word and added to the list of language models. Otherwise, the top scoring language model is updated with the word in question.

As the text structure itself influences the quality of the induced models, the language model induction is run $i$ times ($i \leqslant 1$), with one iteration consisting of two induction steps, once forward and once backward, and $j$ times from a random position ($j \leqslant 0$). The initial model creation thus either picks the first word of the text (as shown in algorithm 3 line 2), or the last word of the text, or a random word.

---

**Algorithm 3** Initial model creation

---

1: **function** CREATEINITIALMODEL
2:     $word \leftarrow words.first$
3:     $model \leftarrow createModel(word)$
4:     $models.add(model)$
5: **end function**

---

**Algorithm 4** Max model and max score

---

1: **function** MAXMODELSCORE($word$)
2:     $maxScore \leftarrow 0$
3:     $maxModel \leftarrow none$
4:     **for** model in models **do**
5:         $score \leftarrow model.probability(word)$
6:         **if** $score > maxScore$ **then**
7:             $maxScore \leftarrow score$
8:             $maxModel \leftarrow model$
9:         **end if**
10:     **end for**
11:     **return** $maxModel, maxScore$
12: **end function**

---

Algorithm 4 returns both the max model and the max score wrapped as a custom object. The individual values can then be read as necessary.

After the models have been induced, the most similar models are merged based on distributional similarity. Distributional similarity is calculated as explained below. This merging step only merges one model from the forward induction group with one model from the backward induction group. The resulting model is added to the set of *probable* (*"silver"*) models.

Merging is performed according to algorithm 5. The merging algorithm only retains the common set of unigrams from both models, and all resulting bi- and trigrams, excluding any bi- and trigrams that contain character that occur only in one of the models. The values for the resulting language model are calculated according to one of four different merge modes.

The merge modes are:

**Algorithm 5** Model merger

---

1: **function** MERGE($model_1$, $model_2$, $mode$)
2:     $merged \leftarrow \emptyset$
3:     **for** unigram $u_1$ in $model_1$.unigrams **do**
4:         **for** unigram $u_2$ in $model_2$.unigrams **do**
5:             **if** $u_1 = u_2$ **then**
6:                 $v_1 \leftarrow f(u_1)$                                 ▷ $f(u_1)$ is the frequency of $u_1$
7:                 $v_2 \leftarrow f(u_2)$
8:                 $value \leftarrow mode(v_1, v_2)$
9:                 $unigram \leftarrow u_1$                          ▷ or $u_2$, since both are equal
10:                $merged \leftarrow (unigram, value)$
11:             **else**
12:                 $exclude \leftarrow u_1$
13:                 $exclude \leftarrow u_2$
14:             **end if**
15:         **end for**
16:     **end for**
17:     **for all** bigrams $b$ in $model_1$ and $model_2$ **do**
18:         **if** not $exclude$ contains any char in $b$ **then**
19:             $v_1 \leftarrow f(b, model_1)$ or $0$                  ▷ frequency of $b$ in $model_1$
20:                                               ▷ or $0$ if it does not exist
21:             $v_2 \leftarrow f(b, model_2)$ or $0$
22:             $value \leftarrow mode(v_1, v_2)$
23:             $merged \leftarrow (b, value)$
24:         **end if**
25:     **end for**
26:     **for all** trigrams $t$ in $model_1$ and $model_2$ **do**
27:         **if** not $exclude$ contains any char in $t$ **then**
28:             $v_1 \leftarrow f(t, model_1)$ or $0$
29:             $v_2 \leftarrow f(t, model_2)$ or $0$
30:             $value \leftarrow mode(v_1, v_2)$
31:             $merged \leftarrow (t, value)$
32:         **end if**
33:     **end for**
34:     **return** $merged$
35: **end function**

- MAX: use the maximum value ($max(v_1, v_2)$)

- MIN: use the minimum value ($min(v_1, v_2)$)

- MEAN: use the mean value ($\frac{v_1 + v_2}{2}$)

- ADD: use the sum of the values ($v_1 + v_2$)

If the random iteration count $j > 0$, a random word is chosen and the induction is run once forward and once backward starting from this position. Then, the most similar models from each set are merged and added to the set of *probable* models. It should be noted that setting the parameter $j > 0$ will make the algorithm non-deterministic.

The model induction is then repeated while the iteration count $i$ has not been reached or until no more models are induced, with the difference that for each word, each *probable* model is first consulted. If any of the probable models yields a score higher than the threshold value $t$, it is assumed that the word is already well represented by one of the probable models and no models are induced for this word. If the score falls below the threshold value $t$, induction is run as described.

At the end of the induction loop, all probable models are checked against each other. While there are two models that have a similarity below the silver threshold value $s$, the two models are merged and added to the set of *very probable* (*"gold"*) models.

If the set of probable models is not empty after this merging step, all remaining probable models are added to the set of very probable models.

In the second stage, the text is segmented according to the induced "gold" models. For each word, the language model with the highest probability for the word is chosen as that word's hypothetical language model.

### 4.4.1 Distributional similarity

Suppose we have three models with the distributions of letters as shown in figures 19, 20 and 21[6]. Similarity could be calculated based on the occurrence of unigrams/letters alone, i.e. if $model_1$ contains the letter 'a' and $model_2$ also contains the letter 'a', their similarity increases by 1.

However, if we calculate similarity in such a way, all three models are equally similar to each other, as each of the letters occurs at least once in each model. Yet, it should be clear that models 1 and 2 are very similar to each other while model 3 is dissimilar.

Therefore, in order to include the distribution of letters in the similarity measure, similarity is calculated as shown in algorithm 6.

---

[6]The figures shown are used for illustration purposes only and do not necessarily reflect real language models.

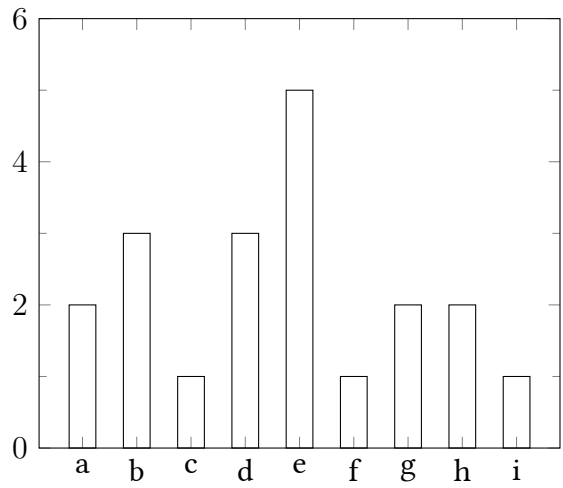

Figure 19: Language model: Distribution 1

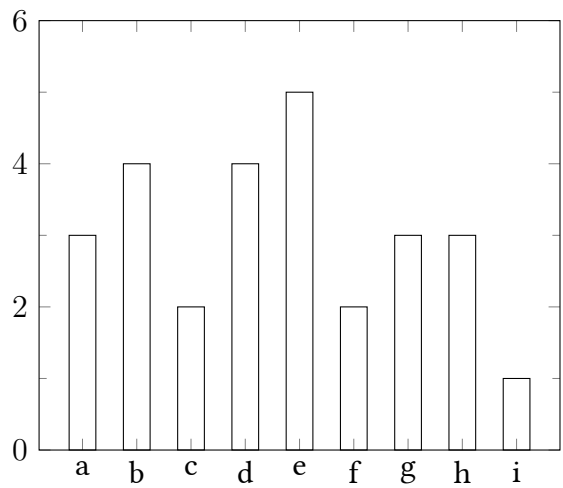

Figure 20: Language Model: Distribution 2

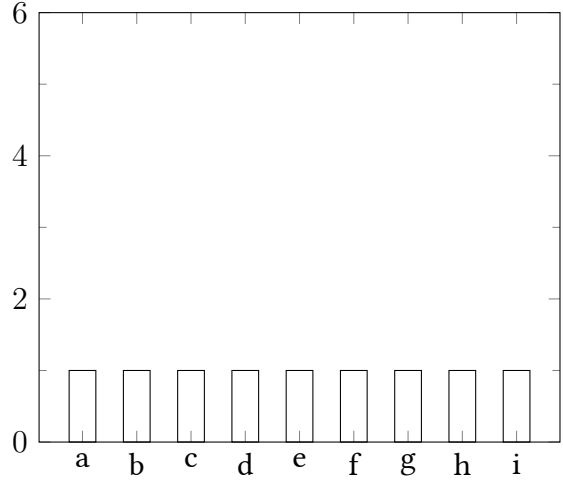

Figure 21: Language model: Distribution 3

---

**Algorithm 6** Distributional Similarity Calculation

---

1: **function** SIMILARITY($model_1$,$model_2$)
2:     similarity $\leftarrow 0$
3:     difference $\leftarrow 1$         ▷ Initialize difference to 1 to avoid division by zero
4:     **for** unigram $u_1$ in $model_1$.unigrams **do**
5:         **for** unigram $u_2$ in $model_2$.unigrams **do**
6:             **if** $u_1 = u_2$ **then**         ▷ unigram occurs in both models
7:                 $v_1 \leftarrow \frac{f(u_1)}{model_1.size}$       ▷ Normalize value by model size
8:                 $v_2 \leftarrow \frac{f(u_2)}{model_2.size}$
9:                 $q \leftarrow \frac{|v_1 - v_2|}{v_1 + v_2}$
10:                similarity $\leftarrow$ similarity $+(2-q)$
11:             **else**
12:                 difference $\leftarrow$ difference $+1$
13:             **end if**
14:         **end for**
15:     **end for**
16:     **return** $\frac{similarity}{difference}$
17: **end function**

---

with $f(c)$ returning the frequency of the character $c$. The number 2 in $(2-q)$ in line 10 can be explained as follows: $q$ expresses the *dissimilarity* of the models with regard to a unigram distribution with $0 \leqslant q \leqslant 1$, hence $(1-q)$ expresses the *similarity*. To this, we add 1, as we increase similarity by 1 due to the match; we augment the simple increase of 1 by the similarity of the distribution.

### 4.4.2 Evaluating results

The results of this approach can be interpreted as clusters, where each language model represents one cluster core and all words assigned to that model making up that cluster. Evaluation will hence be analogous to the evaluation of the clustering approach.

### 4.4.3 Estimating the parameters

As the language model induction can be controlled by parameters, we have to find a combination of parameters that works well for our task. The parameters $i$, $j$ and "merge mode" have been estimated on the development set. The development set contains similar documents to those in the test set. The development set can be found in the appendix.

It has been found that the parameter combination $i = 4, j = 2$, ADD yields good results across the development set. Hence, these values have been used for the test set evaluation.

# 5 Results

'Baseline' indicates the measurement where all words have been thrown into one cluster, measured against the gold standard. For 'Baseline 2', every word has been put into its own cluster and this clustering is evaluated against the gold standard. The column 'F1' stands for the F1 score and the 'F5' column stands for the F5 score.

If any of the 'runs' yields a higher score than any of the baseline values, the maximum score is indicated in bold. If a field contains 'n/a', this means that the value could not be calculated for whatever reason (most often a *division by zero* would have occurred).

## 5.1 N-Gram language model

| | Rand | Jaccard | Fowlkes-Mallows | F1 | F5 |
|---|---|---|---|---|---|
| **German–English** | | | | | |
| Baseline | 0.9259 | 0.9259 | 0.9622 | 0.9615 | 0.9285 |
| Baseline 2 | 0.0000 | 0.0740 | n/a | n/a | n/a |
| NGLM | 0.5200 | 0.4428 | 0.6597 | 0.6138 | 0.9275 |
| **German–Finnish–Turkish** | | | | | |
| Baseline | 0.3312 | 0.3312 | 0.5755 | 0.4976 | 0.3400 |
| Baseline 2 | 0.6721 | 0.0103 | 0.1015 | 0.0204 | 0.2132 |
| NGLM | **0.8104** | **0.3872** | 0.5615 | **0.5582** | **0.5081** |
| **English–French** | | | | | |
| Baseline | 0.7038 | 0.7038 | 0.8389 | 0.8261 | 0.7119 |
| Baseline 2 | 0.3064 | 0.0145 | 0.1207 | 0.0287 | 0.2777 |
| NGLM | 0.6246 | 0.3540 | 0.5322 | 0.5229 | 0.4459 |
| **English–Transliterated Greek** | | | | | |
| Baseline | 0.8809 | 0.8809 | 0.9385 | 0.9385 | 0.8850 |
| Baseline 2 | 0.1269 | 0.0090 | 0.0949 | 0.0178 | 0.1911 |
| NGLM | 0.6932 | 0.5492 | 0.7117 | 0.7090 | 0.7708 |
| **Italian–German** | | | | | |
| Baseline | 0.5807 | 0.5807 | 0.7620 | 0.7347 | 0.5902 |
| Baseline 2 | 0.4227 | 0.0060 | 0.0776 | 0.0119 | 0.1360 |
| NGLM | **0.7010** | 0.2977 | 0.4740 | 0.4589 | **0.5969** |

Table 4: N-Gram language model results: Latin script

|  | Rand | Jaccard | Fowlkes-Mallows | F1 | F5 |
| --- | --- | --- | --- | --- | --- |
| **Greek–Russian** | | | | | |
| Baseline | 0.5578 | 0.5578 | 0.7468 | 0.7161 | 0.5674 |
| Baseline 2 | 0.4440 | 0.0034 | 0.0584 | 0.0068 | 0.0817 |
| NGLM | **0.7597** | 0.5108 | 0.6762 | 0.6762 | **0.6694** |
| **English–Greek** | | | | | |
| Baseline | 0.9179 | 0.9179 | 0.9580 | 0.9571 | 0.9208 |
| Baseline 2 | 0.0946 | 0.0136 | 0.1167 | 0.0269 | 0.2643 |
| NGLM | 0.5665 | 0.3867 | 0.5586 | 0.5577 | 0.5877 |
| **English–Spanish–Arabic** | | | | | |
| Baseline | 0.3354 | 0.3354 | 0.5791 | 0.5023 | 0.3442 |
| Baseline 2 | 0.6682 | 0.0109 | 0.1044 | 0.0215 | 0.2227 |
| NGLM | **0.9204** | **0.7489** | **0.8573** | **0.8564** | **0.8936** |
| **English–Chinese** | | | | | |
| Baseline | 0.8474 | 0.8474 | 0.9205 | 0.9174 | 0.8524 |
| Baseline 2 | 0.1595 | 0.0082 | 0.0909 | 0.0164 | 0.1781 |
| NGLM | 0.6573 | 0.4476 | 0.6259 | 0.6184 | 0.7208 |
| **Ukrainian–Russian** | | | | | |
| Baseline | 0.4950 | 0.4950 | 0.7035 | 0.6622 | 0.5048 |
| Baseline 2 | 0.5060 | 0.0022 | 0.0472 | 0.0044 | 0.0550 |
| NGLM | **0.6755** | 0.3857 | 0.5644 | 0.5567 | 0.4831 |

Table 5: N-Gram language model results: Mixed script

|  | Rand | Jaccard | Fowlkes-Mallows | F1 | F5 |
| --- | --- | --- | --- | --- | --- |
| **Pali 1** | | | | | |
| Baseline | 0.3131 | 0.3131 | 0.5595 | 0.4768 | 0.3216 |
| Baseline 2 | 0.6906 | 0.0118 | 0.1089 | 0.0234 | 0.2379 |
| NGLM | **0.8153** | 0.2069 | 0.3434 | 0.3429 | **0.3608** |
| **Pali 2** | | | | | |
| Baseline | 0.3589 | 0.3589 | 0.5991 | 0.5283 | 0.3680 |
| Baseline 2 | 0.6495 | 0.0238 | 0.1543 | 0.0465 | 0.3880 |
| NGLM | **0.7173** | 0.1958 | 0.3336 | 0.3275 | **0.3971** |
| **Pali 3** | | | | | |
| Baseline | 0.4947 | 0.4947 | 0.7033 | 0.6619 | 0.5045 |
| Baseline 2 | 0.5075 | 0.0045 | 0.0676 | 0.0091 | 0.1067 |
| NGLM | **0.7874** | 0.0816 | 0.1692 | 0.1508 | 0.1064 |
| **Pali 4** | | | | | |
| Baseline | 0.4000 | 0.4000 | 0.6324 | 0.5714 | 0.4094 |
| Baseline 2 | 0.6000 | 0.0000 | n/a | n/a | n/a |
| NGLM | 0.3000 | 0.1250 | 0.2357 | 0.2222 | 0.1699 |
| **Pali 5** | | | | | |
| Baseline | 0.5800 | 0.5800 | 0.7615 | 0.7341 | 0.5895 |
| Baseline 2 | 0.4236 | 0.0063 | 0.0798 | 0.0126 | 0.1430 |
| NGLM | 0.4777 | 0.2496 | 0.4065 | 0.3995 | 0.4816 |

Table 6: N-Gram language model results: Pali data

|  | Rand | Jaccard | Fowlkes-Mallows | F1 | F5 |
|---|---|---|---|---|---|
| **Twitter 1** | | | | | |
| Baseline | 0.4615 | 0.4615 | 0.6793 | 0.6315 | 0.4712 |
| Baseline 2 | 0.5384 | 0.0000 | n/a | n/a | n/a |
| NGLM | **0.8589** | **0.5925** | **0.7542** | **0.7441** | **0.8757** |
| **Twitter 2** | | | | | |
| Baseline | 0.5555 | 0.5555 | 0.7453 | 0.7142 | 0.5652 |
| Baseline 2 | 0.4444 | 0.0000 | n/a | n/a | n/a |
| NGLM | **0.7485** | **0.6090** | **0.7591** | **0.7570** | **0.8121** |
| **Twitter 3** | | | | | |
| Baseline | 0.6583 | 0.6583 | 0.8113 | 0.7939 | 0.6670 |
| Baseline 2 | 0.3416 | 0.0000 | n/a | n/a | n/a |
| NGLM | **0.6750** | 0.4347 | 0.6479 | 0.6060 | **0.8996** |
| **Twitter 4** | | | | | |
| Baseline | 0.8750 | 0.8750 | 0.9354 | 0.9333 | 0.8792 |
| Baseline 2 | 0.1250 | 0.0000 | n/a | n/a | n/a |
| NGLM | 0.7250 | 0.5822 | 0.7597 | 0.7360 | **0.9545** |
| **Twitter 5** | | | | | |
| Baseline | 0.4285 | 0.4285 | 0.6546 | 0.6000 | 0.4382 |
| Baseline 2 | 0.5714 | 0.0000 | n/a | n/a | n/a |
| NGLM | **0.6666** | 0.1250 | 0.2672 | 0.2222 | **0.4561** |

Table 7: N-Gram language model results: Twitter data

## 5.2 Textcat

| | Rand | Jaccard | Fowlkes-Mallows | F1 | F5 |
|---|---|---|---|---|---|
| **German–English** | | | | | |
| Baseline | 0.9259 | 0.9259 | 0.9622 | 0.9615 | 0.9285 |
| Baseline 2 | 0.0000 | 0.0740 | n/a | n/a | n/a |
| Textcat | 0.8632 | 0.8518 | 0.9200 | 0.9200 | 0.9200 |
| **German–Finnish–Turkish** | | | | | |
| Baseline | 0.3312 | 0.3312 | 0.5755 | 0.4976 | 0.3400 |
| Baseline 2 | 0.6721 | 0.0103 | 0.1015 | 0.0204 | 0.2132 |
| Textcat | 0.4095 | 0.1903 | 0.3823 | 0.3198 | 0.2124 |
| **English–French** | | | | | |
| Baseline | 0.7038 | 0.7038 | 0.8389 | 0.8261 | 0.7119 |
| Baseline 2 | 0.3064 | 0.0145 | 0.1207 | 0.0287 | 0.2777 |
| Textcat | 0.3890 | 0.3211 | 0.5476 | 0.4861 | 0.3411 |
| **English–Transliterated Greek** | | | | | |
| Baseline | 0.8809 | 0.8809 | 0.9385 | 0.9385 | 0.8850 |
| Baseline 2 | 0.1269 | 0.0090 | 0.0949 | 0.0178 | 0.1911 |
| Textcat | 0.5202 | 0.4853 | 0.6678 | 0.6535 | 0.5492 |
| **Italian–German** | | | | | |
| Baseline | 0.5807 | 0.5807 | 0.7620 | 0.7347 | 0.5902 |
| Baseline 2 | 0.4227 | 0.0060 | 0.0776 | 0.0119 | 0.1360 |
| Textcat | 0.4030 | 0.3057 | 0.5014 | 0.4682 | 0.3520 |

Table 8: Textcat results: Latin script

|            | Rand   | Jaccard | Fowlkes-Mallows | F1     | F5     |
|------------|--------|---------|-----------------|--------|--------|
| **Greek–Russian** | | | | | |
| Baseline   | 0.5578 | 0.5578  | 0.7468          | 0.7161 | 0.5674 |
| Baseline 2 | 0.4440 | 0.0034  | 0.0584          | 0.0068 | 0.0817 |
| Textcat    | 0.4468 | 0.2971  | 0.4769          | 0.4581 | 0.3644 |
| **English–Greek** | | | | | |
| Baseline   | 0.9179 | 0.9179  | 0.9580          | 0.9571 | 0.9208 |
| Baseline 2 | 0.0946 | 0.0136  | 0.1167          | 0.0269 | 0.2643 |
| Textcat    | 0.5357 | 0.4933  | 0.6730          | 0.6607 | 0.5619 |
| **English–Spanish–Arabic** | | | | | |
| Baseline   | 0.3354 | 0.3354  | 0.5791          | 0.5023 | 0.3442 |
| Baseline 2 | 0.6682 | 0.0109  | 0.1044          | 0.0215 | 0.2227 |
| Textcat    | 0.3956 | 0.2832  | 0.5042          | 0.4414 | 0.3052 |
| **English–Chinese** | | | | | |
| Baseline   | 0.8474 | 0.8474  | 0.9205          | 0.9174 | 0.8524 |
| Baseline 2 | 0.1595 | 0.0082  | 0.0909          | 0.0164 | 0.1781 |
| Textcat    | 0.5018 | 0.4468  | 0.6251          | 0.6177 | 0.5408 |
| **Ukrainian–Russian** | | | | | |
| Baseline   | 0.4950 | 0.4950  | 0.7035          | 0.6622 | 0.5048 |
| Baseline 2 | 0.5060 | 0.0022  | 0.0472          | 0.0044 | 0.0550 |
| Textcat    | 0.3787 | 0.2625  | 0.4472          | 0.4159 | 0.3105 |

Table 9: Textcat results: Mixed script

|  | Rand | Jaccard | Fowlkes-Mallows | F1 | F5 |
|---|---|---|---|---|---|
| **Pali 1** | | | | | |
| Baseline | 0.3131 | 0.3131 | 0.5595 | 0.4768 | 0.3216 |
| Baseline 2 | 0.6906 | 0.0118 | 0.1089 | 0.0234 | 0.2379 |
| Textcat | 0.4531 | 0.2508 | 0.4849 | 0.4011 | 0.2641 |
| **Pali 2** | | | | | |
| Baseline | 0.3589 | 0.3589 | 0.5991 | 0.5283 | 0.3680 |
| Baseline 2 | 0.6495 | 0.0238 | 0.1543 | 0.0465 | 0.3880 |
| Textcat | 0.4307 | 0.2745 | 0.5088 | 0.4307 | 0.2888 |
| **Pali 3** | | | | | |
| Baseline | 0.4947 | 0.4947 | 0.7033 | 0.6619 | 0.5045 |
| Baseline 2 | 0.5075 | 0.0045 | 0.0676 | 0.0091 | 0.1067 |
| Textcat | 0.2032 | 0.0704 | 0.2502 | 0.1315 | 0.0736 |
| **Pali 4** | | | | | |
| Baseline | 0.4000 | 0.4000 | 0.6324 | 0.5714 | 0.4094 |
| Baseline 2 | 0.6000 | 0.0000 | n/a | n/a | n/a |
| Textcat | 0.5000 | 0.1666 | 0.2886 | 0.2857 | 0.2524 |
| **Pali 5** | | | | | |
| Baseline | 0.5800 | 0.5800 | 0.7615 | 0.7341 | 0.5895 |
| Baseline 2 | 0.4236 | 0.0063 | 0.0798 | 0.0126 | 0.1430 |
| Textcat | 0.5090 | 0.3458 | 0.5141 | 0.5140 | 0.5236 |

Table 10: Textcat results: Pali data

|            | Rand   | Jaccard | Fowlkes-Mallows | F1     | F5     |
|------------|--------|---------|-----------------|--------|--------|
| **Twitter 1** |     |         |                 |        |        |
| Baseline   | 0.4615 | 0.4615  | 0.6793          | 0.6315 | 0.4712 |
| Baseline 2 | 0.5384 | 0.0000  | n/a             | n/a    | n/a    |
| Textcat    | 0.3736 | 0.2597  | 0.4460          | 0.4123 | 0.3049 |
| **Twitter 2** |     |         |                 |        |        |
| Baseline   | 0.5555 | 0.5555  | 0.7453          | 0.7142 | 0.5652 |
| Baseline 2 | 0.4444 | 0.0000  | n/a             | n/a    | n/a    |
| Textcat    | 0.4678 | 0.4347  | 0.6158          | 0.6060 | 0.5207 |
| **Twitter 3** |     |         |                 |        |        |
| Baseline   | 0.6583 | 0.6583  | 0.8113          | 0.7939 | 0.6670 |
| Baseline 2 | 0.3416 | 0.0000  | n/a             | n/a    | n/a    |
| Textcat    | **0.6838** | 0.6446 | 0.8011      | 0.7839 | 0.6586 |
| **Twitter 4** |     |         |                 |        |        |
| Baseline   | 0.8750 | 0.8750  | 0.9354          | 0.9333 | 0.8792 |
| Baseline 2 | 0.1250 | 0.0000  | n/a             | n/a    | n/a    |
| Textcat    | **0.8833** | 0.8666 | 0.9309      | 0.9285 | 0.8711 |
| **Twitter 5** |     |         |                 |        |        |
| Baseline   | 0.4285 | 0.4285  | 0.6546          | 0.6000 | 0.4382 |
| Baseline 2 | 0.5714 | 0.0000  | n/a             | n/a    | n/a    |
| Textcat    | 0.3333 | 0.3333  | 0.5773          | 0.5000 | 0.3421 |

Table 11: Textcat results: Twitter data

## 5.3 Clustering

The first run indicates the value after one clustering step, and the second run indicates the value after applying the clustering algorithm to the results of the first run.

| | Rand | Jaccard | Fowlkes-Mallows | F1 | F5 |
|---|---|---|---|---|---|
| **German–English** | | | | | |
| Baseline | 0.9259 | 0.9259 | 0.9622 | 0.9615 | 0.9285 |
| Baseline 2 | 0.0000 | 0.0740 | n/a | n/a | n/a |
| First run | 0.4102 | 0.3929 | 0.6069 | 0.5642 | 0.8549 |
| Second run | 0.2336 | 0.1970 | 0.4199 | 0.3291 | 0.7712 |
| **German–Finnish–Turkish** | | | | | |
| Baseline | 0.3312 | 0.3312 | 0.5755 | 0.4976 | 0.3400 |
| Baseline 2 | 0.6721 | 0.0103 | 0.1015 | 0.0204 | 0.2132 |
| First run | 0.4841 | 0.1764 | 0.3369 | 0.2998 | 0.2110 |
| Second run | 0.6259 | 0.1611 | 0.2840 | 0.2775 | 0.2320 |
| **English–French** | | | | | |
| Baseline | 0.7038 | 0.7038 | 0.8389 | 0.8261 | 0.7119 |
| Baseline 2 | 0.3064 | 0.0145 | 0.1207 | 0.0287 | 0.2777 |
| First run | 0.4051 | 0.2980 | 0.5001 | 0.4592 | 0.3362 |
| Second run | 0.4601 | 0.1836 | 0.3116 | 0.3103 | 0.2857 |
| **English–Transliterated Greek** | | | | | |
| Baseline | 0.8809 | 0.8809 | 0.9385 | 0.9385 | 0.8850 |
| Baseline 2 | 0.1269 | 0.0090 | 0.0949 | 0.0178 | 0.1911 |
| First run | 0.5867 | 0.3977 | 0.5725 | 0.5691 | 0.6320 |
| Second run | 0.5423 | 0.3161 | 0.4909 | 0.4804 | 0.5934 |
| **Italian–German** | | | | | |
| Baseline | 0.5807 | 0.5807 | 0.7620 | 0.7347 | 0.5902 |
| Baseline 2 | 0.4227 | 0.0060 | 0.0776 | 0.0119 | 0.1360 |
| First run | 0.4222 | 0.2838 | 0.4640 | 0.4421 | 0.3453 |
| Second run | 0.4915 | 0.2472 | 0.4006 | 0.3964 | 0.3499 |

Table 12: Clustering results: Latin script

|  | Rand | Jaccard | Fowlkes-Mallows | F1 | F5 |
|---|---|---|---|---|---|
| **Greek–Russian** | | | | | |
| Baseline | 0.5578 | 0.5578 | 0.7468 | 0.7161 | 0.5674 |
| Baseline 2 | 0.4440 | 0.0034 | 0.0584 | 0.0068 | 0.0817 |
| First run | 0.5787 | 0.3811 | 0.5672 | 0.5519 | 0.4549 |
| Second run | **0.7536** | 0.3883 | 0.5899 | 0.4494 | **0.7914** |
| **English–Greek** | | | | | |
| Baseline | 0.9179 | 0.9179 | 0.9580 | 0.9571 | 0.9208 |
| Baseline 2 | 0.0946 | 0.0136 | 0.1167 | 0.0269 | 0.2643 |
| First run | 0.4244 | 0.2482 | 0.4015 | 0.3977 | 0.4553 |
| Second run | 0.3705 | 0.0855 | 0.1784 | 0.1576 | 0.2777 |
| **English–Spanish–Arabic** | | | | | |
| Baseline | 0.3354 | 0.3354 | 0.5791 | 0.5023 | 0.3442 |
| Baseline 2 | 0.6682 | 0.0109 | 0.1044 | 0.0215 | 0.2227 |
| First run | **0.8016** | **0.5650** | **0.7400** | **0.7221** | **0.6008** |
| Second run | **0.7226** | 0.2860 | 0.4495 | 0.4448 | **0.5130** |
| **English–Chinese** | | | | | |
| Baseline | 0.8474 | 0.8474 | 0.9205 | 0.9174 | 0.8524 |
| Baseline 2 | 0.1595 | 0.0082 | 0.0909 | 0.0164 | 0.1781 |
| First run | 0.5480 | 0.3356 | 0.5087 | 0.5025 | 0.5866 |
| Second run | 0.5138 | 0.2584 | 0.4361 | 0.4107 | 0.5957 |
| **Ukrainian–Russian** | | | | | |
| Baseline | 0.4950 | 0.4950 | 0.7035 | 0.6622 | 0.5048 |
| Baseline 2 | 0.5060 | 0.0022 | 0.0472 | 0.0044 | 0.0550 |
| First run | **0.5867** | 0.1953 | 0.3268 | 0.3267 | 0.3305 |
| Second run | **0.5934** | 0.1154 | 0.2178 | 0.2070 | 0.2907 |

Table 13: Clustering results: Mixed script

|            | Rand   | Jaccard | Fowlkes-<br>Mallows | F1     | F5     |
|------------|--------|---------|---------------------|--------|--------|
| **Pali 1** |        |         |                     |        |        |
| Baseline   | 0.3131 | 0.3131  | 0.5595              | 0.4768 | 0.3216 |
| Baseline 2 | 0.6906 | 0.0118  | 0.1089              | 0.0234 | 0.2379 |
| First run  | 0.4674 | 0.2540  | 0.4898              | 0.4051 | 0.2666 |
| Second run | **0.7168** | 0.2547 | 0.4118         | 0.4060 | **0.3516** |
| **Pali 2** |        |         |                     |        |        |
| Baseline   | 0.3589 | 0.3589  | 0.5991              | 0.5283 | 0.3680 |
| Baseline 2 | 0.6495 | 0.0238  | 0.1543              | 0.0465 | 0.3880 |
| First run  | **0.6738** | 0.3026 | 0.4777          | 0.4646 | 0.3825 |
| Second run | **0.6646** | 0.1865 | 0.3147          | 0.3144 | 0.3021 |
| **Pali 3** |        |         |                     |        |        |
| Baseline   | 0.4947 | 0.4947  | 0.7033              | 0.6619 | 0.5045 |
| Baseline 2 | 0.5075 | 0.0045  | 0.0676              | 0.0091 | 0.1067 |
| First run  | **0.5686** | 0.0746 | 0.2002          | 0.1389 | 0.0831 |
| Second run | **0.7534** | 0.0911 | 0.1962          | 0.1670 | 0.1125 |
| **Pali 4** |        |         |                     |        |        |
| Baseline   | 0.4000 | 0.4000  | 0.6324              | 0.5714 | 0.4094 |
| Baseline 2 | 0.6000 | 0.0000  | n/a                 | n/a    | n/a    |
| First run  | 0.5333 | 0.3000  | 0.5477              | 0.4615 | 0.3083 |
| Second run | 0.3000 | 0.3000  | 0.5477              | 0.4615 | 0.3083 |
| **Pali 5** |        |         |                     |        |        |
| Baseline   | 0.5800 | 0.5800  | 0.7615              | 0.7341 | 0.5895 |
| Baseline 2 | 0.4236 | 0.0063  | 0.0798              | 0.0126 | 0.1430 |
| First run  | 0.5294 | 0.2472  | 0.4111              | 0.3965 | 0.5242 |
| Second run | 0.4666 | 0.1214  | 0.2524              | 0.2166 | 0.4117 |

Table 14: Clustering results: Pali data

|  | Rand | Jaccard | Fowlkes-Mallows | F1 | F5 |
|---|---|---|---|---|---|
| **Twitter 1** | | | | | |
| Baseline | 0.4615 | 0.4615 | 0.6793 | 0.6315 | 0.4712 |
| Baseline 2 | 0.5384 | 0.0000 | n/a | n/a | n/a |
| First run | **0.8681** | **0.7142** | **0.8451** | **0.8333** | **0.7222** |
| Second run | **0.8461** | **0.6000** | **0.7745** | **0.7499** | **0.9750** |
| **Twitter 2** | | | | | |
| Baseline | 0.5555 | 0.5555 | 0.7453 | 0.7142 | 0.5652 |
| Baseline 2 | 0.4444 | 0.0000 | n/a | n/a | n/a |
| First run | 0.4575 | 0.3941 | 0.5655 | 0.5654 | 0.5573 |
| Second run | 0.4967 | 0.3888 | 0.5615 | 0.5600 | **0.6012** |
| **Twitter 3** | | | | | |
| Baseline | 0.6583 | 0.6583 | 0.8113 | 0.7939 | 0.6670 |
| Baseline 2 | 0.3416 | 0.0000 | n/a | n/a | n/a |
| First run | 0.4571 | 0.3595 | 0.5525 | 0.5289 | **0.7215** |
| Second run | 0.3523 | 0.2093 | 0.3997 | 0.3461 | 0.6428 |
| **Twitter 4** | | | | | |
| Baseline | 0.8750 | 0.8750 | 0.9354 | 0.9333 | 0.8792 |
| Baseline 2 | 0.1250 | 0.0000 | n/a | n/a | n/a |
| First run | **0.9019** | **0.8584** | 0.9265 | 0.9238 | 0.8631 |
| Second run | 0.6250 | 0.5000 | 0.6789 | 0.6666 | 0.8080 |
| **Twitter 5** | | | | | |
| Baseline | 0.4285 | 0.4285 | 0.6546 | 0.6000 | 0.4382 |
| Baseline 2 | 0.5714 | 0.0000 | n/a | n/a | n/a |
| First run | **0.7142** | **0.4666** | **0.6831** | **0.6363** | **0.4764** |
| Second run | 0.5714 | 0.3076 | 0.4780 | 0.4705 | 0.4046 |

Table 15: Clustering results: Twitter data

## 5.4   Language model induction

In addition to highlighting results that outperform the baseline values, the following tables have been color coded. Results that outperform the clustering algorithm are indicated in red and results that outperform both the clustering algorithm and the n-gram language model are indicated in blue.[7]

| | Rand | Jaccard | Fowlkes-Mallows | F1 | F5 |
|---|---|---|---|---|---|
| **German–English** | | | | | |
| Baseline | 0.9259 | 0.9259 | 0.9622 | 0.9615 | 0.9285 |
| Baseline 2 | 0.0000 | 0.0740 | n/a | n/a | n/a |
| Induced | 0.6837 | 0.6574 | 0.7988 | 0.7932 | 0.8896 |
| **German–Finnish–Turkish** | | | | | |
| Baseline | 0.3312 | 0.3312 | 0.5755 | 0.4976 | 0.3400 |
| Baseline 2 | 0.6721 | 0.0103 | 0.1015 | 0.0204 | 0.2132 |
| Induced | 0.6438 | 0.1771 | 0.3057 | 0.3009 | 0.2588 |
| **English–French** | | | | | |
| Baseline | 0.7038 | 0.7038 | 0.8389 | 0.8261 | 0.7119 |
| Baseline 2 | 0.3064 | 0.0145 | 0.1207 | 0.0287 | 0.2777 |
| Induced | 0.6171 | 0.2835 | 0.4427 | 0.4418 | 0.4692 |
| **English–Transliterated Greek** | | | | | |
| Baseline | 0.8809 | 0.8809 | 0.9385 | 0.9385 | 0.8850 |
| Baseline 2 | 0.1269 | 0.0090 | 0.0949 | 0.0178 | 0.1911 |
| Induced | 0.4436 | 0.2398 | 0.4277 | 0.3868 | 0.6382 |
| **Italian–German** | | | | | |
| Baseline | 0.5807 | 0.5807 | 0.7620 | 0.7347 | 0.5902 |
| Baseline 2 | 0.4227 | 0.0060 | 0.0776 | 0.0119 | 0.1360 |
| Induced | 0.5658 | 0.1536 | 0.2871 | 0.2664 | 0.4065 |

Table 16: Induction results: Latin script

---

[7]Results that outperform only the n-gram language model would have been indicated in green, but there is no score that outperforms only the n-gram language model.

|  | Rand | Jaccard | Fowlkes-Mallows | F1 | F5 |
|---|---|---|---|---|---|
| **Greek–Russian** | | | | | |
| Baseline | 0.5578 | 0.5578 | 0.7468 | 0.7161 | 0.5674 |
| Baseline 2 | 0.4440 | 0.0034 | 0.0584 | 0.0068 | 0.0817 |
| Inducted | **0.7142** | <span style="color:red">0.4222</span> | <span style="color:red">0.5940</span> | <span style="color:red">0.5937</span> | **0.6125** |
| **English–Greek** | | | | | |
| Baseline | 0.9179 | 0.9179 | 0.9580 | 0.9571 | 0.9208 |
| Baseline 2 | 0.0946 | 0.0136 | 0.1167 | 0.0269 | 0.2643 |
| Inducted | <span style="color:red">0.4769</span> | <span style="color:red">0.3266</span> | <span style="color:red">0.5089</span> | <span style="color:red">0.4924</span> | <span style="color:blue">0.6423</span> |
| **English–Spanish–Arabic** | | | | | |
| Baseline | 0.3354 | 0.3354 | 0.5791 | 0.5023 | 0.3442 |
| Baseline 2 | 0.6682 | 0.0109 | 0.1044 | 0.0215 | 0.2227 |
| Inducted | **0.7783** | <span style="color:red">0.5677</span> | <span style="color:red">0.7534</span> | <span style="color:red">0.7242</span> | **0.5773** |
| **English–Chinese** | | | | | |
| Baseline | 0.8474 | 0.8474 | 0.9205 | 0.9174 | 0.8524 |
| Baseline 2 | 0.1595 | 0.0082 | 0.0909 | 0.0164 | 0.1781 |
| Inducted | <span style="color:red">0.5657</span> | 0.3343 | <span style="color:red">0.5258</span> | 0.5011 | <span style="color:red">0.6953</span> |
| **Ukrainian–Russian** | | | | | |
| Baseline | 0.4950 | 0.4950 | 0.7035 | 0.6622 | 0.5048 |
| Baseline 2 | 0.5060 | 0.0022 | 0.0472 | 0.0044 | 0.0550 |
| Inducted | <span style="color:red">0.6289</span> | 0.1000 | 0.1935 | 0.1818 | 0.2659 |

Table 17: Induction results: Mixed script

|  | Rand | Jaccard | Fowlkes-Mallows | F1 | F5 |
|---|---|---|---|---|---|
| **Pali 1** | | | | | |
| Baseline | 0.3131 | 0.3131 | 0.5595 | 0.4768 | 0.3216 |
| Baseline 2 | 0.6906 | 0.0118 | 0.1089 | 0.0234 | 0.2379 |
| Inducted | 0.7856 | 0.1683 | 0.2898 | 0.2882 | 0.3188 |
| **Pali 2** | | | | | |
| Baseline | 0.3589 | 0.3589 | 0.5991 | 0.5283 | 0.3680 |
| Baseline 2 | 0.6495 | 0.0238 | 0.1543 | 0.0465 | 0.3880 |
| Inducted | 0.8148 | 0.5000 | 0.6686 | 0.6666 | 0.7176 |
| **Pali 3** | | | | | |
| Baseline | 0.4947 | 0.4947 | 0.7033 | 0.6619 | 0.5045 |
| Baseline 2 | 0.5075 | 0.0045 | 0.0676 | 0.0091 | 0.1067 |
| Inducted | 0.8492 | 0.0569 | 0.1083 | 0.1078 | 0.1186 |
| **Pali 4** | | | | | |
| Baseline | 0.4000 | 0.4000 | 0.6324 | 0.5714 | 0.4094 |
| Baseline 2 | 0.6000 | 0.0000 | n/a | n/a | n/a |
| Inducted | 0.6000 | 0.0000 | 0.0000 | n/a | n/a |
| **Pali 5** | | | | | |
| Baseline | 0.5800 | 0.5800 | 0.7615 | 0.7341 | 0.5895 |
| Baseline 2 | 0.4236 | 0.0063 | 0.0798 | 0.0126 | 0.1430 |
| Inducted | 0.4033 | 0.2082 | 0.3504 | 0.3446 | 0.4134 |

Table 18: Induction results: Pali data

|            | Rand   | Jaccard | Fowlkes-Mallows | F1     | F5     |
|------------|--------|---------|-----------------|--------|--------|
| **Twitter 1** |     |         |                 |        |        |
| Baseline   | 0.4615 | 0.4615  | 0.6793          | 0.6315 | 0.4712 |
| Baseline 2 | 0.5384 | 0.0000  | n/a             | n/a    | n/a    |
| Inducted   | **0.6282** | 0.3695 | 0.5515      | 0.5396 | 0.4533 |
| **Twitter 2** |     |         |                 |        |        |
| Baseline   | 0.5555 | 0.5555  | 0.7453          | 0.7142 | 0.5652 |
| Baseline 2 | 0.4444 | 0.0000  | n/a             | n/a    | n/a    |
| Inducted   | <span style="color:blue">**0.7719**</span> | <span style="color:red">**0.6020**</span> | <span style="color:blue">**0.7687**</span> | <span style="color:red">**0.7515**</span> | <span style="color:blue">**0.9325**</span> |
| **Twitter 3** |     |         |                 |        |        |
| Baseline   | 0.6583 | 0.6583  | 0.8113          | 0.7939 | 0.6670 |
| Baseline 2 | 0.3416 | 0.0000  | n/a             | n/a    | n/a    |
| Inducted   | <span style="color:red">0.5916</span> | 0.3000 | 0.5236 | <span style="color:red">0.4615</span> | **0.8185** |
| **Twitter 4** |     |         |                 |        |        |
| Baseline   | 0.8750 | 0.8750  | 0.9354          | 0.9333 | 0.8792 |
| Baseline 2 | 0.1250 | 0.0000  | n/a             | n/a    | n/a    |
| Inducted   | 0.5250 | 0.3736  | 0.5615          | 0.5439 | 0.7055 |
| **Twitter 5** |     |         |                 |        |        |
| Baseline   | 0.4285 | 0.4285  | 0.6546          | 0.6000 | 0.4382 |
| Baseline 2 | 0.5714 | 0.0000  | n/a             | n/a    | n/a    |
| Inducted   | <span style="color:blue">**1.0000**</span> | <span style="color:blue">**1.0000**</span> | <span style="color:blue">**1.0000**</span> | <span style="color:blue">**1.0000**</span> | <span style="color:blue">**1.0000**</span> |

Table 19: Induction results: Twitter data

# 6    Discussion

The work by Seldin et al. (2001) is similar to the work presented here. They propose an unsupervised language (and protein sequence) segmentation approach that yields accurate segmentations. While their work looks promising, it also has its drawbacks. Their method requires longer monolingual text fragments and a sizable amount of text. Furthermore, they disallow switching language models after each word. This presumption will fail to detect single-word inclusions and structures as shown in figure 22, where the language alternates after each word.

$$w_1 \quad w_2 \quad w_3 \quad w_4 \quad w_5 \quad w_6 \quad w_7 \quad \cdots$$

Figure 22: Alternating language structure

While this structure looks very artificial, such a structure is found, for instance, in the fifth Pali dictionary text, in the passage "Pacati, [Ved. pacati, Igd. *peqŭō, Av. pac-;". In this case, 'red' corresponds to Pali, 'blue' to (abbreviations in) English and 'green' to reconstructed Indo-european.

## 6.1    N-Gram language models

The trained n-gram language model approach works well on the Latin script data, managing to single out the German inclusion from the English–German text (even though it is classified as "other" instead of German).

For German–Finnish–Turkish, English–French, English–Transliterated Greek and Italian–German, the separation of the main languages involved is good, although there appear to be some problems when words contain non-word characters such as quotes or parentheses.

Some puzzling misclassifications happen in the English–Transliterated Greek case: *agápe* is considered English and *éros* is considered Transliterated Amharic.

In the Italian–German text, the Italian language leads to a rather important Spanish cluster due to the relatedness of the two Romance languages.

On the mixed script data set, the results are more diverse. Greek–Russian, English–Spanish–Arabic and Ukrainian–Russian are segmented well, with English–Spanish–Arabic having Spanish split into Spanish, French and Italian due to the relatedness of the languages.

In contrast, the segmentation of English–Greek did not work well at all. Of the two Greek words ἀγάπη and ἔρως, ἀγάπη was considered French and ἔρως was considered Russian. It must be noted, though, that these words bear polytonic diacritics, whereas the model was trained on monotonic Greek.

Also, the segmentation of English–Chinese did not work well. This is probably due to the way the model was trained. Chinese script is written without whitespace characters between words, and the correct segmentation of a text written in Chinese requires in-depth knowledge of the language. Some words are written with only one character, but others are composed of two or more characters, with the meaning often being non-compositional; the meaning of a two-character word is different from the sum of the meaning of the two characters. Sometimes, more than one segmentation would be possible and the context decides on which segmentation is correct. In other cases, more than one segmentation might be correct. This problem occurs with all scripts that are written without whitespace.

As with the simplified assumption in the tokenization of whitespace-scripts, where I consider a word to be a character sequence delineated by whitespace, I have treated each character as a word. Adapting the method to Chinese and similar scripts would have been possible, but would have introduced the need for large amounts of external linguistic knowledge. Indeed, every possible non-whitespace-script would have to be considered, and each of the tokenizers would be language dependent, i.e. a tokenizer for Chinese would not work on Korean or Japanese.

The supervised approach did not work well on the Pali dictionary data. While English words could be isolated somewhat successfully, the rest of the data proved difficult to segment. As an example, let us look at the first Pali text. The English cluster contains almost only English words, but not all, the "other" cluster contains mainly marked up words, and the rest is seemingly haphazardly distributed among the other models.

**Pali 1: abbha**

- (AR) ., 134., 289.

- (DE) Miln), imber, dark), Miln

- (EL) (=, (abbhaŋ

- (EN) water, mountain, of, free, (used, or, like, referred, (also, A, This, cloudy, clouds, later, a, froth, 1, summit, thundering, by, mass, Pv, Oir, obscure, scum, that, water]., thick, As, from, It, is, at, as, the, in, clouds, things, also

- (ES) (dense, f., sense, expl, rajo

- (FI) 239., rain;, Lat., Vin, perhaps, SnA

- (FR) cloud, Dh, adj., point, cloud, Dhs, A), rain, VvA, DhsA, list

- (IT) \"dark, &, ambha, 3, 1, 317, J, sunshine, cp., abhra, [Vedic, (megho

- (PL) 487, =, S, 295, <br, moon−, 249

- (RU) 348, 53

- (TR) viz., ambu, Vv

- (TrAM) 687, PvA, (°sama, 101, (nīl°, (cp., 64;, (nt.), 581, m., Sn, 1064;

- (TrEL) Th, Gr., Sk., Idg., to, pabbata, nt.

- (UK) 12)., 273, 617, 348)., 250;, 251)., 382).

- (other) <b> −saŋvilāpa </b>, <b> −mutta </b>, <smallcaps> vi. </smallcaps>, (mahiyā, <smallcaps> iv. </smallcaps>, cloud\";, <b> Rāhu </b>, <b> abbhā </b>, <b> abbhaŋ, <superscript> 9 </superscript>, marajo </b>, abbhāmutta, valāhaka);, <smallcaps> i. </smallcaps>, <b> abbhāmatta </b>, valāhaka−sikhara, <superscript> s. </superscript>, <smallcaps> ii. </smallcaps>, <b> dhū-, storm−cloud, /><b> −kūṭa </b>, thunder−cloud);, <at> a)fro\\s </at>, <b> −paṭala </b>, <at>o)/mbros</at>, nīla−megha, <superscript>1</superscript>, *ṁbhro, \"dull\";, acchādesi);, mahikā</b>, <b> −ghana </b>

On the Twitter data, the supervised approach achieved passable results. While the numbers look great, the actual segmentations do not. For Twitter 1, too many clusters were generated, for Twitter 2 and 3, the recognition of French words worked somewhat, also recognizing English words as French and French words as English. For Twitter 4, the Polish inclusion was isolated but recognized as "other", together with "strawberries". The recognition of transliterated Amharic worked satisfactorily, yielding 'naw' to the Polish model.

As the number of language models increases, so does the risk of misclassification. As can be seen, we already have quite some misclassification with only 15 language models. For example, in our data, the English preposition 'to' is often erroneously classified as 'transliterated Greek'. The Greek particle το 'to' can be either the neuter singular accusative or nominative definite article 'the', the masculine singular accusative or nominative definite article 'the' or the 3rd person neuter singular nominative/accusative weak pronoun 'it', and as such is rather frequent in the language. This is especially problematic with the transliterated Greek language model, which tends to misclassify the English preposition 'to' as transliterated Greek.

A quick corpus study using the Corpus of Modern Greek[8] and the Corpus of Contemporary American English[9] reveals that the frequency per million words for the Greek particle το is 22666, while the English preposition 'to' has a frequency per million words of 25193. Their relative frequencies are very close together, and it might just have happened that the training data used in this work contained more Greek 'to's than English 'to's, leading to this misclassification.

Other reasons for misclassification include relatedness of the modeled languages as in the case of Germanic or Romance language families. Also, the text types used for training and the text types used for testing play an important role, as well as the amount of training data.

For n-gram language models, the quality of the model is dependent on the texts used for training and the texts used in evaluation. It is probable that a different training set would have yielded different results. This is also the problem with the supervised approach; it is necessary to have language data for training and the trained models reflect the training data to some extent.

## 6.2 Textcat

Textcat works well on monolingual texts. However, it fails on multilingual texts and does not work well on short fragments of text, such as single words. Many of the words are tagged as *unknown*, and if a language has been identified, the language guess often is not correct. Hence, Textcat cannot be used for language segmentation purposes.

Indeed, Textcat fails to exceed the baseline values except for two cases: 'Twitter 3' and 'Twitter 4' yield better values than the baseline values. However, upon closer inspection, it is clear that the numerical index values do not give a reliable picture of the quality of the clustering.

Indeed, while the clustering of 'Twitter 3' is not nonsensical, it is not very good, failing to extract the French insertion 'breuvages'. The Rand Index also only shows a slightly better value than the baseline values. It seems that the outstanding score for 'Twitter 4' is achieved because both the clustering by Textcat and the gold standard have the same number of clusters.

Tables 20 and 21 show the clusterings side by side. Clearly, Textcat performed poorly despite the high numerical index values. A closer inspection of all the Textcat results shows that Textcat performs poorly at the task of language segmentation; often, a word cannot be assigned a language and thus is added to the cluster of 'unknown' language words. For the words where a language has been identified, it most often is not the correct language. While language identification is not necessary for the task of language segmentation, it helps to understand why Textcat failed at the task of language segmentation.

|  | Textcat | Gold standard |
|---|---|---|
| Cluster 1 | ∅ | breuvages |
| Cluster 2 | #bilingualism | #FWWC2015, #bilingualism |
| Cluster 3 | Food, and, breuvages, in, Edmonton, are, ready, to, go, just, waiting, for, the, fans, #FWWC2015 | Food, and, in, Edmonton, are, ready, to, go, just, waiting, for, the, fans |

Table 20: 'Twitter 3': Textcat versus Gold clustering

|  | Textcat | Gold standard |
|---|---|---|
| Cluster 1 | strawberries, | żubrówka |
| Cluster 2 | my, dad, comes, back, from, poland, with, two, crates, of, żubrówka, and, adidas, jackets, omg | my, dad, comes, back, from, poland, with, two, crates, of, strawberries, and, adidas, jackets, omg |

Table 21: 'Twitter 4': Textcat versus Gold clustering

## 6.3 Clustering

The clustering results are more difficult to interpret. Often, the first distinction made seems to be based on case, i.e. words that begin with a capital letter versus words that are all lowercase letters. The second run on the 'mixed script: English – Greek' data shows that the first cluster from the first run has been separated into a cluster with words that begin with a capital letter and two clusters with words that don't begin with a capital letter.

**English–Greek: First run: First cluster**

- "intimate, "without, Although, Aquinas, Christians, Corinthians, Socrates, Symposium, Testament, Whether, affection, ancient, another.", appreciation, aspires, attraction, attraction.", becomes, benevolence., biblical, brotherly, chapter,", charity;, children, children., contemplation, content, continues, contributes, definition:, described, existence;, explained, express, feeling, feelings, finding, further, holding, initially, inspired, knowledge, marriage., necessary, non-corporeal, passage, passion.", philosophers, physical, platonic, refined, relationships, returned, self-benefit)., sensually, spiritual, subject, suggesting, through, throughout, transcendence., unconditional, understanding, without, youthful

**English–Greek: Second run: Splitting of first cluster**

- affection, ancient, another.", aspires, becomes, biblical, chapter,", charity;, children, children., content, definition:, feeling, feelings, finding, holding, marriage., necessary, passage, passion.", platonic, refined, returned, subject, through, without

- Although, Aquinas, Christians, Corinthians, Socrates, Symposium, Testament, Whether

- "intimate, appreciation, attraction, attraction.", benevolence., brotherly, contemplation, continues, contributes, described, existence;, explained, express, further, initially, inspired, knowledge, non-corporeal, philosophers, physical, relationships, self-benefit)., sensually, spiritual, suggesting, throughout, transcendence., unconditional, understanding, youthful

Another important distinction seems to be the length of words. Indeed, the results often show clusters that clearly are based on the length of the contained words. The first run on the 'latin script: German – Italian' data shows that short words have been singled out into the first cluster.

**Italian–German: First run: First cluster**

- (il, E, So, a, ad, da, di, e, es, ha, i, il, in, la, le, lo, ma, ne, se, si, un, va, zu

The clustering works well when the scripts involved are dissimilar, as in the case of the English–Chinese text, where the Chinese characters were isolated after the first run, and also the English–Spanish–Arabic example, where the Arabic part was completely isolated in the first run.

The closer the scripts become, the less well clear cut the results are. For Greek–Russian, the results are acceptable, with one mixed cluster. However, the number of clusters is too high for the number of languages involved and the separation is only achieved after two consecutive clusterings.

The clustering of closer scripts, such as Ukrainian–Russian does not work well. The clusters, with the exception of the cluster containing the datum '9—13' are all impure, consisting of Ukrainian and Russian words. The second run also fails at improving the clustering.

Finally, clustering of latin based scripts does not perform well unless diacritics are involved and the diacritics form the most salient distinction. Word containing letters with diacritics are then generally separated from words containing no diacritics, as in the German–Finnish-Turkish example. The first run generates a cluster for numbers, two clusters with diacritics and one cluster without diacritics.

Probably for this reason, the clustering of Transliterated Greek–English and Greek–English worked surprisingly well. In both cases, the first run managed to separate the (transliterated) Greek parts from the English words. However, unaccented Greek words such as *Agape*, *erotas* or *eros* were clustered with English.

**English–Transliterated Greek: First run: Transliterated Greek cluster**

- agápe, philía, storgē., éros

**English–Greek: First run: Greek cluster**

- (ἀγάπη, (ἔρως, Agápe, agápē), Éros, érōs), –

The problem is that when there are other salient distinguishing features besides diacritics, the result is less good, as can be seen on the Pali data.

**Pali: abhijjhitar: Second run**

- abhijjhita, abhijjhātar, covets, function], med., one, who, °itar), °itar, °ātar).
- (T, A, M
- =, l., v.
- <smallcaps> i. </smallcaps>, <smallcaps> v. </smallcaps>, ag., fr., in
- 265, 287
- [n.

In some cases, the clustering fails at the task of language segmentation, as in the case of the various English–French texts and the English–German example with the German inclusion. We can thus say that the surface structure or morphology, or in other words the basis from which we can extract features, is not sufficient to deduce relevant information about 'language'.

When there are more than two languages that are to be separated, the clustering also does not work well. Indeed, the most dissimilar objects are separated first. In the case of English–Spanish–Arabic, the Arabic part is separated first, as well as words with diacritics, while English and Spanish words without diacritics are thrown together. Subsequent runs show no improvement of the clustering concerning the separation of English and Spanish.

In the case of German–Finnish–Turkish, the clustering algorithm seems to cluster out Turkish first, followed by Finnish. The results are however much less clear-cut than for English–Spanish–Arabic.

## 6.4 Language model induction

The language model induction does not seem to work very well on the Latin script data. There are almost only impure clusters, containing more than one language. However, the approach consistently outperforms the clustering approach when we look at the F5 score. For the English–French data set, the clustering approach even outperforms the n-gram language model approach. Indeed, the French words are relatively well separated from the English text, with the exception of 'sucré', which is still thrown together with English words.

**Latin script: English–French**

- both, "soft", in, English, although, their, is, is, the, opposite, of, "rough", or, is, the, opposite, of, sweet, only, for, wines, (otherwise, is

- mou, :, mou, but

- doux,

- Doux, (rugueux), Doux

- while

- "hard"., used).,

- translate, as, meaning, very, different., "coarse", can, also, mean, almost,sucré,

In contrast, the approach works well on the mixed script data. Indeed, we achieve a good separation of the languages by script. However, when there are also Latin based scripts, we encounter the same problems as mentioned above with rather modest results. For example, for the English–Greek text, the approach separates out the Greek character words but it fails to separate transliterated Greek and English. Also, for the English–Spanish–Arabic text, Arabic is separated out, but English and Spanish are not separated well.

One interesting observation can be made in the case of the English–Chinese text. The Chinese characters have been isolated, but the Pinyin transcription is thrown together with the Chinese characters. Based on the prior observations, this is rather unexpected. This raises the question of whether Pinyin ought to be clustered out, or clustered together with English or Chinese.

Again, the language model induction approach outperforms the clustering approach, and also the n-gram language model approach in the case of the English–Greek text.

On the larger Pali dictionary entries, the language model induction approach yields acceptable results. On the shorter Pali dictionary entries, the language model induction approach yields good results.

The quite low performance must be blamed on the data. Indeed, the Pali dictionary data contain various problematic characters such as 'comma/dot and whitespace' as one character. On such characters, whitespace tokenization fails, yielding big chunks of nonsense tokens. For example, the fourth Pali dictionary entry was split into five chunks (while it might not be displayed as such, all commata and all dots are in fact not followed by whitespace, the whitespace is part of the character,[10] hence whitespace tokenization fails).

**Pali: gūhanā: Chunks**

- Gūhanā，（f.）

- [abstr. fr. gūhati]=gūhanā

- （q. v.）

- Pug. 19. Cp. pari°.（Page

- 253）

Furthermore, the data contains markup, abbreviations, references, typing mistakes and signs such as <-> that are difficult to assign to a language.

On the Twitter data, the language model induction approach works rather well. For example, on the first text, separation is not perfect with the Greek cluster still containing some English words.

**Twitter 1: English–Greek**

- BUSINESS, EXCELLENCE.

- Μόλις, ψήφισα, αυτή, τη, λύση, <span style="color:red">Internet, of</span>, στο, διαγωνισμό

- Things, IT

For the third and fourth text, the approach manages to single out the other-language inclusions, but not exclusively. Both times, there is one additional item in the cluster (the relevant clusters are marked in red).

---

[10]The comma has the Unicode codepoint U+FF0C (FULLWIDTH COMMA) and the dot has the Unicode codepoint U+FF0E (FULLWIDTH FULL STOP)

**Twitter 3: French–English**

- #FWWC2015

- <span style="color:red">breuvages, go</span>

- Food, Edmonton, to, for, the

- in, waiting, #bilingualism

- and, are, ready, just, fans

**Twitter 4: English–Polish**

- comes, from, with, two, crates, of, strawberries, jackets, omg

- my, dad, poland, and, adidas

- <span style="color:red">back, żubrówka</span>

The approach exceeded expectations on the second and fifth Twitter text. On the second text, the 'French' cluster does not only contain the French words 'Demain' and 'par', but also the French way of notating time '18h'.

**Twitter 2: French–English**

- Keynote, "The, collective, of, science-publish, or, perish;, it, all, that, counts?"

- Demain, 18h, par

- #dhiha6, David

- @dhiparis, dynamics, is

On the fifth text, an almost perfect result was achieved, with only one additional subdivision of the 'English' cluster.

**Twitter 5: Transliterated Amharic–English**

- (coffee

- bread). is, our

- Buna, dabo, naw

It seems that the language model approach does not work very well on longer texts, especially on longer texts in Latin-based scripts, with the chosen parameter set; still, the approach outperforms the clustering approach and achieves scores in the vicinity of the scores achieved with the supervised trained n-gram language model approach. On mixed script texts, the approach consistently outperforms the clustering approach and we also reach scores in the vicinity of the scores achieved with the supervised trained n-gram language model approach.

Moreover, on short texts, the approach works rather well. We succeed in outperforming the supervised trained n-gram language model approach on a number of texts, and we achieve scores close to the scores achieved with the supervised trained n-gram language model approach.

Although the language model induction approach tends to generate too many clusters, it also generally succeeds at separating the languages involved.

## 6.5   Scores

Of the scores I used for evaluation purposes, it seems that a combination of a high Rand Index and a high F5 score indicate a good language segmentation. A high F5 score alone is not significant. For example, the clustering algorithm achieves an F5 score of 0.7215 on 'Twitter 3'. This score looks good, but the Rand Index score is at 0.4571, and the segmentation is not good.

**Twitter 3: Cluster analysis**

- Edmonton, Food

- go, in, to

- and, are, breuvages, fans, for, just, ready, the, waiting

Similarly, a high Rand Index score alone is not significant. For example, the clustering algorithm achieves a Rand Index score of 0.6738 on the 'Pali 2' text, but the F5 score is at 0.3825 and the clustering is not good.

**Pali 2: Cluster analysis**

- abhijjhita, abhijjhātar, covets, function], med., one, who, °itar), °itar, °ātar).

- (T., <smallcaps>i.</smallcaps>, <smallcaps>v.</smallcaps>, =, A, M, ag., fr., in, l., v.

- 265, 287

- [n.

# 7 Conclusion

In this thesis, I have asked the question of whether unsupervised approaches to language segmentation perform better on short and difficult texts than supervised approaches by overcoming some of the difficulties associated with supervised approaches, such as the need for (enough and adequate)[11] training data, the language-specificity of the language model or the inflexibility of trained language models when it comes to spelling variation and abbreviations, unless the training data also contained spelling variation and abbreviations.

I have given an overview over related work, presenting supervised approaches that have been used in monolingual language identification and the amelioration of such approaches through unsupervised approaches such as clustering.

Unfortunately, the body of literature covering the topic of language segmentation is sparse. The work by Yin et al. (2007) and the work by Seldin et al. (2001) are closest in topic to this thesis. However, Yin et al. (2007) concern themselves with spoken language, with requires a different approach than dealing with written language. As I concentrated on written language, their work was not conducive to this thesis.

In contrast, Seldin et al. (2001) present a work that looks promising. They present a system that finds language borders in a text with great accuracy using unsupervised algorithms. However, they restrict their algorithm in such a way that switching language models after each word is disallowed. Thus, they are unable to detect single-word inclusions and cannot handle situations where the language switches every word, as has been shown to occur in the test data used in section 4.

Another major drawback of the approach is that it also needs longer fragments of monolingual text and an overall longer text. Hence, their approach would not work well on short texts, if at all.

Next, I have presented the theoretical foundations of a supervised n-gram language model approach and an unsupervised clustering approach. Finally, I have introduced a weakly supervised n-gram language model inducing approach devised by myself. All of these approaches can be used for language segmentation. In order to test how well the different approaches perform on different text types, I have performed experiments.

Section 4 presents the experiments made. I have first compiled a small corpus of texts ranging from longer texts with clearly separated languages to one-sentence Twitter messages containing foreign language inclusions. I have also included a set of dictionary entries from the Pali dictionary by the Pali Text Society. Indeed, these entries contain a lot of different languages and abbreviations, and (unfortunately) are not consistently formatted.

I have then presented my implementations of the supervised and weakly super-vised approaches and the choice of the unsupervised clustering algorithms. Then, I have presented the results of their application to the data.

---

[11]The question of what is to be considered 'enough' or 'adequate' is another point of contention; the data always influences the resulting models.

It can be said that the supervised approach works reasonably well. The drawbacks are that the approach needs training data to train the models on. The problems of the training data and its influence on the models have been raised more than once.

The supervised approach failed for non-whitespace scripts. The models would have to be adapted for non-whitespace scripts, introducing more complexity. Also, the training and test texts would have to be split in meaningful ways, introducing the need for a vast array of language-specific text splitters, should the approach work on a wide range of languages.

The unsupervised approach generally succeeded in separating languages by script when different scripts were involved. Other than that, it seems that the chosen morphological features, or possibly morphological features in general, are insufficient for the algorithm to separate languages effectively.

The weakly supervised approach worked well on short texts and on difficult short texts, but less well on long texts, while still outperforming the clustering approach on long texts. The approach consistently outperforms the clustering approach and reaches scores in the vicinity of the scores achieved by the supervised approach, even surpassing the supervised approach in some cases. These results are promising, but more thorough investigations have to be undertaken.

In conclusion, it can be said that some unsupervised (or weakly supervised) approaches can perform better on the task of language segmentation on difficult and short texts. The presented weakly supervised approach does not only outperform the unsupervised clustering approach, it also achieves scores comparable to the scores achieved with the supervised approach.

Future work could concentrate on the reduction of the number of generated clusters, ideally getting down to one cluster per language; it would also be thinkable to prevent overly frequent language model switching by taking a word's context into account. Finally, the parameters could conceivably be adapted automatically. With an increased interest in the area of multilingual text processing lately, the emergence and evolution of the texts themselves will influence the direction of the work in that direction.

> "Il est venu le temps des cathédrales
>     le monde est entré
> dans un nouveau millénaire
>
> L'homme a voulu monter vers les étoiles
>     écrire son histoire
> dans le verre ou dans la pierre"
>
>     — Gringoire

# 8 Appendix

## 8.1 Development data

### 8.1.1 Latin script data

Karl Marx anses som en af de fire klassiske sociologer. Marx er epokegørende for den historiske videnskab. Og Marx spillede en vigtig rolle for den samtidige og efterfølgende arbejderbevægelse.

1891, nach einer Tuberkuloseerkrankung Hopes, eröffnete das Ehepaar ein modernes Lungensanatorium in Nordrach im Schwarzwald, das sie bis 1893 gemeinsam führten. 1895 wurde die Ehe geschieden.

*Sources:*
https://da.wikipedia.org/wiki/Karl_Marx
https://de.wikipedia.org/wiki/Hope_Bridges_Adams_Lehmann

### 8.1.2 Mixed script data

Capitalism is an economic system and a mode of production in which trade, industries, and the means of production are largely or entirely privately owned. Private firms and proprietorships usually operate in order to generate profit, but may operate as private nonprofit organizations.

او را ولودیا خطاب می‌کردند که مخفف ولادمیر است نام اصلی او ولادمیر ایلیچ اولیانوف بود ولی در دنیا به اسم لنین مشهور شد. ولودیا سومین فرزند از شش فرزند خانواده اولیانوف بود که در سال ۱۸۷۰ یعنی یک سال قبل از کمون پاریس، در یک خانواده مرفه در سیمبیرسک در ساحل رود ولگا که در آن زمان شهرکی بیش نبود ولی بعدها به صورت شهر بزرگی به نام اولیاء نوفسک در آمد متولد گردید. پدرش یک خرده بورژوای لیبرال و معلم ریاضی و مادرش دختر یک پزشک المانی بود وبه همین جهت لنین در تمام مدت عمر به المانیها و طرز تفکر المانی که مارکس مولود آن بود به دیده اغماض می‌نگریست. ولودیا در دبیرستان شاگرد خوبی بود و قوه استدلال درخشانی داشت ولی در عین حال بچه‌ای موذی بود.

*Sources:*
https://en.wikipedia.org/wiki/Capitalism
https://fa.wikipedia.org/wiki/ولادیمیر_لنین

### 8.1.3 Twitter data

**Twitter 1**  »Fallo ergo sum«: On being wrong.
*Source:*
Roland Hieber (daniel_bohrer). "»Fallo ergo sum«: On being wrong.". 26 July 2015, 16:47. Tweet.

**Twitter 2**   Music for Airports > le piano en libre-accès dans l'aéroport Charles-de-Gaulles
   *Source:*
Yannick Rochat (yrochat). "Music for Airports > le piano en libre-accès dans l'aéroport Charles-de-Gaulles". 26 July 2015, 18:12. Tweet.

### 8.1.4   Pali dictionary data

All entries have been taken from the Pali Text Society's Pali-English dictionary (T. W. Rhys Davids, William Stede, editors, The Pali Text Society's Pali–English dictionary. Chipstead: Pali Text Society, 1921–5). 8 parts [738 pp.].)

**Hambho**   Hambho，（indecl.）[haṁ+bho] a particle expressing surprise or haughtiness J.I，184，494．See also ambho.（Page 729）

**Ussada**   Ussada，[most likely to ud + syad；see ussanna]：this word is beset with difficulties，the phrase satt-ussada is applied in all kinds of meanings，evidently the result of an original application & meaning having become obliterated．satt° is taken as *sapta（seven）as well as *sattva（being），ussada as prominence，protuberance，fulness，arrogance．The meanings may be tabulated as follows：（1）prominence（cp. Sk．utsedha），used in characterisation of the Nirayas，as "projecting，prominent hells"，ussadanirayā（but see also below 4）J．I，174；IV，3，422（pallaṅkaṁ，v．l．caturassaṁ，with four corners）；V，266．– adj．prominent ThA．13（tejussadehi ariyamaggadhammehi，or as below 4？）．– 2．protuberance，bump，swelling J．IV，188；also in phrase sattussada having 7 protuberances，a qualification of the Mahāpurisa D．III，151（viz．on both hands，feet，shoulders，and on his back）．– 3．rubbing in，anointing，ointment；adj．anointed with（-°），in candan° J．III，139；IV，60；Th．1，267；Vv 537；DhA．I，28；VvA．237．– 4．a crowd adj．full of（-°）in phrase sattussada crowded with（human beings）D．I，87（cp．DA．I，245；aneka-satta-samākiṇṇa；but in same sense BSk．sapt-otsada Divy 620，621）；Pv IV．18（of Niraya = full of beings，expld．by sattehi ussanna uparûpari nicita PvA．221．– 5．qualification，characteristic，mark，attribute，in catussada "having the four qualifications（of a good village）" J．IV，309（viz．plenty of people，corn，wood and water C．）．The phrase is evidently shaped after D．I，87（under 4）．As "preponderant quality，characteristic" we find ussada used at Vism．103（cf．Asl．267）in combns．lobh°，dos°，moh°，alobh° etc．（quoted from the "Ussadakittana"），and similarly at VvA．19 in Dhammapāla's definition of manussa（lobh'ādīhi alobh'ādīhi sahitassa manassa ussannatāya manussā），viz．sattā manussa-jātikā tesu lobh'‹-› ādayo alobh'ādayo ca ussadā．– 6．（metaph.）self-elevation，arrogance，conceit，haughtiness Vin．I，3；Sn．515，624（an° = taṇhā-ussada-abhāvena SnA 467），783

（expld. by Nd1 72 under formula sattussada；i. e. showing 7 bad qualities，viz. rāga，dosa，moha etc. )，855. – See also ussādana，ussādeti etc.（Page 157）

## 8.2 Test data

### 8.2.1 Latin script data

**English - German**   The German word Nabelschau means "navel-gazing" or "staring at your navel". But in this case, it doesn't refer to anyone else's belly button – just your own.

*Source:*

Glass, Nicole (2015): "German Missions in the United States - Word of the Week". Germany.info.

**English - French**   doux, mou : both translate as "soft" in English, although their meaning is very different. Doux is the opposite of "rough" or "coarse" (rugueux), while mou is the opposite of "hard". Doux can also mean sweet, but almost only for wines (otherwise sucré is used).

*Source:*

Maciamo, (2015): "French words and nuances that don't exist in English". Eupedia.

**English - Transliterated Greek**   The Greek language distinguishes at least four different ways as to how the word love is used. Ancient Greek has four distinct words for love: agápe, éros, philía, and storgē. However, as with other languages, it has been historically difficult to separate the meanings of these words when used outside of their respective contexts. Nonetheless, the senses in which these words were generally used are as follows.

*Source:*

https://en.wikipedia.org/wiki/Greek_words_for_love

**Italian - German**   Milano ne custodisce l'esempio più struggente: quel Cenacolo che il vinciano affrescò con amore, cura e rivoluzionaria psicologia (il Giuda non viene privato dell'aureola, ma si condanna da solo, con la consapevolezza del peccato) cominciò subito ad autodistruggersi, con un cancro che solo un lunghissimo restauro ha di recente arginato.

Kaum eine Woche vergeht, in der es keine neue Studie, Umfrage oder Warnung zum Thema Fachkräftemangel in Deutschland gibt.

Certo, lo faceva per definire le idee, ma anche perché consapevole che le intuizioni sono periture, che la vita stessa va catturata in qualche modo.

Dabei mehren sich letzter Zeit auch Stimmen, die Entwarnung geben. So kam jüngst eine Studie des Stifterverbands für die Deutsche Wissenschaft zu dem Ergebnis, dass "ein allgemeiner Fachkräftemangel in den MINT-Berufen eher nicht mehr" drohe.

Come anche i riccioli del Battista richiamano il movimento delle acque, moto che poi Leonardo studierà più approfonditamente a Venezia, nelle ricerche sui bacini in chiave di difesa anti-Turchi. E si vada alla bellissima Annunciazione, con un occhio attento alle ali dell'angelo: la delicatezza delle punte all'insù che cosa sono se non il barbaglio di un sogno che lo ossessionava da anni, ovvero quello di volare?

Ist das seit Jahren angemahnte Szenario vom drohenden Fachkräftemangel bei Ingenieuren und Naturwissenschaftlern also nur ein Mythos?

*Source:*

Stalinski, Sandra (2015): "Ingenieure: Mythos Fachkräftemangel?". tagesschau.de.
Scorranese, Roberta (2015): "Nelle grandi opere il racconto sofferto della natura mortale". Archiviostorico.corriere.it.

**German - Finnish - Turkish**  Der Sommer ist die wärmste der vier Jahreszeiten in der gemäßigten und arktischen Klimazone. Je nachdem, ob er gerade auf der Nord- oder Südhalbkugel herrscht, spricht man vom Nord- oder Südsommer. Der Nordsommer findet gleichzeitig mit dem Südwinter statt.

Kesä eli suvi on vuodenaika kevään ja syksyn välissä. Kesä on vuodenajoista lämpimin, koska maapallo on silloin kallistunut niin, että aurinko säteilee maan pinnalle jyrkemmässä kulmassa kuin muina vuodenaikoina. Pohjoisella pallonpuoliskolla kesäkuukausiksi lasketaan tavallisesti kesä-. heinä- ja elokuu, eteläisellä pallonpuoliskolla joulu-, tammi- ja helmikuu.

Yaz, en sıcak mevsimdir. Kuzey Yarım Küre'de en uzun günler yazda gerçekleşir. Dünya ısıyı depo ettiği için en sıcak günler genellikle yaklaşık iki ay sonra ortaya çıkar. Sıcak günler Kuzey Yarım Küre'de 21 Haziran ile 22 Eylül arasında, Güney Yarım Küre'de ise 22 Aralık ile 21 Mart arasındadır.

*Source:*

https://fi.wikipedia.org/wiki/Kesä
https://de.wikipedia.org/wiki/Sommer
https://tr.wikipedia.org/wiki/Yaz

### 8.2.2  Mixed script data

**Greek - Russian**  Η ελληνική γλώσσα είναι μία από τις ινδοευρωπαϊκές γλώσσες. Αποτελεί το μοναδικό μέλος ενός ανεξάρτητου κλάδου της ινδοευρωπαϊκής οικογένειας γλωσσών. Ανήκει επίσης στον βαλκανικό γλωσσικό δεσμό. Στην ελληνική γλώσσα, έχουμε γραπτά κείμενα από τον 15ο αιώνα π.Χ. μέχρι σήμερα.

На греческом языке на всех этапах его существования была создана богатейшая литература. В Римской империи знание греческого языка считалось обяза-тельным для всякого образованного человека. В латинском языке присутствует большое количество греческих заимствований, а в греческом —значительное количество латинских и романских слов. В новое время древнегреческий язык стал (наряду с латинским) источником создания новых научных и технических терминов (так называемая международная лексика). В русский язык греческие слова проникали в основном двумя путями —через международную лексику и через церковнославянский язык.

**English - Greek - Transliterated Greek** Agápe (ἀγάπη agápē) means "love: esp. brotherly love, charity; the love of God for man and of man for God." Agape is used in the biblical passage known as the "love chapter," 1 Corinthians 13, and is described there and throughout the New Testament as brotherly love, affection, good will, love, and benevolence. Whether the love given is returned or not, the person continues to love (even without any self-benefit). Agape is also used in ancient texts to denote feelings for one's children and the feelings for a spouse, and it was also used to refer to a love feast. It can also be described as the feeling of being content or holding one in high regard. Agape is used by Christians to express the unconditional love of God for his children. This type of love was further explained by Thomas Aquinas as "to will the good of another."

Éros (ἔρως érōs) means "love, mostly of the sexual passion." The Modern Greek word "erotas" means "intimate love." It can also apply to dating relationships as well as marriage. Plato refined his own definition: Although eros is initially felt for a person, with contemplation it becomes an appreciation of the beauty within that person, or even becomes appreciation of beauty itself. Plato does not talk of physical attraction as a necessary part of love, hence the use of the word platonic to mean, "without physical attraction."

In the Symposium, the most famous ancient work on the subject, Plato has Socrates argue that eros helps the soul recall knowledge of beauty, and contributes to an understanding of spiritual truth, the ideal "Form" of youthful beauty that leads us humans to feel erotic desire – thus suggesting that even that sensually based love aspires to the non-corporeal, spiritual plane of existence; that is, finding its truth, just like finding any truth, leads to transcendence. Lovers and philosophers are all inspired to seek truth through the means of eros.

**English - Spanish - Arabic**   A black ribbon is a symbol of remembrance or mourn-ing. Wearing or displaying a black ribbon has been used for POW/MIA remembrance, mourning tragedies or as a political statement.

El crespón negro o lazo negro es un símbolo utilizado por personas, estados, so-ciedades y organizaciones, representando un sentimiento político-social en señal de duelo.

الرمز يعني الرسم الذي يعبر عن شيء معين وعموما فأن العلامة ينبغي أن تنقل رسالتها بنظرة واحدة دون الحاجة لاية كلمات و من المعروف أن قدماء المصريين والأغريق أستخدموا العلامات ولكن أكثر من استخدم العلامات هم

*Source:*

https://es.wikipedia.org/?title=Lazo_negro
https://en.wikipedia.org/wiki/Black_ribbon
https://ar.wikipedia.org/wiki/رمز

**English - Chinese - (Pinyin)**   The Chinese word for "crisis" (simplified Chinese: 危机; traditional Chinese: 危機; pinyin: wēijī) is frequently invoked in Western motivational speaking because the word is composed of two Chinese characters that can represent "danger" and "opportunity". Some linguists have criticized this usage because the component pronounced jī (simplified Chinese: 机; traditional Chinese: 機) has other meanings besides "opportunity". In Chinese tradition, certain numbers are believed by some to be auspicious (吉利) or inauspicious (不利) based on the Chinese word that the number name sounds similar to. The numbers 0, 6, 8, and 9 are believed to have auspicious meanings because their names sound similar to words that have positive meanings.
*Source:*
https://en.wikipedia.org/w/index.php?title=Chinese_word_for_"crisis"

**Ukrainian - Russian**   Віддавна на території України існували держави скіфів, сарматів, готів та інших народів, але відправним пунктом української державності й культури вважається Київська Русь 9—13 століття.
На юге омывается водами Чёрного и Азовского морей. Имеет сухопутную границу с Россией, Белоруссией, Польшей, Словакией, Венгрией, Румынией и Молдавией.
*Source:*
https://uk.wikipedia.org/wiki/Україна
Surgut-safari.ru, (2015): "Страны - Safari Tour".

### 8.2.3 Twitter data

**Tweet 1: Greek – English**   Μόλις ψήφισα αυτή τη λύση  Internet of Things, στο διαγωνισμό  BUSINESS IT EXCELLENCE.
*Source:*
GaloTyri. "Μόλις ψήφισα αυτή τη λύση Internet of Things, στο διαγωνισμό BUSINESS IT EXCELLENCE.". 19 June 2015, 12:06. Tweet

**Tweet 2: English – French**   Demain #dhiha6 Keynote 18h @dhiparis "The collective dynamics of science-publish or perish; is it all that counts?" par David @chavalarias
*Source:*
Claudine Moulin (ClaudineMoulin). "Demain #dhiha6 Keynote 18h @dhiparis "The collective dynamics of science-publish or perish; is it all that counts?" par David @chavalarias". 10 June 2015, 17:35. Tweet.

**Tweet 3: English – French**   Food and breuvages in Edmonton are ready to go, just waiting for the fans #FWWC2015 #bilingualism
*Source:*
HBS (HBS_Tweets). "Food and breuvages in Edmonton are ready to go, just waiting for the fans #FWWC2015 #bilingualism". 6 June 2015, 23:29. Tweet.

**Tweet 4: English – Polish**   my dad comes back from poland with two crates of strawberries, żubrówka and adidas jackets omg
*Source:*
katarzyne (wifeyriddim). "my dad comes back from poland with two crates of strawberries, żubrówka and adidas jackets omg". 8 June 2015, 08:49. Tweet.

**Tweet 5: Transliterated Amharic – English**   Buna dabo naw (coffee is our bread).
*Source:*
TheCodeswitcher. "Buna dabo naw (coffee is our bread).". 9 June 2015, 02:12. Tweet.

### 8.2.4 Pali dictionary data

All entries have been taken from the Pali Text Society's Pali-English dictionary (T. W. Rhys Davids, William Stede, editors, The Pali Text Society's Pali–English dictionary. Chipstead: Pali Text Society, 1921–5. 8 parts [738 pp.].)

**abbha**   (nt.) [Vedic abhra nt. & later Sk. abhra m. "dark cloud"; Idg. *m̐bhro, cp. Gr. <at>a)fro\\s</at> scum, froth, Lat. imber rain; also Sk. ambha water, Gr. <at>o)/mbros</at> rain, Oir ambu water]. A (dense & dark) cloud, a cloudy mass A <smallcaps>ii.</smallcaps> 53 = Vin <smallcaps>ii.</smallcaps> 295 = Miln 273 in list of to things that obscure moon– & sunshine, viz. **abbhaŋ mahikā** (mahiyā A) **dhū- marajo** (megho Miln), **Rāhu** . This list is referred to at SnA 487 & VvA 134. S <smallcaps>i.</smallcaps> 101 (°sama pabbata a mountain like a thunder–cloud); J <smallcaps>vi.</smallcaps> 581 (abbhaŋ rajo acchādesi); Pv <smallcaps>iv.</smallcaps> 3 <superscript>9</superscript> (nīl° = nīla–megha PvA 251). As f. **abbhā** at Dhs 617 & DhsA 317 (used in sense of adj. "dull"; DhsA expl <superscript>s.</superscript> by valāhaka); perhaps also in **abbhāmatta** . <br />**–kūṭa** the point or summit of a storm–cloud Th 1, 1064; J <smallcaps>vi.</smallcaps> 249, 250; Vv 1 <superscript>1</superscript> (= valāhaka–sikhara VvA 12). **–ghana** a mass of clouds, a thick cloud It 64; Sn 348 (cp. SnA 348). **–paṭala** a mass of clouds DhsA 239. **–mutta** free from clouds Sn 687 (also as abbhāmutta Dh 382). **–saŋvilāpa** thundering S <smallcaps>iv.</smallcaps> 289.

**abhijjhitar**   [n. ag. fr. abhijjhita in med. function] one who covets M <smallcaps>i.</smallcaps> 287 (T. abhijjhātar, v. l. °itar) = A <smallcaps>v.</smallcaps> 265 (T. °itar, v. l. °ātar).

**ajja**   Ajja, & Ajjā （adv.）[Vedic adya & adyā, a + dyā, a° being base of demonstr. pron. （see a3）and dyā an old Loc. of dyaus （see diva）, thus "on this day"] to-day, now Sn.75, 153, 158, 970, 998; Dh.326; J.I, 279; III, 425 （read bahutaṁ ajjā; not with Kern, Toev. s. v. as "food"）; Pv.I, 117 （= idāni PvA.59）; PvA.6, 23; Mhvs 15, 64. ‹-› Freq. in phrase ajjatagge （= ajjato + agge （?）or ajja-tagge, see agga3）from this day onward, henceforth Vin.I, 18; D.I, 85; DA.I, 235. –kālaṁ （adv.）this morning J.VI, 180; –divasa the present day Mhvs 32, 23. （Page 10）

**gūhanā**   Gūhanā, （f.）[abstr. fr. gūhati]=gūhanā （q. v.）Pug. 19. Cp. pari°. （Page 253）

**pacati**   Pacati, [Ved. pacati, Idg. *peqŭō, Av. pac-; Obulg. peka to fry, roast, Lith, kepū bake, Gr. péssw cook, pépwn ripe] to cook, boil, roast Vin. IV, 264; fig. torment in purgatory （trs. and intrs. ）: Niraye pacitvā after roasting in N. S. II, 225, PvA. 10, 14. – ppr. pacanto tormenting, Gen. pacato （+Caus. pācayato）D. I, 52 （expld at DA. I, 159, where read pacato for paccato, by pare daṇḍena pīḷentassa）. – pp. pakka （q. v.）. ‹-› Caus. pacāpeti & pāceti （q. v.）. – Pass. paccati to be roasted or tormented （q. v. ）. （Page 382）

## 8.3 Results

### 8.3.1 N-Gram Language Models

For the n-gram language model approach, the identified language is indicated in parentheses. The language abbreviations are:

| Abbreviation | Language |
| --- | --- |
| AR | Arabic |
| DE | German |
| EL | Greek |
| EN | English |
| ES | Spanish |
| FI | Finnish |
| FR | French |
| IT | Italian |
| PL | Polish |
| RU | Russian |
| UK | Ukrainian |
| TR | Turkish |
| TrAM | Transliterated Amharic |
| TrEL | Transliterated Greek |
| ZH | Chinese |

**Data:** Latin script: German – English

- (EN) own., belly, refer, button, But, it, or, your, at, in, "staring, anyone, doesn't, else's, word, this

- (FI) –

- (FR) case, just, means, navel".

- (TrAM) The

- (TrEL) to, German

- (other) Nabelschau, "navel-gazing"

**Data:** Latin script: German – Finnish – Turkish

- (DE) ob, oder, Sommer, und, Nord-, arktischen, der, Der, dem, gemäßigten, mit, er, Südsommer., spricht, Jahreszeiten, Südwinter, herrscht, wärmste, vom, die, statt., nachdem, auf

- (EN) ist, Nordsommer, Mart, in

- (ES) en, depo

- (FI) joulu-, kevään, suvi, on, eli, vuodenajoista, syksyn, koska, kesä-., kuin, Pohjoisella, man, helmikuu., tammi-, lämpimin, heinä-, niin, maapallo, maan, pinnalle, Kesä, säteilee, tavallisesti, vuodenaika, kallistunut, lasketaan, muina, ettiği, jyrkemmässä, elokuu, välissä., että, eteläisellä, silloin, ja, kulmassa

- (FR) vier, Je

- (PL) aurinko

- (RU) 22, 21

- (TR) yaklaşık, ortaya, genellikle, Eylül, Sıcak, çıkar., Yaz, sonra, arasında, Kuzey, Güney, Aralık, gerade, ısıyı, gerçekleşir., Küre'de, günler, için, findet, mevsimdir., arasındadır., Haziran, iki, yazda, uzun, ise, ay, sıcak, ile, Yarım, Dünya

- (TrAM) Der

- (other) Klimazone., gleichzeitig,kesäkuukausiksi, vuodenaikoina., pallonpuoliskolla,Südhalbkugel

**Data:** Latin script: English – French

- (EL) "coarse"

- (EN) but, both, for, while, wines, almost, sweet, of, although, only, is, "rough", used)., or, as, meaning, the, in, translate, "hard"., their, English, also, different., very

- (ES) can

- (FI) mean

- (FR) opposite, Doux, doux, sucré, :

- (RU) "soft"

- (TrEL) mou

- (other) (otherwise, (rugueux)

**Data:**   Latin script: English – Transliterated Greek

- (EN) for, meanings, least, used, been, distinct, love, of, were, are, when, agápe, these, how, and, Greek, word, used., outside, ways, different, other, follows., words, respective, generally, However, is, with, it, at, as, historically, the, in, which, their

- (ES) has, separate

- (FR) language, senses, Ancient, languages, difficult, four

- (IT) contexts.

- (TrAM) éros, The, love:

- (TrEL) to, storgē., philía

- (other) Nonetheless, distinguishes

**Data:**   Latin script: Italian – German

- (DE) drohe., geben., allgemeiner, Studie, jüngst, für, Ergebnis, keine, kam, drohenden, oder, und, letzter, neue, Mythos?, Deutschland, Ist, sich, der, vergeht, studierà, Dabei, Studie, den, dem, auch, Entwarnung, dass, nur, eher, nicht, gibt., Umfrage, Woche, eine, Kaum, Jahren, bei, mehren, Stimmen, Deutsche, das, zum, mehr", angemahnte, "ein, Zeit, ein, So, vom, zu, die, seit, Warnung, Wissenschaft

- (EL) affrescò

- (EN) moto, attento, a, in, ad, also

- (ES) custodisce, cura, subito, Certo, Giuda, lo, del, difesa, con, definire, restauro, se, modo., la, arginato., recente, vada, movimento, Leonardo, Szenario, quel, cominciò

- (FI) va, si, Battista, Thema

- (FR) l'esempio, non, des, acque, perché, un, es, le, sui, condanna

- (IT) solo, faceva, catturata, chiave, peccato), periture, (il, delicatezza, cancro, privato, bellissima, anni, bacini, ovvero, delle, sogno, di, barbaglio, ma, qualche, e, amore, ricerche, Come, per, richiamano, ne, intuizioni, punte, occhio, struggente:, nelle, vita, riccioli, solo, che, volare?, sono, alla, alle, anche, Cenacolo, quello, cosa, ali, viene, il, psicologia, vinciano, Venezia

- (PL) i

- (TR) ha, più, da

- (TrAM) Milano, E

- (TrEL) poi, idee, stessa

- (other) MINT-Berufen, Fachkräftemangel, dell'angelo:, consapevole, anti-Turchi., Annunciazione, lunghissimo, consapevolezza, ossessionava, dell'aureola, approfonditamente, autodistruggersi, rivoluzionaria, Stifterverbands, all' insù, Naturwissenschaftlern, Ingenieuren

**Data:** Mixed script: Greek – Russian

- (EL) κείμενα, βαλκανικό, από, το, αιώνα, Αποτελεί, ελληνική, μία, επίσης, στον, γλωσσικό, γλωσσών., είναι, Στην, έχουμε, μέλος, ανεξάρτητου, τις, γλώσσες., 15ο, Ανήκει, γραπτά, π.Χ., σήμερα., γλώσσα, γλώσσα, κλάδου, οικογένειας, τον, της, δεσμό., μέχρι, μοναδικό, ενός

- (RU) слов., с, богатейшая, образованного, человека., этапах, значительное, знание, научных, лексика)., называемая, технических, источником, стал, латинских, существования, слова, греческом, всех, —, В, романских, новых, Римской, и, проникали, в, греческие, терминов, присутствует, греческих, новое, русский, империи, латинском, литература., создана, создания, путями, основном, язык., язык, (так, его, количество, считалось, обязательным, время, двумя, была, греческого, большое, языке, языка

- (TrAM) Н

- (UK) лексику, (наряду, через, всякого, а, На, для, на

- (other) ινδοευρωπαϊκές, ινδοευρωπαϊκής,латинским), международную, международная, церковнославянский, заимствований, древнегреческий

**Data:** Mixed script: English – Greek

- (DE) Symposium, Modern, being, felt

- (EL) "Form"

- (EN) sensually, platonic, for, holding, existence;, refined, its, explained, attraction, of, (even, are, spiritual, given, refer, Agape, beauty, or, attraction.", like, without, not, further, will, own, love, knowledge, will, one's, most, use, express, This, another.", The, leads, truth, suggesting, dating, relationships, inspired, "love, mostly, hence, definition:, regard., appreciation, a, ideal, us, helps, seek, Agápe, plane, recall, feeling, within, returned, chapter,", based, described, apply, physical, Although, good, by, used, love, God.", children., his, any, charity;, Socrates, be, work, throughout, and, that, Greek, even, word, agápē), love.", known, biblical, feelings, does, famous, In, subject, becomes, one, understanding, children, "love, through, beauty, well, It, was, initially, feast., finding, itself., 13, all, "without, feel, with, is, it, thus, New, as, the, brotherly, in, is, an, there, God, youthful, necessary, high, Lovers, also, Whether

- (ES) person, Aquinas, esp., continues, has, Thomas, truth, can, erotic, sexual, desire

- (FI) on, –, man, mean

- (FR) (ἀγάπη, spouse, not, ancient, marriage., soul, person, content, Christians, Testament, Éros, just, part, type, passage, means, humans, passion.", aspires, contemplation, contributes, argue, affection

- (IT) texts, 1, "intimate, Plato, "to

- (RU) (ἔρως

- (TR) talk

- (TrAM) érōs), "love:

- (TrEL) "erotas", denote, eros., to, eros

- (other) non-corporeal, Corinthians, self-benefit)., benevolence., unconditional, philosophers, transcendence.

**Data:** Mixed script: English – Spanish – Arabic

- (AR) المصريين رسالتها الذي الرسمْ فٱنْ دونْ لايةْ أكثرْ ٱستخدموٱ وْ يبيعْنِ العلامةْ العلاماتْ واحدةْ قدماءْ منْ كلماتْ ولكنْ الرمزْ همْ أنْ شيءْ استخدمْ يعبرْ عنْ بنظرةْ المعروفْ تنقلْ وعموماً والٱغريقْ معينْ الحاجة

- (EN) for, used, been, displaying, of, ribbon, black, or, mourning., statement., tragedies, is, political, a, Wearing, as, mourning

- (ES) por, has, crespón, sociedades, personas, sentimiento, representando, estados, de, El, señal, lazo, símbolo, en, utilizado, y

- (FR) remembrance, remembrance, un, es

- (IT) negro, duelo., POW/MIA

- (TrAM)

- (TrEL) symbol, o

- (other) político-social, organizaciones

**Data:** Mixed script: English – Chinese

- (DE) 机;, Chinese:, Western

- (EL) 機)

- (EN) Some, for, meanings, by, of, are, 8, positive, speaking, be, composed, or, meanings., tradition, number, and, that, sound, linguists, word, some, this, other, In, have, invoked, criticized, 6, because, The, believed, words, numbers, sounds, frequently, is, pronounced, besides, traditional, the, in, represent, two, motivational, usage, their, based

- (ES) 危機;, has, 危机;, can, Chinese, "crisis", similar

- (FI) on

- (FR) (吉利), component, "danger", characters, (不利), certain, jī

- (PL) pinyin:

- (RU) 0, 9, wēijī

- (TrEL) to, to., names, name

- (other) inauspicious, "opportunity"., (simplified, auspicious

**Data:** Mixed script: Ukrainian – Russian

- (RU) Польшей, Румынией, Венгрией, юге, границу, с, омывается, Имеет, 9 —13, Молдавией., Азовского, водами, Россией, Чёрного, Русь, и, пунктом, Словакией

- (TrAM) й

- (UK) держави, скіфів, України, народів, На, державності, вважається, відправним, території, української, готів, культури, але, сарматів, існували, століття., Київська, на, Віддавна, інших, та, морей.

- (other) сухопутную, Белоруссией

**Data:** Pali: abbha

- (AR) ., 134., 289.

- (DE) Miln), imber, dark), Miln

- (EL) (=, (abbhaŋ

- (EN) water, mountain, of, free, (used, or, like, referred, (also, A, This, cloudy, clouds, later, a, froth, 1, summit, thundering, by, mass, Pv, Oir, obscure, scum, that, water]., thick, As, from, It, is, at, as, the, in, clouds, things, also

- (ES) (dense, f., sense, expl, rajo

- (FI) 239., rain;, Lat., Vin, perhaps, SnA

- (FR) cloud, Dh, adj., point, cloud, Dhs, A), rain, VvA, DhsA, list

- (IT) \"dark, &, ambha, 3, 1, 317, J, sunshine, cp., abhra, [Vedic, (megho

- (PL) 487, =, S, 295, <br, moon−, 249

- (RU) 348, 53

- (TR) viz., ambu, Vv

- (TrAM) 687, PvA, (°sama, 101, (nīl°, (cp., 64;, (nt.), 581, m., Sn, 1064;

- (TrEL) Th, Gr., Sk., Idg., to, pabbata, nt.

- (UK) 12)., 273, 617, 348)., 250;, 251)., 382).

- (other) <b> −saŋvilāpa </b>, <b> −mutta </b>, <smallcaps> vi. </smallcaps>, (mahiyā, <smallcaps> iv. </smallcaps>, cloud\";, <b> Rāhu </b>, <b> abbhā </b>, <b> abbhaŋ, <superscript> 9 </superscript>, marajo </b>, abbhāmutta, valāhaka);, <smallcaps> i. </smallcaps>, <b> abbhāmatta </b>, valāhaka−sikhara, <superscript> s. </superscript>, <smallcaps> ii. </smallcaps>, <b> dhū−, storm−cloud, /><b> −kūṭa </b>, thunder−cloud);, <at>a)fro\\s</at>, <b>−paṭala</b>, <at>o)/mbros</at>, nīla−megha, <superscript>1</superscript>, *ṁbhro, \"dull\";, acchādesi);, mahikā</b>, <b> −ghana </b>

**Data:** Pali: abhijjhitar

- (DE) v.
- (EN) A, one, in, who, covets, med., function]
- (IT) ag., M, fr.
- (PL) 287, =
- (RU) 265
- (TrAM) l., [n.
- (TrEL) (T.
- (other) <smallcaps> v. </smallcaps>, abhijjhātar, abhijjhita, °ātar)., <smallcaps> i. </smallcaps>, °itar, °itar)

**Data:** Pali: ajja

- (DE) （see, v., being, Ajjā
- (EN) of, or, and, not, present, Freq., day, this, "on, from, adyā，a, with, as, the, morning, in, day"], an
- (ES) bahutaṁ,
- (FI) 32，23., ajjato
- (FR) Loc., dyaus, 15，64., dyā, pron.
- (IT) [Vedic, Mhvs, &, −divasa
- (PL) （=, +, demonstr., s.
- (RU) III，425, agge（?）
- (TR) old, adya, 10）, idāni
- (TrAM) ‹–›
- (TrEL) phrase, base
- (UK) a3）
- (other) onward，henceforth, ajjā；, DA.I，235.,（adv.）, J.I，279；, D.I，85；, ajja-tagge，see, Sn.75，153，158，970，998；, J.VI，180；, PvA.6，23；, −kālaṁ, diva）, thus, PvA.59）；, agga3）, Kern，Toev., Pv.I，117, Dh.326；, ajjatagge, （read,（Page, Vin.I，18；, dyā, a°, Ajja，&, to-day，now, "food"）；

**Data:** Pali: gūhanā

- (ES) 253）
- (other) [abstr. fr. gūhati]=gūhanā, Pug. 19. Cp. pari°.（Page,（q. v.）, Gūhanā，（f.）

**Data:** Pali: pacati

- (EL) 382）
- (EN) for, after, roasting, read, roasted, be, or, at, tormented, in
- (FR) pare, D. I，52
- (IT) &, pacato, purgatory
- (TrAM) pāceti, ripe]
- (TrEL) to, daṇḍena
- (other) bake，Gr. péssw,（+Caus. pācayato）,（q. v.）.（Page, DA. I, 159，where, Caus. pacāpeti, intrs.）: Niraye, pacitvā, Pass. paccati,（trs. and, tormenting，Gen. pacato, piḷentassa）. –, fig. torment, cook, pépwn, Pacati, [Ved. pacati, Idg. *peqŭō, Av. pac-；, paccato，by, ppr. pacanto, cook，boil，roast, fry，roast，Lith, kepū,（q. v.）. –,（expld, Vin. IV, 264；, Obulg. peka, pp. pakka,（q. v.）. ‹-›, N. S. II, 225，PvA. 10, 14. –

**Data:** Twitter 1 (Greek–English)

- (DE) Internet
- (EL) στο, τη, αυτή, διαγωνισμό, λύση, ψήφισα
- (EN) of, IT, Things
- (ES) BUSINESS
- (TrAM) Μόλις
- (other) EXCELLENCE.

**Data:** Twitter 2 (French–English)

- (EN) David, "The, is, it, perish;, or, collective, Demain, counts?", that, of, dynamics, all
- (FI) 18h
- (FR) par, Keynote
- (other) #dhiha6, @dhiparis, science-publish

**Data:** Twitter 3 (French–English)

- (EN) for, Food, waiting, the, in, ready, and, are
- (ES) go
- (FI) Edmonton
- (FR) just, breuvages, fans
- (TrEL) to
- (other) #bilingualism, #FWWC2015

**Data:** Twitter 4 (English–Polish)

- (EN) with, back, from, comes, crates, and, poland, two, of, jackets
- (ES) dad, adidas
- (TrAM) my
- (TrEL) omg
- (other) żubrówka, strawberries

**Data:** Twitter 5 (Transliterated Amharic–English)

- (EN) is, bread).
- (FR) our
- (IT) (coffee
- (PL) naw
- (TrAM) Buna, dabo

### 8.3.2 Textcat

For Textcat, the identified language is indicated in parentheses. As Textcat returns *unknown* for many words, I merely indicate the non-unknown categories to save space and write **rest** to indicate that all other words of the text have been classified as *unknown*. The language abbreviations are:

| Abbreviation | Language |
|---|---|
| DA | Danish |
| DE | German |
| EL | Greek |
| EN | English |
| ES | Spanish |
| FI | Finnish |
| FR | French |
| HU | Hungarian |
| ID | Indonesian |
| IT | Italian |
| LT | Lithuanian |
| LV | Latvian |
| NL | Dutch |
| PT | Portuguese |
| RU | Russian |
| TH | Thai |
| ZH | Chinese |

**Data:** Latin script: German – English

- (HU) "navel-gazing"

- (ZH) Nabelschau

- (unknown) **rest**

**Data:** Latin script: German – Finnish – Turkish

- (DA) Südsommer., genellikle,

- (DE) Jahreszeiten, arktischen,

- (FI) vuodenajoista, kallistunut, tavallisesti,

- (ZH) gemäßigten, Klimazone., Südhalbkugel, Nordsommer, gleichzeitig, vuoden-aika, jyrkemmässä, vuodenaikoina., Pohjoisella, pallonpuoliskolla, kesäkuukausik-si, eteläisellä, mevsimdir., gerçekleşir., arasındadır.,

- (unknown) **rest**

**Data:** Latin script: English – French

- (HU) different.,

- (ZH) (rugueux),(otherwise,

- (unknown) **rest**

**Data:** Latin script: English – Transliterated Greek

- (EN) historically, respective,

- (LT) languages,

- (ZH) distinguishes, Nonetheless,

- (unknown) **rest**

**Data:** Latin script: Italian – German

- (DE) allgemeiner, angemahnte,

- (ES) delicatezza,

- (HU) bellissima,

- (IT) dell'aureola, consapevole, richiamano, anti-Turchi., ossessionava,

- (NL) Ingenieuren,

- (PT) approfonditamente,

- (ZH) custodisce, struggente:, rivoluzionaria, psicologia, consapevolezza, auto-distruggersi, lunghissimo, Fachkräftemangel, Deutschland, intuizioni, Entwar-nung, Stifterverbands, Wissenschaft, MINT-Berufen, Annunciazione, dell'an-gelo:, Naturwissenschaftlern,

- (unknown) **rest**

**Data:**   Mixed script: Greek – Russian

- (EL) ανεξάρτητου, οικογένειας,

- (RU) существования, богатейшая, литература., греческого, обязательным, образованного, присутствует, количество, заимствований, значительное, источником, технических, называемая, международная,

- (TH) латинским),

- (ZH) ινδοευρωπαϊκές, ινδοευρωπαϊκής, древнегреческий, международную, церковнославянский,

- (unknown) **rest**

**Data:**   Mixed script: English – Greek

- (DA) definition:, understanding,

- (EN) affection, unconditional, suggesting,

- (FR) relationships, contemplation, appreciation, attraction, attraction.", transcendence.,

- (HU) benevolence., self-benefit).,

- (IT) non-corporeal,

- (PT) contributes,

- (ZH) Corinthians, throughout, Christians, Symposium, existence;, philosophers,

- (unknown) **rest**

**Data:**   Mixed script: English – Spanish – Arabic

- (ES) sociedades, organizaciones, sentimiento, político-social,

- (FR) remembrance, remembrance, statement.,

- (ID) displaying,

- (PT) representando,

- (unknown) **rest**

**Data:**    Mixed script: English – Chinese

- (EN) traditional, motivational, pronounced, tradition„
- (FR) characters,
- (ZH) simplified, frequently, "opportunity"., criticized, auspicious, inauspicious,
- (unknown) **rest**

**Data:**    Mixed script: Ukrainian – Russian

- (RU) державності, Словакией, Молдавией.,
- (TH) вважається,
- (ZH) відправним, української, сухопутную, Белоруссией,
- (unknown) **rest**

**Data:**    Pali: abbha

- (DA) storm–cloud, thundering,
- (HU) marajo</b>, nīla–megha, valāhaka–sikhara,
- (ZH)
  <at> a)fro\\</at>, <at> o)/mbros </at>, <smallcaps> ii. </smallcaps>, mahikā</b>, <b> Rāhu </b>, <smallcaps> i. </smallcaps>, thunder–cloud);, <smallcaps> vi. </smallcaps>, acchādesi);, <smallcaps> iv. </smallcaps>, <superscript> 9 </superscript>, <b> abbhā </b>, <superscript> s. </superscript>, valāhaka);, <b> abbhāmatta </b>, /><b> –kūṭa </b>, <superscript> 1 </superscript>, <b> –ghana </b>, <b> –paṭala </b>, <b> –mutta </b>, abbhāmutta, <b> –saŋvilāpa </b>
- (unknown) **rest**

**Data:**    Pali: abhijjhitar

- (ZH) abhijjhita, <smallcaps> i. </smallcaps>, abhijjhātar, <smallcaps> v. </smallcaps>,
- (unknown) **rest**

**Data:**  Pali: ajja

- (ZH) diva）, thus, to-day，now, Sn.75，153，158，970，998；, Kern，Toev., ajja-tagge，see, onward，henceforth,

- (unknown) **rest**

**Data:**  Pali: gūhanā

- (ZH) Gūhanā，（f.）, [abstr. fr. gūhati]hanā, Pug. 19. Cp. pari°.（Page,

- (unknown) **rest**

**Data:**  Pali: pacati

- (ZH) fig. torment, Pacati，[Ved. pacati，Idg. *peqŭō，Av. pac-；, Obulg. peka, fry，roast，Lith，kepū, bake，Gr. péssw, cook，pépwn, cook，boil，roast, Vin. IV，264；, intrs.）: Niraye, N. S. II，225，PvA. 10，14. –, ppr. pacanto, tormenting，Gen. pacato，（+Caus. pācayato）, DA. I, 159, where, paccato, by, pīḷentassa）. –,（q. v.）. ‹–›, Caus. pacāpeti, Pass. paccati,（q. v.）.（Page,

- (unknown) **rest**

**Data:**  Twitter 1 (Greek–English)

- (ZH) διαγωνισμό, EXCELLENCE.,

- (unknown) **rest**

**Data:**  Twitter 2 (French–English)

- (IT) collective,

- (ZH) science-publish,

- (unknown) **rest**

**Data:**  Twitter 3 (French–English)

- (ZH) #bilingualism,

- (unknown) **rest**

**Data:** Twitter 4 (English–Polish)

- (LV) strawberries,
- (unknown) **rest**

**Data:** Twitter 5 (Transliterated Amharic–English)

- (unknown) **rest**

### 8.3.3 Clustering

Clustering the different data sets produced the following clusters. The second run uses the clusters from the first run and possibly subdivides each cluster into two or more clusters.

**Data:** Latin script: German – English

**First run**

- "navel-gazing", doesn't, else's
- "staring, But, German, Nabelschau, anyone, belly, button, case, just, means, navel"., own., refer, this, word, your
- at, in, it, or, to
- –, The

**Second run**

- doesn't, else's
- "navel-gazing"
- "staring, But, German, Nabelschau, belly, case, means, navel"., refer, this
- anyone, button, just, own., word, your
- it, or, to
- at, in
- –, The

**Data:** Latin script: German – Finnish – Turkish

**First run**

- Dünya, Güney, Küre'de, Südhalbkugel, Südsommer., Südwinter, Sıcak, arasında, gemäßigten, günler, için, kesäkuukausiksi, lämpimin, säteilee, sıcak, wärmste, çıkar., Der

- Aralık, Eylül, Kesä, Yarım, arasındadır., eteläisellä, ettiği, että, gerçekleşir., heinä-, jyrkemmässä, kesä-., kevään, välissä., yaklaşık, ısıyı

- 21, 22

- Der, Haziran, Jahreszeiten, Je, Klimazone., Kuzey, Mart, Nord-, Nordsommer, Pohjoisella, Sommer, Yaz, arktischen, auf, aurinko, ay, dem, depo, der, die, eli, elokuu, en, er, findet, genellikle, gerade, gleichzeitig, helmikuu., herrscht, iki, ile, in, ise, ist, ja, joulu-, kallistunut, koska, kuin, kulmassa, lasketaan, maan, maapallo, man, mevsimdir., mit, muina, nachdem, niin, ob, oder, on, ortaya, pallonpuoliskolla, pinnalle, silloin, sonra, spricht, statt., suvi, syksyn, tammi-, tavallisesti, und, uzun, vier, vom, vuodenaika, vuodenaikoina., vuodenajoista, yazda

**Second run**

- Südhalbkugel, Südsommer., Südwinter, arasında, gemäßigten, kesäkuukausiksi, lämpimin, säteilee, wärmste

- Dünya, Güney, Küre'de, Sıcak, günler, için, sıcak, çıkar., Der

- arasındadır., eteläisellä, ettiği, että, gerçekleşir., heinä-, jyrkemmässä, kesä-., kevään, välissä., yaklaşık, ısıyı

- Aralık, Eylül, Yarım

- Kesä

- 22

- 21

- Der, Haziran, Jahreszeiten, Klimazone., Kuzey, Mart, Nord-, Nordsommer, Pohjoisella, Sommer, Yaz,

- arktischen, auf, aurinko, dem, depo, der, die, eli, elokuu, findet, genellikle, gerade, gleichzeitig, helmikuu., herrscht, iki, ile, ise, ist, joulu-, kallistunut, koska, kuin, kulmassa, lasketaan, maan, maapallo, man, mevsimdir., mit, muina, nachdem, niin, oder, ortaya, pallonpuoliskolla, pinnalle, silloin, sonra, spricht, statt., suvi, syksyn, tammi-, tavallisesti, und, uzun, vier, vom, vuodenaika, vuodenaikoina., vuodenajoista, yazda

- Je, ay, en, er, in, ja, ob, on

**Data:**    Latin script: English – French

**First run**

- "coarse", "hard"., "rough", "soft", (otherwise, (rugueux), Doux, English, almost, also, although, both, but, can, different., doux, for, mean, meaning, mou, only, opposite, sucré, sweet, the, their, translate, used)., very, while, wines

- is, or

- as, in, of

**Second run**

- Doux, English,

- "coarse", (otherwise, (rugueux), almost, although, different., meaning, opposite, translate

- "hard"., "rough", "soft", also, both, but, can, doux, for, mean, mou, only, sucré, sweet, the, their, used)., very, while, wines

- or

- is

- in

- of

- as

**Data:** Latin script: English – Transliterated Greek

**First run**

- The

- agápe, philía, storgē., éros,

- Ancient, However, Nonetheless, contexts., different, difficult, distinct, distinguishes, follows., generally, historically, language, languages, meanings, outside, respective, senses, separate, which, words

- Greek, and, are, as, at, been, for, four, has, how, in, is, it, least, love, love:, of, other, the, their, these, to, used, used., ways, were, when, with, word

**Second run**

- The

- philía, storgē.

- agápe, éros,

- Ancient, However, Nonetheless, contexts., different, difficult, distinct, distinguishes, follows., generally, historically, meanings, respective

- words

- language, languages, outside, senses, separate, which

- and, are, as, at, been, for, four, has, how, in, is, it, least, love, love:, of, other, the, their, these, to, used, used., ways, were, when, with, word

- Greek

**Data:** Latin script: German – Italian

**First run**

- (il, E, So, a, ad, da, di, e, es, ha, i, il, in, la, le, lo, ma, ne, se, si, un, va, zu

- "ein , Annunciazione, Battista, Cenacolo, Certo, Come, Dabei, Deutsche, Deutschland, Entwarnung, Ergebnis, Giuda, Ingenieuren, Ist, Jahren, Kaum, Leonardo, MINT-Berufen, Mythos?, Naturwissenschaftlern, Stifterverbands, Stimmen, Studie, Studie, Szenario, Thema, Umfrage, Venezia, Warnung, Wissenschaft, Woche, Zeit, acque, ali, alla, alle, allgemeiner, also, amore, anche, angemahnte, anni, anti-Turchi., approfonditamente, arginato., attento, auch, autodistruggersi, bacini, barbaglio, bei, bellissima, cancro, catturata, che, chiave, con, condanna, consapevole, consapevolezza, cosa, cura, custodisce, das, dass, definire, del, delicatezza, delle, dem, den, der, des, die, difesa, drohe., drohenden, eher, ein, eine, faceva, geben., gibt., idee, intuizioni, kam, keine, letzter, lunghissimo, mehr", mehren, modo., moto, movimento, nelle, neue, nicht, non, nur, occhio, oder, ossessionava, ovvero, peccato), per, periture, poi, privato, psicologia, punte, qualche, quel, quello, recente, restauro, riccioli, ricerche, richiamano, rivoluzionaria, seit, sich, sogno, solo, solo, sono, stessa, struggente:, subito, sui, und, vada, vergeht, viene, vinciano, vita, volare?, vom, zum

- all' insù, dell' angelo:, dell' aureola, l' esempio, Milano

- Fachkräftemangel, affrescò, cominciò, für, jüngst, perché, più, studierà

## Second run

- a, e, i

- E

- So

- (il, ad, da, di, es, ha, il, in, la, le, lo, ma, ne, se, si, un, va, zu

- Annunciazione, Battista, Cenacolo, Certo, Come, Dabei, Deutsche, Deutschland, Entwarnung, Ergebnis, Giuda, Ingenieuren, Ist, Jahren, Kaum, Leonardo, MINT-Berufen, Mythos?, Naturwissenschaftlern, Stifterverbands, Stimmen, Studie, Studie, Szenario, Thema, Umfrage, Venezia, Warnung, Wissenschaft, Woche, Zeit

- "ein, acque, ali, alla, alle, allgemeiner, also, amore, anche, angemahnte, anni, anti-Turchi., approfonditamente, arginato., attento, auch, autodistruggersi, bacini, barbaglio, bei, bellissima, cancro, catturata, che, chiave, con, condanna, consapevole, consapevolezza, cosa, cura, custodisce, das, dass, definire, del, delicatezza, delle, dem, den, der, des, die, difesa, drohe., drohenden, eher, ein, eine, faceva, geben., gibt., idee, intuizioni, kam, keine, letzter, lunghissimo, mehr", mehren, modo., moto, movimento, nelle, neue, nicht, non, nur, occhio, oder, ossessionava, ovvero, peccato), per, periture, poi, privato, psicologia, punte, qualche, quel, quello, recente, restauro, riccioli, ricerche, richiamano, ri-voluzionaria, seit, sich, sogno, solo, solo, sono, stessa, struggente:, subito, sui, und, vada, vergeht, viene, vinciano, vita, volare?, vom, zum

- all'insù, dell'angelo:, dell'aureola, l'esempio, Milano

- Fachkräftemangel

- affrescò, cominciò, jüngst, perché, studierà

- für

- più

**Data:** Mixed script: Greek – Russian

**First run**

- 15o, —, Η

- το, Β, Ηа, α, в, и, на, с

- (наряду, (так, γλωσσών., γλώσσα, γλώσσες., δεσμό., π.Χ., σήμερα., заимство-ваний, латинским), лексика)., литература., слов., человека., язык.

- Ανήκει, Αποτελεί, Στην, έχουμε, αιώνα, ανεξάρτητου, από, βαλκανικό, γλωσ-σικό, γλώσσα, γραπτά, είναι, ελληνική, ενός, επίσης, ινδοευρωπαϊκές, ινδο-ευρωπαϊκής, κείμενα, κλάδου, μέλος, μέχρι, μία, μοναδικό, οικογένειας, στον, της, τις, τον, Римской, богатейшая, большое, была, время, всех, всякого, греческие, греческих, греческого, греческом, двумя, для, древнегреческий, его, знание, значительное, империи, источником, количество, латинских, латинском, лексику, международная, международную, называемая, науч-ных, новое, новых, образованного, обязательным, основном, присутству-ет, проникали, путями, романских, русский, слова, создана, создания, стал, существования, считалось, терминов, технических, церковнославянский, через, этапах, язык, языка, языке

**Second run**

- 15o

- —

- Η

- а, в, и, с

- В

- то, На, на

- (наряду, (так

- γλωσσών., γλώσσα, γλώσσες., δεσμό., π.Χ., σήμερα., заимствований, латинским), лексика)., литература., слов., человека., язык.

- έχουμε, αιώνα, ανεξάρτητου, από, βαλκανικό, γλωσσικό, γλώσσα, γραπτά, είναι, ελληνική, ενός, επίσης, ινδοευρωπαϊκές, ινδοευρωπαϊκής, κείμενα, κλάδου, μέλος, μέχρι, μία, μοναδικό, οικογένειας, στον, της, τις, τον

- Ανήκει, Αποτελεί, Στην

- богатейшая, греческие, греческих, греческого, греческом, древнегреческий, значительное, источником, количество, латинских, латинском, международная, международную, называемая, образованного, обязательным, основном, присутствует, проникали, романских, создания, существования, считалось, терминов, технических, церковнославянский

- Римской, большое, была, время, всех, всякого, двумя, для, его, знание, империи, лексику, научных, новое, новых, путями, русский, слова, создана, стал, через, этапах, язык, языка, языке

**Data:** Mixed script: English – Greek

**First run**

- "intimate, "without, Although, Aquinas, Christians, Corinthians, Socrates, Symposium, Testament, Whether, affection, ancient, another.", appreciation, aspires, attraction, attraction.", becomes, benevolence., biblical, brotherly, chapter,", charity;, children, children.., contemplation, content, continues, contributes, definition:, described, existence;, explained, express, feeling, feelings, finding, further, holding, initially, inspired, knowledge, marriage., necessary, non-corporeal, passage, passion.", philosophers, physical, platonic, refined, relationships, returned, self-benefit)., sensually, spiritual, subject, suggesting, through, throughout, transcendence., unconditional, understanding, without, youthful

- (ἀγάπη, (ἔρως, Agápe, agápē), Éros, érōs), –

- "Form", "erotas", "love, "love, "love:, (even, Agape, Greek, Lovers, Modern, Plato, This, Thomas, also, apply, argue, based, beauty, beauty, being, dating, denote, desire, does, eros, eros., erotic, even, famous, feast., feel, felt, given, good, helps, hence, high, humans, ideal, itself., just, known, leads, like, love, love, love.", mean, means, most, mostly, one's, part, person, person, plane, recall, refer, regard., seek, sexual, soul, spouse, talk, texts, that, there, thus, truth, truth, type, used, well, will, will, with, within, word, work

- "to, 1, 13, God, God.", In, It, New, The, a, all, an, and, any, are, as, be, by, can, esp., for, has, his, in, is, is, it, its, man, not, not, of, on, one, or, own, the, to, us, use, was

## Second run

- affection, ancient, another.", aspires, becomes, biblical, chapter,", charity;, children, children., content, definition:, feeling, feelings, finding, holding, marriage., necessary, passage, passion.", platonic, refined, returned, subject, through, without

- Although, Aquinas, Christians, Corinthians, Socrates, Symposium, Testament, Whether

- "intimate, appreciation, attraction, attraction.", benevolence., brotherly, contemplation, continues, contributes, described, existence;, explained, express, further, initially, inspired, knowledge, non-corporeal, philosophers, physical, relationships, self-benefit)., sensually, spiritual, suggesting, throughout, transcendence., unconditional, understanding, youthful

- Agápe, agápē), Éros, érōs)

- (ἀγάπη, (ἔρως

- –

- "erotas", beauty, beauty, dating, denote, desire, erotic, famous, humans, itself., mostly, person, person, recall, regard., sexual, spouse, within

- "Form", Agape, Greek, Lovers, Modern, Plato, This, Thomas, based, being, feast., hence, ideal, leads, means, plane, refer, there

- apply, felt, helps, high, just, known, most, part, talk, texts, that, thus, truth, truth, type, well, will, will, with, word, work

- "love, "love, "love:, (even, also, argue, does, eros, eros., even, feel, given, good, like, love, love, love.", mean, one's, seek, soul, used

- 1, 13, In, It

- "to, a, an, as, be, by, in, is, is, it, of, on, or, to, us

- God, God.", New, The, all, and, any, esp., its, own, the

- are, can, for, has, his, man, not, not, one, use, was

**Data:** Mixed script: English – Spanish – Arabic

## First run

- El, POW/MIA, Wearing, a, as, been, black, de, displaying, duelo., en, es, estados, for, has, is, lazo, mourning, mourning., negro, o, of, or, organizaciones, personas, political, por, remembrance, remembrance, representando, ribbon, sentimiento, sociedades, statement., symbol, tragedies, un, used, utilizado, y

- crespón, político-social, señal, símbolo

- A

- ٱستخدموا، ٱكثر، ٱن، استخدم، الحاجة، الذي، الرسم، الرمز، العلامات، العلامة، المصريين، المعروف، بنظرة، تنقل، دون، رسالتها، شيء، عن، فٱن، قدماء، كلمات، لاية، معين، من، هم، و، واحدة، والٱغريق، وعموما، ولكن، يعبر، يعني، ينبغي

## Second run

- a, o, y

- El, as, de, en, es, is, of, or, un

- Wearing, been, black, displaying, duelo., estados, for, has, lazo, mourning, mourning., negro, organizaciones, personas, political, por, remembrance, remembrance, representando, ribbon, sentimiento, sociedades, statement., symbol, tragedies, used, utilizado

- POW/MIA

- político-social, símbolo

- crespón, señal

- A

- ٱستخدموا، استخدم، الحاجة، العلامات، العلامة، المصريين، المعروف، رسالتها، والٱغريق، وعموما

- ٱكثر، الذي، الرسم، الرمز، بنظرة، تنقل، دون، شيء، فٱن، قدماء، كلمات، لاية، معين، واحدة، ولكن، يعبر، يعني، ينبغي

- اُن، عن، من، هم

- و

**Data**:   Mixed script: English – Chinese

**First run**

- "crisis", "danger", "opportunity"., (simplified, Chinese, Chinese:, Western, auspicious, because, believed, besides, certain, characters, component, composed, criticized, frequently, inauspicious, invoked, linguists, meanings, meanings., motivational, number, numbers, pinyin:, positive, pronounced, represent, similar, sounds, speaking, tradition, traditional, wēijī)

- (不利), (吉利), 危机;, 危機;, 机;, 機)

- 0, 6, 8, 9

- In, Some, The, and, are, based, be, by, can, for, has, have, in, is, jī, name, names, of, on, or, other, some, sound, that, the, their, this, to, to., two, usage, word, words

**Second run**

- Chinese, Chinese:

- Western

- "crisis", "danger", "opportunity"., (simplified, auspicious, because, believed, besides, certain, characters, component, composed, criticized, frequently, inauspicious, invoked, linguists, meanings, meanings., motivational, number, numbers, pinyin:, positive, pronounced, represent, similar, sounds, speaking, tradition, traditional, wēijī)

- (不利), (吉利)

- 危机;, 危機;

- 机;, 機)

- 6, 8, 9

- 0,

- Some, The, and, are, based, can, for, has, have, name, names, other, some, sound, that, the, their, this, two, usage, word, words

- In, be, by, in, is, of, on, or, to, to.

- jī

**Data:**  Mixed script: Ukrainian – Russian

**First run**

- 9—13

- Белоруссией, Венгрией, Молдавией., Польшей, Россией, Словакией, морей., народів, сарматів, скіфів, століття.

- Азовского, Віддавна, Київська, Румынией, України, , Чёрного, вважається, відправним, , границу, держави, державності, , культури, омывается, пунктом, сухопутную, території, української, існували,

- Имеет, На, Русь, але, водами, готів, и, й, на, с, та, юге, інших

**Second run**

- 9—13

- морей., народів, сарматів, скіфів, століття.

- Белоруссией, Венгрией, Молдавией., Польшей, Россией, Словакией,

- Азовского, Віддавна, Київська, Румынией, України, Чёрного, границу, держави, культури, пунктом, існували

- вважається, відправним, державності, омывается, сухопутную, території, української

- и, й, с

- На, на, та

- але, водами, готів, юге, інших

- Имеет, Русь

**Data:** Pali: abbha

**First run**

- (also, (cp., (dense, (megho, (used, (°sama, 1, 1, 101, 1064;, 12)., 134., 239., 249, 250;, 251)., 273, 289., 295, 3, 317, 348, 348)., 382)., 487, 53, 581, 617, 64;, 687, <at> a)fro\\s </at>, <at> o)/mbros </at>, <smallcaps> i. </smallcaps>, <smallcaps> ii. </smallcaps>, <smallcaps> iv. </smallcaps>, <smallcaps> vi. </smallcaps>, <superscript> 1 </superscript>, <superscript> 9 </superscript>, <superscript> s. </superscript>, A, A), As, Dh, Dhs, DhsA, Gr., Idg., It, J, Lat., Miln, Miln), Oir, Pv, PvA, S, Sk., Sn, SnA, Th, This, Vin, Vv, VvA, [Vedic, a, abhra, adj., also, ambha, ambu, as, at, by, cloud, cloud, cloud\";, clouds, clouds, cloudy, cp., dark), expl, f., free, from, froth, imber, in, is, later, like, list, m., marajo</b>, mass, moon−, mountain, nt., obscure, of, or, pabbata, perhaps, point, rain, rain;, rajo, referred, scum, sense, storm−cloud, summit, sunshine, that, the, thick, things, thunder−cloud);, thundering, to, viz., water, water].

- &, (=, <b>−ghana</b>, <b>−mutta</b>, <br, =, \"dark, \"dull\";

- (abbhaŋ, (mahiyā, (nīl°, <b> −saŋvilāpa </b>, <b> Rāhu </b>, <b> abbhā </b>, <b> abbhāmatta </b>, abbhāmutta, acchādesi);, mahikā </b>, nīla−megha, valā− haka);, valāhaka− sikhara

- *ṁbhrocite /><b>−kūṭa</b>, <b>−paṭala</b>, <b>abbhaŋ, <b>dhū-, (nt.)

**Second run**

- (cp., Dhs, DhsA, Idg., Lat., Miln, Miln), Oir, PvA, SnA, This, Vin, VvA, [Vedic, as, at, by, cp., in, is, nt., of, or, to

- (also, (dense, (megho, (used, (°sama, <at> a)fro\\s </at>, <at> o)/mbros </at>, <smallcaps> ii. </smallcaps>, <smallcaps> iv. </smallcaps>, <smallcaps> vi. </smallcaps>, abhra, adj., also, ambha, ambu, cloud, cloud, cloud\";, clouds, clouds, cloudy, dark), expl, free, from, froth, imber, later, like, list, marajo </b>, mass, moon−, mountain, obscure, pabbata, perhaps, point, rain, rain;, rajo, referred, scum, sense, storm− cloud, summit, sunshine, that, the, thick, things, thunder− cloud);, thundering, viz., water, water].

- 1, 1, 101, 1064;, 12)., 134., 239., 249, 250;, 251)., 273, 289., 295, 3, 317, 348, 348)., 382)., 487, 53, 581, 617, 64;, 687, <superscript> 1 </superscript>, <superscript> 9 </superscript>

- <smallcaps> i. </smallcaps>, <superscript> s. </superscript>, A, A), As, Dh, Gr., It, J, Pv, S, Sk., Sn, Th, Vv, a, f., m.

- <b> −ghana </b>, <b> −mutta </b>, <br, \"dark, \"dull\";

- &, (=, =

- (abbhaŋ, (mahiyā, (nīl°, <b> Rāhu </b>, <b> abbhā </b>, nīla−megha

- <b> −saŋvilāpa </b>, <b> abbhāmatta </b>, abbhāmutta, acchādesi);, mahikā </b>, valāhaka);, valāhaka−sikhara

- *m̐bhro, /><b> −kūṭa </b>, <b> −paṭala </b>, <b> abbhaŋ, <b> dhū-

- (nt.)

**Data:**    Pali: abhijjhitar

## First run

- abhijjhita, abhijjhātar, covets, function], med., one, who, °itar), °itar, °ātar).

- (T., <smallcaps> i. </smallcaps>, <smallcaps> v. </smallcaps>, =, A, M, ag., fr., in, l., v.

- 265, 287

- [n.

## Second run

- abhijjhita, abhijjhātar, covets, function], med., one, who, °itar), °itar, °ātar).

- (T., A, M

- =, l., v.

- <smallcaps> i. </smallcaps>, <smallcaps> v. </smallcaps>, ag., fr., in

- 265, 287

- [n.

**Data:** Pali: ajja

## First run

- −divasa, Freq., Loc., [Vedic, adya, ajjatagge, ajjato, an, and, as, base, being, day, demonstr., dyaus, from, in, morning, not, of, old, or, phrase, present, pron., the, this, with

- &, +, Mhvs, s., v.

- −kālaṁ, 10）, 15, 64., 32, 23., Ajjā, D.I, 85；, DA.I, 235., Dh.326；, III, 425, J.I, 279；, J.VI, 180；, Kern, Toev., Pv.I, 117, PvA.59）；, PvA.6, 23；, Sn.75, 153, 158, 970, 998；, Vin.I, 18；, a3）, adyā, a, agga3）, agge（?）, ajja-tagge, see, ajjā；, bahutaṁ, day〞], diva）, thus, dyā, dyā, a°, idāni, onward，henceforth, to-day，now，"food"）；, "on, ‹-›, Ajja, &,（=,（Page,（adv.）,（read,（see

## Second run

- an, as, in, of, or

- Freq., Loc., [Vedic

- −divasa, adya, ajjatagge, ajjato, and, base, being, day, demonstr., dyaus, from, morning, not, old, phrase, present, pron., the, this, with

- &

- +

- Mhvs

- s., v.

- "on, ‹-›, Ajja, &,（=,（Page,（adv.）,（read,（see

- −kālaṁ, 10）, 15, 64., 32, 23., Ajjā, D.I, 85；, DA.I, 235., Dh.326；, III, 425, J.I, 279；, J.VI, 180；, Kern, Toev., Pv.I, 117, PvA.6, 23；, Sn.75, 153, 158, 970, 998；, Vin.I, 18；, a3）, agga3）, ajja-tagge，see, ajjā；, bahutaṁ, day〞], diva）, thus, dyā, idāni, onward，henceforth, to-day，now

- PvA.59）；, adyā, a, agge（?）, dyā, a°, "food"）；

**Data:** Pali: gūhanā

**First run**

- 253）, Pug. 19. Cp. pari°.（Page, [abstr. fr. gūhati]=gūhanā, Gūhanā,（f.）,（q. v.）

**Second run**

- 253）, Pug. 19. Cp. pari°.（Page, [abstr. fr. gūhati]=gūhanā, Gūhanā,（f.）,（q. v.）

**Data:** Pali: pacati

**First run**

- 382）, Caus. pacāpeti, DA. I, 159，where, Obulg. peka, Pass. paccati, Vin. IV, 264；, bake，Gr. péssw, cook，boil，roast, cook，pépwn, daṇḍena, fig. torment, fry，roast，Lith，kepū, intrs. ）: Niraye, paccato，by, ppr. pacanto, pp. pakka, pīḷentassa). –, tormenting，Gen. pacato

- D. I, 52, N. S. II, 225, PvA. 10, 14. –, Pacati, [Ved. pacati, Idg. *pequ̯ō, Av. pac-；,（+Caus. pācayato）,（expld,（q. v.）. –,（q. v.）. <->,（q. v.）.（Page,（trs. and

- after, at, be, for, in, or, pacato, pare, purgatory, read, ripe], roasted, roasting, to, tormented

- &, pacitvā, pāceti

**Second run**

- Caus. pacāpeti, DA. I, 159，where, Obulg. peka, Pass. paccati, Vin. IV, 264；, bake，Gr. péssw, cook，boil，roast, cook，pépwn, daṇḍena, fig. torment, fry，roast，Lith，kepū, intrs. ）: Niraye, paccato，by, ppr. pacanto, pīḷentassa). –, tormenting，Gen. pacato

- 382）, pp. pakka

- D. I, 52, N. S. II, 225, PvA. 10, 14. –,（q. v.）. –

- Pacati, [Ved. pacati, Idg. *pequ̯ō, Av. pac-；,（+Caus. pācayato）,（expld,（q. v.）. <->,（q. v.）.（Page,（trs. and

- for, pacato, pare, read, ripe]

- after, purgatory, roasted, roasting, tormented

- or, to

- at, be, in

- &

- pacitvā, pāceti

**Data:** Twitter 1 (Greek–English)

**First run**

- αυτή, διαγωνισμό, λύση, στο, τη, ψήφισα, Μόλις

- BUSINESS, EXCELLENCE., IT, Internet, Things, of

**Second run**

- Μόλις

- αυτή, διαγωνισμό, λύση, στο, τη, ψήφισα

- IT, of

- Internet, Things,

- BUSINESS, EXCELLENCE.

**Data:** Twitter 2 (French–English)

**First run**

- "The, 18h, @dhiparis, David, Demain, Keynote, all, collective, counts?", dynam-ics, par, perish;, science-publish, that

- is, it, of, or

**Second run**

- "The, @dhiparis, David, Demain, Keynote, all, collective, counts?", dynamics, par, perish;, science-publish, that

- 18h

- is, it, or

- of

**Data:** Twitter 3 (French–English)

**First run**

- Edmonton, Food

- go, in, to

- and, are, breuvages, fans, for, just, ready, the, waiting

**Second run**

- Edmonton, Food

- to

- go, in

- for, just

- and, are, breuvages, fans, ready, the, waiting

**Data:** Twitter 4 (English–Polish)

**First run**

- żubrówka, my

- adidas, and, back, comes, crates, dad, from, jackets, of, omg, poland, strawberries, two, with

**Second run**

- żubrówka, my

- adidas, comes, dad, of

- and, back, crates, from, jackets, omg, poland, strawberries, two, with

**Data:** Twitter 5 (Transliterated Amharic–English)

**First run**

- Buna

- (coffee, bread)., dabo, is, naw, our

**Second run**

- Buna

- our

- (coffee, bread)., dabo, is, naw

### 8.3.4 Language Model Induction

For all language model induction tasks, the threshold value $t$ has been set $t = 0.02$ and the silver threshold value $s$ has been set $s = 0.1$. The other parameters have been set to "maximum iteration count" $i = 4$, "maximum random iteration count" $j = 2$ and "merge mode ADD".

**Data:** Latin script: German–English

- The, German, word, Nabelschau, means, or, "staring, at, your, But, in, this, it, doesn't, refer, to, anyone, else's, button, just, your, own.,

- –

- "navel-gazing", navel"., case, belly

**Data:** Latin script: German–Finnish–Turkish

- die, in, und, Klimazone., Je, ob, auf, Südhalbkugel, vom, eli, on, vuodenaika, ja, on, vuodenajoista, koska, maapallo, on, silloin, kallistunut, aurinko, maan, pinnalle, kulmassa, muina, vuodenaikoina., Pohjoisella, pallonpuoliskolla, lasketaan, tavallisesti, ja, elokuu, eteläisellä, pallonpuoliskolla, joulu-, ja, helmikuu., en, sıcak, en, yazda, Dünya, depo, en, sıcak, yaklaşık, ay, sonra, ortaya, Sıcak, Haziran, Eylül, ise, Aralık, arasındadır.

- Der, ist, wärmste, der, vier, Jahreszeiten, der, arktischen, nachdem, er, der, Nord-, oder, herrscht, spricht, Nord-, oder, Der, findet, mit, Südwinter, statt., suvi, lämpimin, niin, että, säteilee, heinä-, Yaz, mevsimdir., Küre'de, Küre'de, 21, 22, arasında, Küre'de, 22, 21, Mart

- gemäßigten, gerade, gleichzeitig, kuin, Kuzey, uzun, günler, gerçekleşir., ettiği, için, günler, genellikle, iki, günler, Kuzey, ile, ile

- Sommer, man, Südsommer., Nordsommer, dem, Kesä, kevään, syksyn, välissä., Kesä, jyrkemmässä, kesäkuukausiksi, kesä-., tammi-, Yarım, ısıyı, çıkar., Yarım, Güney, Yarım

**Data:** Latin script: English–French

- both, "soft", in, English, although, their, is, is, the, opposite, of, "rough", or, is, the, opposite, of, sweet, only, for, wines, (otherwise, is

- mou, :, mou, but

- doux,

- Doux, (rugueux), Doux

- while

- "hard"., used).,

- translate, as, meaning, very, different., "coarse", can, also, mean, almost,sucré,

**Data:** Latin script: English–Transliterated Greek

- at, least, ways, as, to, is, has, philía, and, storgē., as, has, historically, difficult, to, which, generally, as

- The, language, distinguishes, different, the, Ancient, distinct, with, languages, it, been, separate, the, meanings, these, used, outside, their, respective, the, senses, in, these, used

- Greek, how, word, Greek, agápe, éros, However, other, when, were, are

- four, love, used., four, words, for, love:, of, words, of, contexts., Nonetheless, words, follows.

**Data:**   Latin script: Italian–German

- affrescò, privato, Studie, definire, periture,Stifterverbands, Wissenschaft,studierà, difesa, ovvero, Szenario, Naturwissenschaftlern

- dell' aureola, da, del, di, der, zum, modo., dem, den, drohe., Come, vom

- custodisce, quel, es, oder, per, le, idee, stessa, des, dass, delle, E, se, Ist, das, seit

- più, Cenacolo, vinciano, rivoluzionaria, Giuda, condanna, con, peccato), cominciò, con, cancro, faceva, intuizioni, vita, va, Dabei, Ergebnis, in, i, riccioli, poi, più, bacini, in, Annunciazione, con, ali, la, cosa, barbaglio, anni, bei,

- ne, struggente:, che, amore, e, non, viene, ma, consapevolezza, ad, che, ha, recente, Kaum, eine, Woche, vergeht, keine, neue, Umfrage, Warnung, Thema, Fachkräftemangel, Deutschland, Certo, ma, anche, consapevole, che, qualche, mehren, letzter, Zeit, Stimmen, Entwarnung, geben., kam, jüngst, eine, Deutsche, "ein, allgemeiner, Fachkräftemangel, eher, mehr", anche, Battista, che, Leonardo, approfonditamente, a, Venezia, nelle, vada, alla, attento, alle, dell' angelo:, delicatezza, punte, che, non, che, volare?, Jahren, angemahnte, drohenden, Fachkräftemangel, Ingenieuren, ein

- Milano, l' esempio, psicologia, (il, subito, autodistruggersi, solo, lunghissimo, So, il, movimento, moto, sui, si, bellissima, occhio, all' insù, sono, sogno, lo, ossessionava, quello, und, also, Mythos?

- un, für, MINT-Berufen

- cura, restauro, arginato., gibt., perché, catturata, sich, auch, zu, nicht, richiamano, acque, ricerche, chiave, anti-Turchi., nur

**Data:** Mixed script: Greek–Russian

- ελληνική, γλώσσα, είναι, μία, από, τις, ινδοευρωπαϊκές, γλώσσες., Αποτελεί, το, μοναδικό, μέλος, ενός, ανεξάρτητου, κλάδου, της, ινδοευρωπαϊκής, οικογένειας, γλωσσών., Ανήκει, επίσης, στον, βαλκανικό, γλωσσικό, δεσμό., Στην, ελληνική, γλώσσα, έχουμε, γραπτά, κείμενα, από, τον, 15ο, αιώνα, μέχρι, σήμερα.

- На, греческом, на, всех, его, существования, была, создана, богатейшая, греческого, обязательным, всякого, образованного, большое, заимствова-ний, а, в, греческом, новое, время, (наряду, новых, научных, терминов, на-зываемая, международная, слова, в, основном, двумя, через

- Н, п.Х.,языке, этапах, литература., В, Римской, империи, знание, языка, считалось, для, человека., В, латинском, языке, присутствует, количество, греческих, —, значительное, количество, латинских, и, романских, слов., В, древнегреческий, язык, стал, с, латинским), источником, создания, и, тех-нических, (так, лексика)., В, русский, язык, греческие, проникали, путями, —, международную, лексику, и, церковнославянский, язык.

**Data:** Mixed script: English–Greek

- is, biblical, is, will, is, without, self-benefit)., is, feelings, feelings, it, be, feeling, being, high, is, by, his, This, by, will, mostly, sexual, "intimate, well, refined, his, definition:, is, initially, felt, with, it, beauty, within, beauty, itself., use, "with-out, helps, soul, beauty, spiritual, youthful, beauty, feel, suggesting, sensually, spiritual, finding, its, like, finding, all, seek

- (ἀγάπη, (ἔρως

- Agápe, "love:, brotherly, love, love, of, God, for, of, for, in, known, "love, 1, 13, throughout, New, brotherly, love, affection, good, love, love, given, or, not, per-son, continues, love, (even, in, for, one's, for, spouse, refer, love, of, content, or, holding, one, in, unconditional, love, of, God, for, of, love, "to, good, of, Éros, "love, of, The, Modern, Greek, word, love.", own, Although, eros, for, person, contemplation, becomes, of, person, or, even, becomes, of, not, of, of, love, of, word, mean, In, Symposium, work, on, subject, eros, knowledge, of, of, "Form", of, erotic, –, even, love, non-corporeal, of, is, Lovers, philosophers, through, of,

- agápē), means, esp., charity;, the, man, and, man, God.", Agape, used, the, pas-sage, as, the, chapter;", Corinthians, and, described, there, and, the, Testament, as, and, benevolence., Whether, the, returned, the, to, any, Agape, also, used, ancient, texts, to, denote, children, and, the, a, and, was, also, used, to, to, a, feast., It, can, also, described, as, the, regard., Agape, used, Christians, to, express, the, children., type, was, further, explained, Thomas, Aquinas, as, the, another.",
érōs), means, the, passion.", "erotas", means, It, can, also, apply, to, dating, re-
lationships, as, as, marriage., Plato, a, an, appreciation, the, that, appreciation,
Plato, does, talk, physical, attraction, as, a, necessary, part, hence, the, the, pla-
tonic, to, physical, attraction.", the, the, most, famous, ancient, the, Plato, has,
Socrates, argue, that, the, recall, and, contributes, to, an, understanding, truth,
the, ideal, that, leads, us, humans, to, desire, thus, that, that, based, aspires, to,
the, plane, existence;, that, truth, just, any, truth, leads, to, transcendence., and,
are, inspired, to, truth, the, means, eros.

**Data:**    Mixed script: English–Spanish–Arabic

- الرمز، يعني، الرَسم، الذي، يعبر، عن، شيء، معين، وعموما، فُلان، العلامة، ينبغي، أن، تنقل، رسائلها، بنظرة، واحدة، دون،
الحاجة، لأية، كلمات، و، من، المعروف، أن، قدماء، المصريين، والأغريق، استخدموا، العلامات، ولكن، أكثر، من، استخدم،
العلامات، هم

- ribbon, symbol, mourning., ribbon, mourning, El, un, y, un, en

- black, is, a, of, remembrance, or, Wearing, or, displaying, a, black, has, been, used,
for, remembrance, tragedies, or, as, a, political, statement., crespón, negro, o, lazo,
negro, es, símbolo, utilizado, por, personas, estados, sociedades, organizaciones,
representando, sentimiento, político-social, señal, de, duelo.

- A, POW/MIA

**Data:**    Mixed script: English–Chinese

- The, Chinese, (simplified, traditional, Chinese:, invoked, motivational, speaking,
because, the, composed, characters, that, represent, linguists, have, criticized,
this, usage, because, the, component, (simplified, Chinese:, traditional, Chinese:,
has, other, besides, Chinese, certain, some, be, based, the, Chinese, that, the, The,
numbers, believed, have, because, their, similar, words, that, have, positive

- (不利)

- Western, can, and, Some, meanings, In, are, number, name, and, are, meanings,
names, meanings.

- 0, 6, 8, 9

- "crisis", is, auspicious, inauspicious, sounds, sound

- for, pinyin:, frequently, in, word, of, two, "danger", "opportunity"., pronounced,
tradition, by, or, on, word, to., to

- 危机;, 危機;, wēijī), jī, 机;, 機), (吉利)

**Data:** Mixed script: Ukrainian–Russian

- й, Русь, морей., Россией, Белоруссией, Польшей, Словакией, Венгрией, Румынией

- але, 9—13, юге, Имеет, Молдавией.

- існували, інших

- Чёрного, Азовского, границу

- культури

- території, України, пунктом, української, и, сухопутную, и

- Віддавна, на, держави, скіфів, сарматів, готів, народів, відправним, державності, На, водами

- та, вважається, Київська, століття., омывается, с

**Data:** Pali: abbha

- (nt.), nt., Sk., \"dark, Idg., cp., Gr., Lat., Sk., water, Gr., water]., dark), at, SnA, S, at, It, Sn, (cp., SnA, Sn, S

- &, A, A), ., J, 251)., 1, 1064;, 249, 250;, 12)., 64;, 348)., 382).

- viz., 134., 101, 581, f., 289.

- 53, 295, 273, 487, 3, 617, 317, 348, 239., 687

- cloud\";, also, cloud, cloudy, <smallcaps> ii. </smallcaps>, =, list, is, <smallcaps> i. </smallcaps>, (°sama, <smallcaps> vi. </smallcaps>, (abbhaŋ, <smallcaps> iv. </smallcaps>, (nīl°, As, Dhs, DhsA, (used, (=, clouds, cloud, (also, as

- m., adj.

- abhra, (mahiyā, VvA, acchādesi);, Pv, PvA, \"dull\";, valāhaka);, Vv, valāhaka–sikhara

- <at>a)fro\\s</at>, froth, of, <superscript> 9 </superscript>, <superscript> s. </superscript>, <superscript> 1 </superscript>

- later, scum, rain;, ambha, rain, a, Miln, (megho, Miln), nīla–megha, sense, expl, Th, Dh

- *m̐bhro, <at>o)/mbros</at>, ambu, mass, to, obscure, moon−, <b>abbhaŋ, mahikā</b>, <b>dhū−, marajo</b>, <b>Rāhu</b>, pabbata, rajo, <b>abbhā</b>, by, perhaps, <b>abbhāmatta</b>, <br /><b>−kūṭa</b>, or, summit, storm−cloud, <b>−ghana</b>, <b>−paṭala</b>, mass, <b>−mutta</b>, from, abbhāmutta, <b>−saŋvilāpa</b>,

- [Vedic, imber, Oir, (dense, Vin, in, things, that, sunshine, This, referred, mountain, like, thunder−cloud);, the, point, thick, free, thundering

**Data:** Pali: abhijjhitar

- <smallcaps>i.</smallcaps>, v., l., <smallcaps>v.</smallcaps>

- abhijjhita, abhijjhātar, °itar), °itar, °ātar).,

- [n., ag., fr., med., M, 287, (T., =, A, 265

- in, function], one, who, covets

**Data:** Pali: ajja

- Ajja, &, Ajjā, （adv.）, base, a3）, diva）, thus, Dh.326；, ajjā；, v., PvA.59）；, PvA.6, 23；, phrase, ajjatagge, ajjato, agge（?）, ajja-tagge, see, agga3）, （adv.）, the, 32, 23., （Page

- ‹-›, −kālaṁ

- [Vedic, &, +, being, （see, （see, （read, as, （=, Mhvs, （=, +, Mhvs

- of, of, "on, "food"）；

- and, an, old, not

- adya, adyā, a, dyā, a°, dyā, dyaus, day"], to-day, now, bahutaṁ, with, day, −divasa, day

- demonstr., pron., Loc., this, Kern, Toev., s., Freq., or, from, this, onward, henceforth, this, morning, present

- Sn.75, 153, 158, 970, 998；, J.I, 279；, III, 425, Pv.I, 117, idāni, 15, 64., in, Vin.I, 18；, D.I, 85；, DA.I, 235., J.VI, 180；, 10）

**Data:** Pali: gūhanā

- Pug. 19. Cp. pari°.（Page

- Gūhanā,（f.）, [abstr. fr. gūhati]=gūhanā

- 253）,（q. v.）

**Data:** Pali: pacati

- Vin. IV, 264；, N. S. II, 225, PvA. 10, 14. –, D. I, 52

- DA. I, 159, where, 382）

- in

- at, &

- cook, pépwn, cook, boil, roast

- Pacati, [Ved. pacati, Idg. *pequō, Av. pac-；, Obulg. peka, to, fry, roast, Lith, kepū, ripe], to, fig. torment, purgatory,（trs. and, pacitvā, after, roasting, ppr. pacanto, tormenting, Gen. pacato,（+Caus. pācayato）, read, pacato, for, paccato, by, pare, pp. pakka, Caus. pacāpeti, pāceti, Pass. paccati, to, roasted, or, tormented

- bake, Gr. péssw, intrs.）: Niraye,（expld, daṇḍena, pīḷentassa）. –,（q. v.）. <->,（q. v.）. –, be,（q. v.）.（Page

**Normalized data**

- pacati, peka, péssw, pépwn, pacitvā, ppr., pacanto, Gen., pacato, (+Caus., pācay-ato), pacato, paccato, pare, pīḷentassa)., pp., pakka, Caus., pacāpeti, Pass., paccati

- *pequō, bake

- pac-;, 264;, 52, &, 382)

- 10,14.–, 159, –, <->, –

- fry, Niraye, I, I, by

- Av., Obulg., Gr., (trs., D., DA., (q.v.)., (q.v.)., (q.v.).

- [Ved., to, roast, kepū, cook, ripe], to, cook, roast, torment, purgatory, and, after, roasting, tormenting, (expld, at, where, read, for, daṇḍena, pāceti, to, be, roasted, or, tormented, (Page

- Pacati, Idg., Lith, boil, Vin.IV, fig., in, intrs.):, in, N.S.II,225,PvA.

**Data:** Twitter 1 (Greek–English)

- BUSINESS, EXCELLENCE.
- Μόλις, ψήφισα, αυτή, τη, λύση, Internet, of, στο, διαγωνισμό
- Things, IT

**Data:** Twitter 2 (French–English)

- Keynote, "The, collective, of, science-publish, or, perish;, it, all, that, counts?"
- Demain, 18h, par
- #dhiha6, David
- @dhiparis, dynamics, is

**Data:** Twitter 3 (French–English)

- #FWWC2015
- breuvages, go,
- Food, Edmonton, to, for, the
- in, waiting, #bilingualism
- and, are, ready, just, fans

**Data:** Twitter 4 (English–Polish)

- comes, from, with, two, crates, of, strawberries, jackets, omg
- my, dad, poland, and, adidas
- back, żubrówka

**Data:** Twitter 5 (Transliterated Amharic–English)

- (coffee
- bread). is, our
- Buna, dabo, naw